\documentclass[sigconf]{acmart}

\AtBeginDocument{%
  \providecommand\BibTeX{{%
    \normalfont B\kern-0.5em{\scshape i\kern-0.25em b}\kern-0.8em\TeX}}}
\copyrightyear{2023}
\acmYear{2023}
\setcopyright{acmlicensed}\acmConference[SIGSPATIAL '23]{The 31st ACM International Conference on Advances in Geographic Information Systems}{November 13--16, 2023}{Hamburg, Germany}
\acmBooktitle{The 31st ACM International Conference on Advances in Geographic Information Systems (SIGSPATIAL '23), November 13--16, 2023, Hamburg, Germany}
\acmPrice{15.00}
\acmDOI{10.1145/3589132.3625604}
\acmISBN{979-8-4007-0168-9/23/11}

\usepackage{graphicx}
\usepackage{amsmath}
\usepackage{amsfonts}
\usepackage{algorithm}
\usepackage[noend]{algpseudocode}
\usepackage[skip=0.5\baselineskip,font=normalsize]{caption}
\usepackage{subcaption}
\usepackage{nicefrac}
\usepackage{multirow}
\usepackage{array}
\usepackage{booktabs}
\usepackage{times}
\usepackage{epsfig}

\usepackage{hyperref}       

\usepackage[capitalize]{cleveref}
\crefname{section}{Sec.}{Secs.}
\Crefname{section}{Section}{Sections}
\Crefname{table}{Table}{Tables}
\crefname{table}{Tab.}{Tabs.}

\newcolumntype{L}{>{\arraybackslash}p{2.6cm}}
\newcolumntype{X}{>{\centering\arraybackslash}p{1.8cm}}
\newcolumntype{Y}{>{\centering\arraybackslash}p{1.1cm}}
\newcolumntype{Z}{>{\centering\arraybackslash}p{1.15cm}}
\newcolumntype{M}{>{\centering\arraybackslash}p{1.8cm}}
\newcolumntype{N}{>{\centering\arraybackslash}p{2.6cm}}
\newcolumntype{V}{>{\centering\arraybackslash}p{3.9cm}}
\newcolumntype{U}{>{\centering\arraybackslash}p{3.3cm}}
\newcolumntype{K}{>{\centering\arraybackslash}p{1.3cm}}
\newcolumntype{J}{>{\centering\arraybackslash}p{1.8cm}}
\newcolumntype{A}{>{\centering\arraybackslash}p{4.9cm}}
\newcolumntype{B}{>{\centering\arraybackslash}p{0.8cm}}
\newcolumntype{C}{>{\centering\arraybackslash}p{1.2cm}}
\newcolumntype{D}{>{\centering\arraybackslash}p{3.0cm}}
\newcolumntype{E}{>{\centering\arraybackslash}p{1.2cm}}

\algnewcommand\algorithmicinput{\textbf{Input:}}
\algnewcommand\INPUT{\item[\algorithmicinput]}
\algnewcommand\algorithmicoutput{\textbf{Output:}}
\algnewcommand\OUTPUT{\item[\algorithmicoutput]}
\algnewcommand\algorithmicparameters{\textbf{Parameters:}}
\algnewcommand\PARAMETERS{\item[\algorithmicparameters]}
\algnewcommand\algorithmicinitialize{\textbf{Initialize:}}
\algnewcommand\INITIALIZE{\item[\algorithmicinitialize]}
\algnewcommand\algorithmicbreak{\textbf{break}}
\algnewcommand\BREAK{\item[\algorithmicbreak]}
\algtext*{EndWhile}
\algtext*{EndIf}
\algtext*{EndFor}
\algtext*{EndFunction}
\algtext*{EndProcedure}
\algrenewcommand\textproc{\texttt}
\algnewcommand{\LineComment}[1]{\State \(\triangleright\) \textit{#1}}

\newbox{\bigpicturebox}

\newtheorem{definition}{Definition}[section]

\begin{document}
\setlength{\abovedisplayskip}{7pt}
\setlength{\belowdisplayskip}{7pt}

\setkeys{Gin}{draft=false}

\title{SeMAnD: Self-Supervised Anomaly Detection in Multimodal Geospatial Datasets}

\author{Daria Reshetova}
\authornote{Equal contribution.}
\email{resh@stanford.edu}
\affiliation{
    \institution{Stanford University}
    \city{Stanford}
    \state{CA}
    \country{USA}
    \postcode{94305}
}

\author{Swetava Ganguli}
\authornotemark[1]
\authornote{Corresponding Author. Alternative email: swetava@cs.stanford.edu}
\email{swetava@apple.com}
\affiliation{%
    \institution{Apple}
    \city{Cupertino}
    \state{CA}
    \country{USA}
    \postcode{95014}
}

\author{C. V. K. Iyer, V. Pandey}
\email{{cvk,vipul}@apple.com}
\affiliation{%
    \institution{Apple}
    \city{Cupertino}
    \state{CA}
    \country{USA}
    \postcode{95014}
}

\renewcommand{\shortauthors}{Reshetova and Ganguli, et al.}

\begin{abstract}
We propose a \textbf{Se}lf-supervised \textbf{An}omaly \textbf{D}etection technique, called \textbf{SeMAnD}, to detect geometric anomalies in \textbf{M}ultimodal geospatial datasets. Geospatial data comprises of acquired and derived heterogeneous data modalities that we transform to semantically meaningful, image-like tensors to address the challenges of representation, alignment, and fusion of multimodal data. SeMAnD is comprised of (i) a simple data augmentation strategy, called \textit{RandPolyAugment}, capable of generating diverse augmentations of vector geometries, and (ii) a self-supervised training objective with three components that incentivize learning representations of multimodal data that are discriminative to local changes in one modality which are not corroborated by the other modalities. Detecting local defects is crucial for geospatial anomaly detection where even small anomalies (e.g., shifted, incorrectly connected, malformed, or missing polygonal vector geometries like roads, buildings, landcover, etc.) are detrimental to the experience and safety of users of geospatial applications like mapping, routing, search, and recommendation systems. Our empirical study on test sets of different types of real-world geometric geospatial anomalies across 3 diverse geographical regions demonstrates that SeMAnD is able to detect real-world defects and outperforms domain-agnostic anomaly detection strategies by 4.8--19.7\% as measured using anomaly classification AUC. We also show that model performance increases (i) up to 20.4\% as the number of input modalities increase and (ii) up to 22.9\% as the diversity and strength of training data augmentations increase.   
\end{abstract}

\ccsdesc[500]{Computing methodologies~Computer vision}
\ccsdesc[500]{Computing methodologies~Anomaly detection}
\ccsdesc[500]{Computing methodologies~Learning latent representations}

\keywords{Self-Supervised Learning, Multimodal Learning, Anomaly Detection, Computer Vision}

\maketitle

\section{Introduction}\label{section:introduction}
Anomaly detection is the task of identifying data instances containing anomalous or defective patterns causing them to significantly deviate from some concept of normality \cite{chandola2009anomaly}. Supervised approaches for anomaly detection are impractical due to the acute difficulty or infeasibility of producing labeled datasets since (i) anomalies are rare and can be diverse and heterogeneous in nature, (ii) different classes of anomalies may have large variations between them and their nature remains unknown until they occur, and (iii) anomalies can be correlated to the aleatoric and epistemic uncertainties in the data generation process. Instead, unsupervised learning of a model of \textit{normality} from error-free (\textit{normal}) data is a central theme in anomaly detection \cite{ruff2021unifying,salehi2021unified}. The synergy between this theme and self-supervised learning techniques, which utilize vast amounts of unlabeled data to learn semantically meaningful universal representations of the data, has made self-supervised anomaly detection a successful and popular approach \cite{hojjati2022self,reiss2023anomaly}. Self-supervised learning (SSL) uses \textit{auxiliary} or \textit{pretext} tasks with training objectives similar to those used for supervised learning while obtaining supervision signals directly from unlabeled data by making neural networks predict withheld or altered parts or properties of the inputs \cite{chen2020,caron2021emerging,he2022masked}.  

Geospatial datasets are diverse, naturally spatiotemporal, and inherently multimodal (composed of two or more distinct signal types or modalities) e.g., satellite/aerial imagery (RGB, multispectral), road network graphs, vector geometry, sensor data (LiDAR-based point cloud, IMU data, GPS-based mobility data), active sensing imagery (SAR, Sonar, Radar). The richness of geospatial datasets results from the rapid growth in sensor technologies and their deployment for measuring variegated geospatial signals.  

\begin{figure}
	\centering
	\scriptsize
	\begin{subfigure}[b]{0.092\textwidth}
		\centering
		\includegraphics[width=\textwidth]{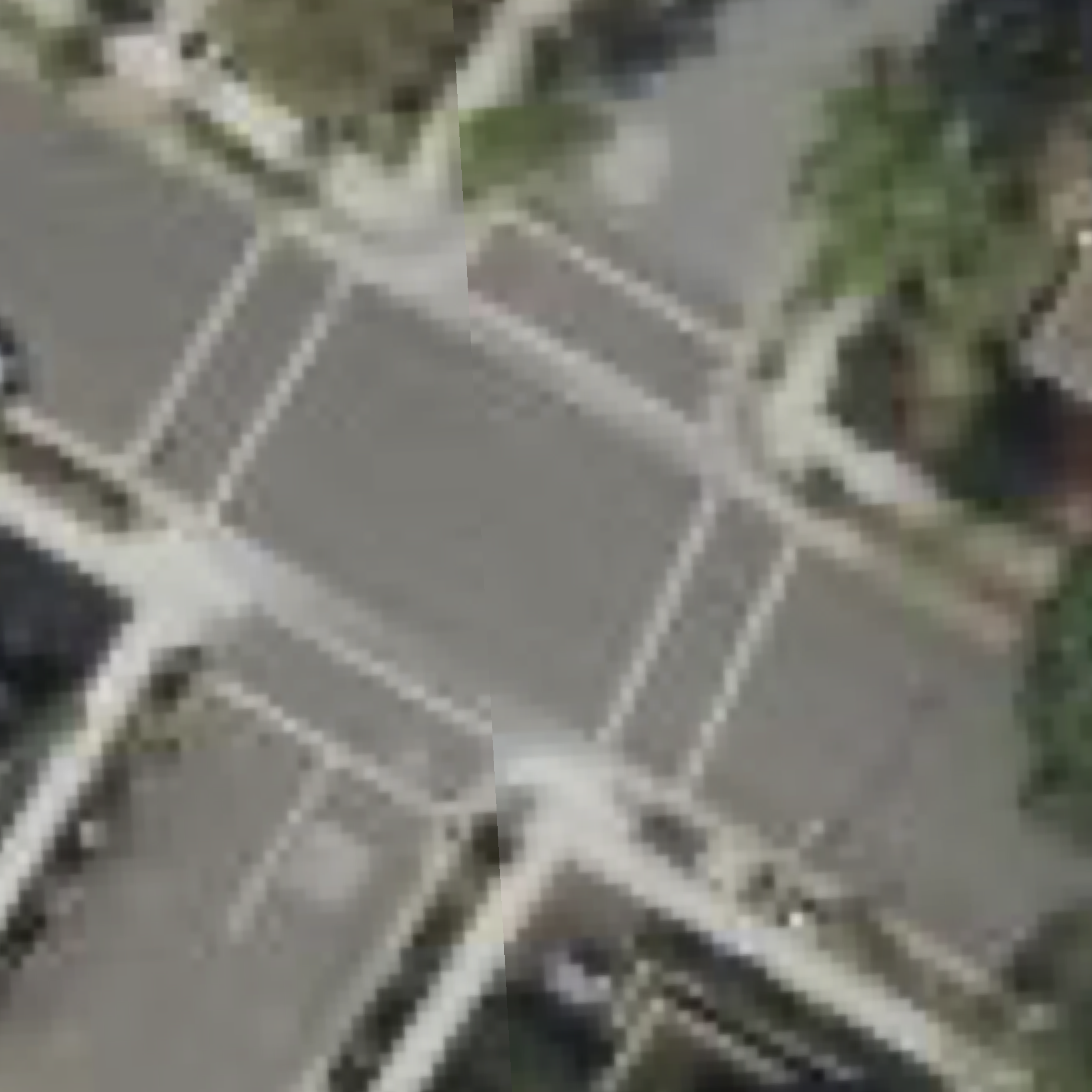}
		\caption{}
		\label{fig::example_mulimodal_satellite}
	\end{subfigure}
	\hfill
	\begin{subfigure}[b]{0.092\textwidth}
		\centering
		\includegraphics[width=\textwidth]{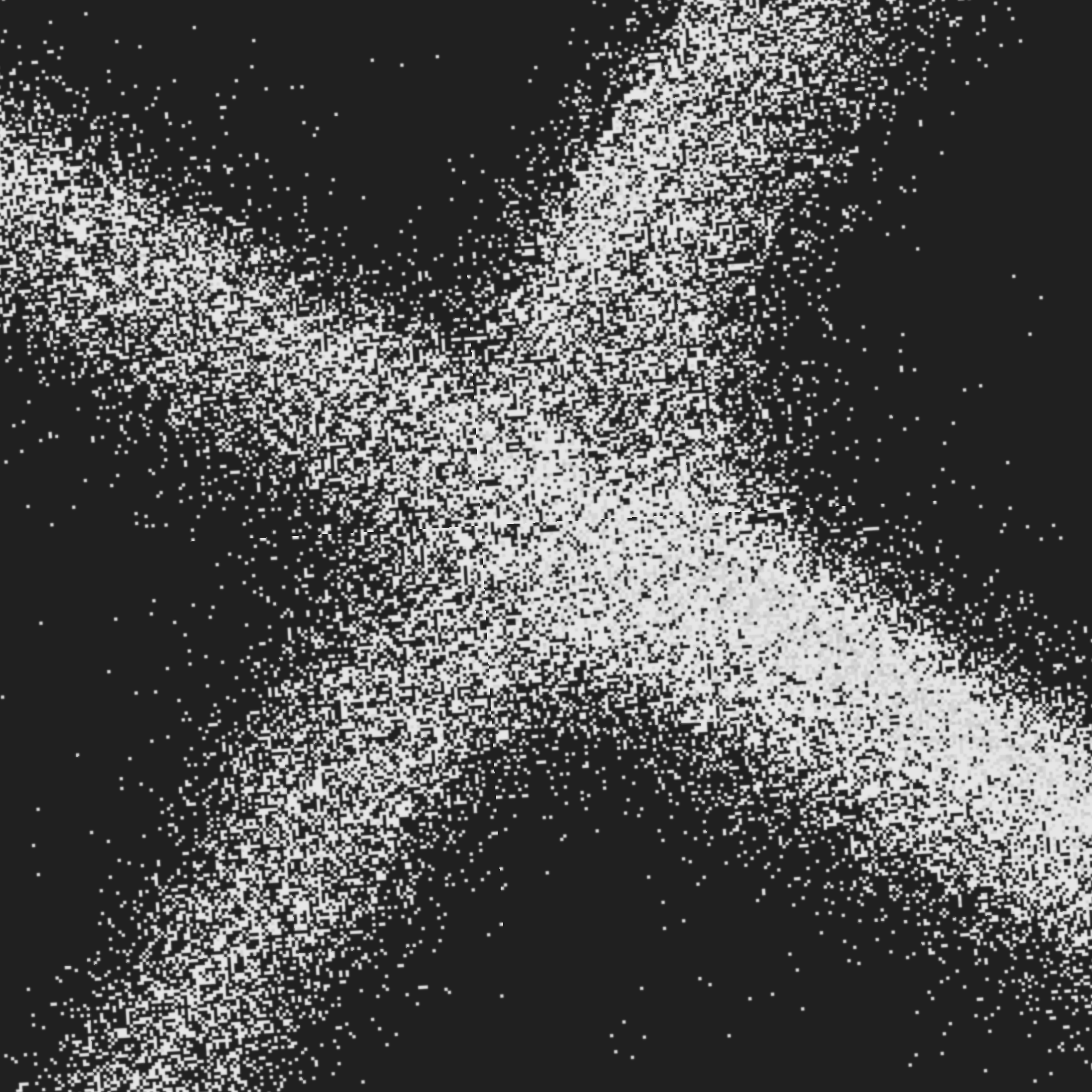}
		\caption{}
		\label{fig::example_mulimodal_driving}
	\end{subfigure}
	\hfill
	\begin{subfigure}[b]{0.092\textwidth}
		\centering
		\includegraphics[width=\textwidth]{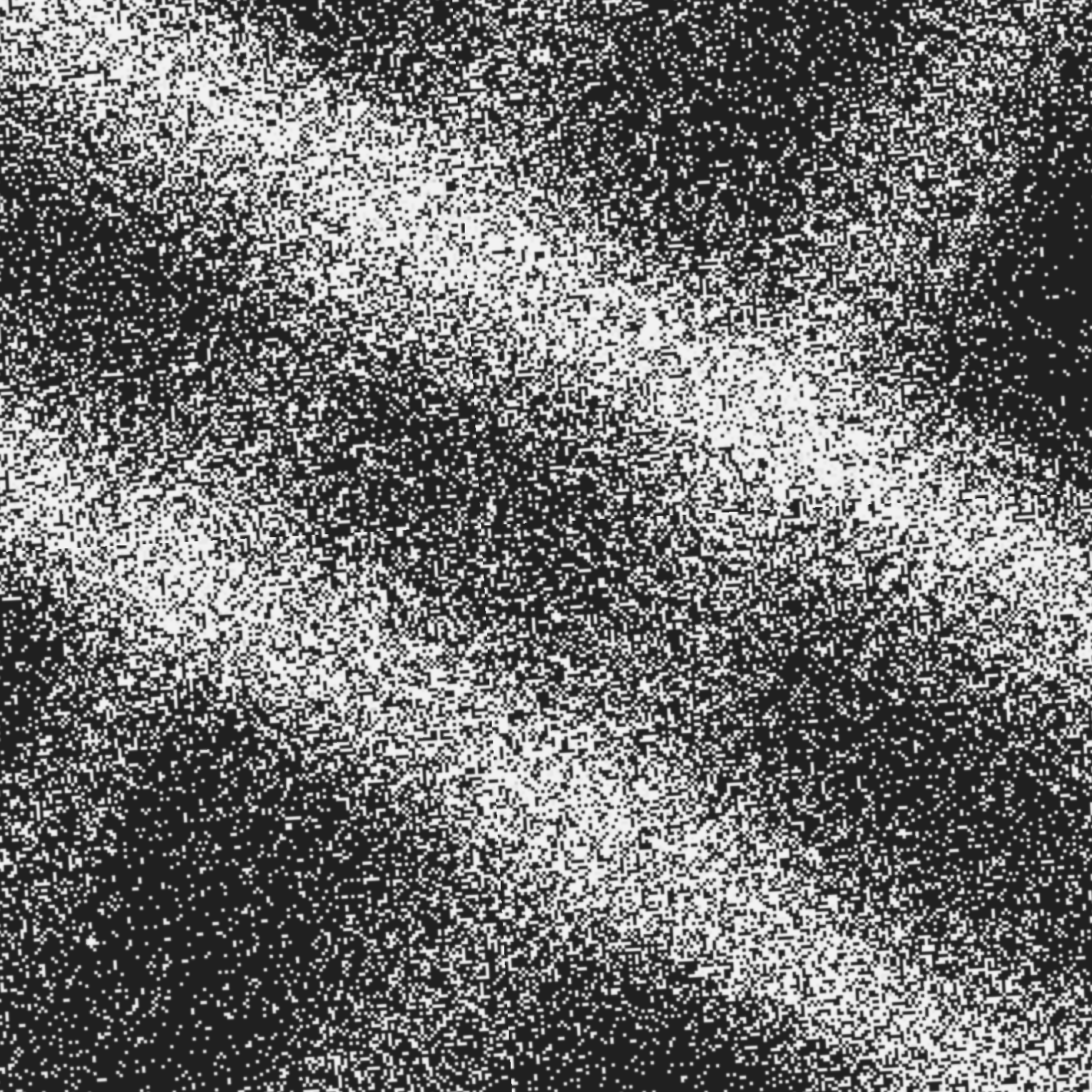}
		\caption{}
		\label{fig::example_mulimodal_walking}
	\end{subfigure}
	\hfill
	\begin{subfigure}[b]{0.092\textwidth}
		\centering
		\includegraphics[width=\textwidth]{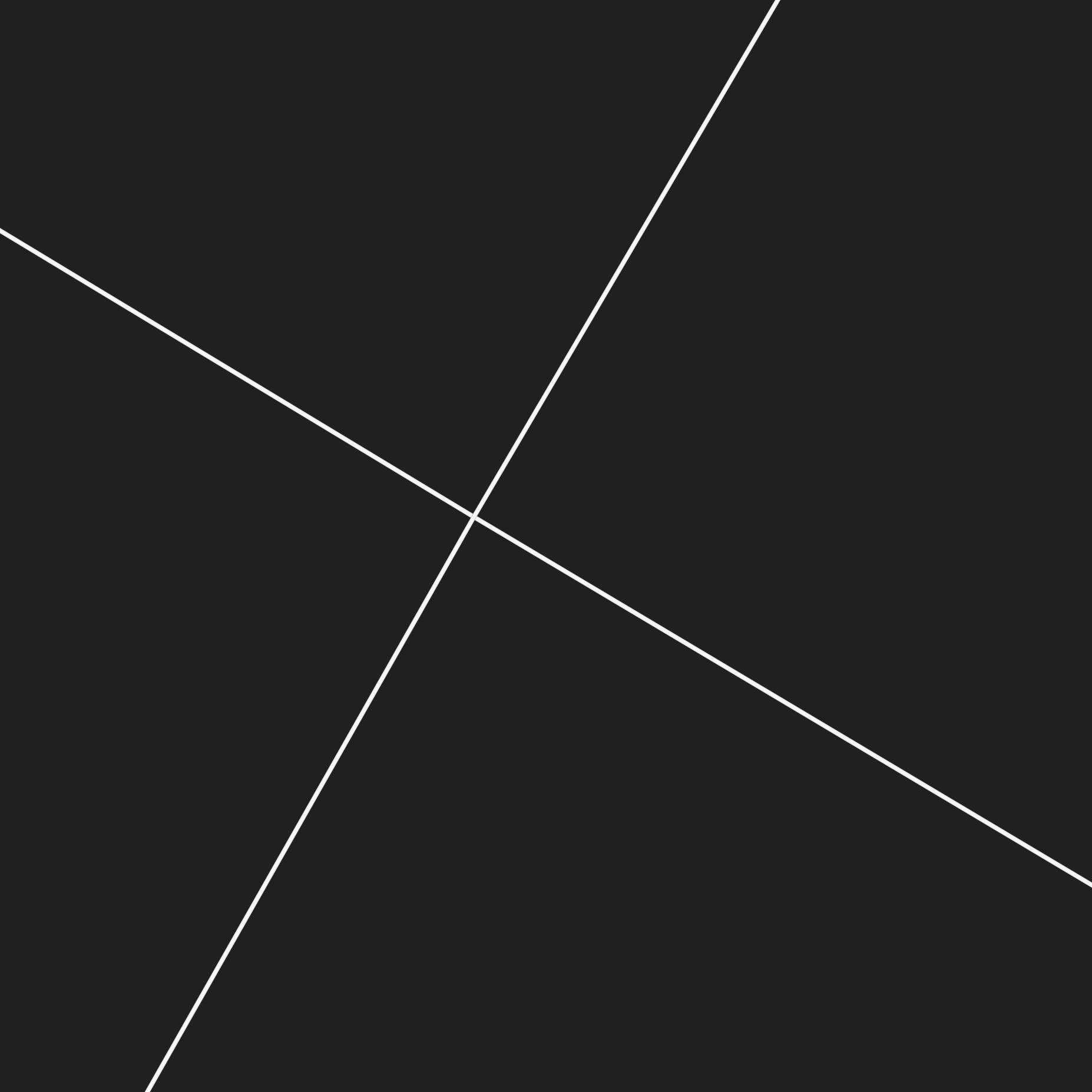}
		\caption{}
		\label{fig::example_mulimodal_network}
	\end{subfigure}
	\hfill
	\begin{subfigure}[b]{0.092\textwidth}
		\centering
		\includegraphics[width=\textwidth]{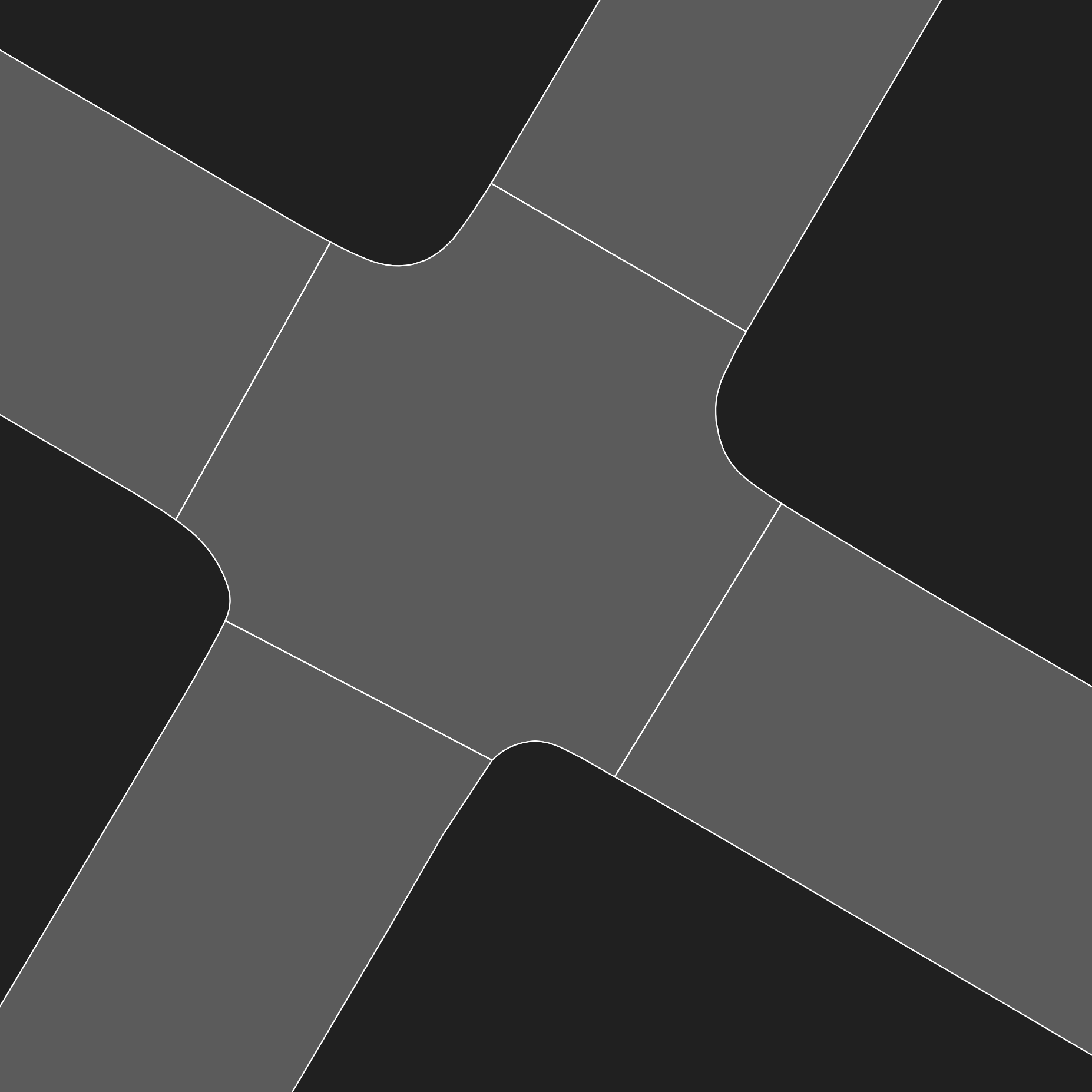}
		\caption{}
		\label{fig::example_mulimodal_vector}
	\end{subfigure}
	\caption{A visual example of the complementary and supplementary nature of multimodal data at a given location: (a) satellite imagery, (b) driving and (c) walking mobility heatmaps, (d) road network, and (e) road casement polygons.}
	\label{fig::example_mulimodal}
\end{figure}

Despite the heterogeneity and syntactic dissimilarities in the natural representations of the different data modalities, they are synchronized since they correspond to the same semantic reality i.e., the events, objects, geographic features, and phenomena associated with a given geographic location. When the descriptions overlap, they serve to corroborate each other. When they are exclusive, they serve to overcome each others' information deficit. \cref{fig::example_mulimodal} shows a visual example of the complementary and supplementary information provided by various geospatial data modalities at a road junction: pedestrian crosswalks in the imagery are corroborated by walking mobility, road centerlines are corroborated by imagery and driving mobility, and road casement polygon widths are corroborated by imagery and driving mobility. While multimodal data presents a challenge due to the possibility of (multiscale) defects (simultaneously) occurring in multiple modalities, there is an opportunity to build multimodal anomaly detection models that are more performant and holistic compared to unimodal models by meaningfully choosing a combination of data modalities and exploiting the synchronization between them.

Geospatial datasets can broadly be classified into: (i) \textit{acquired} data resulting from active or passive sensing of the physical environment (e.g., satellite imagery, mobility data), and (ii) \textit{derived} data which are mathematical or cartographic representations of one or combination of elements in raw or processed acquired data (e.g., road network, vector geometries, demographics). Polygonal vector geometries are an important derived modality in geospatial modeling which are used to represent a multitude of geospatial features such as road casements, buildings, areas of interest (AoIs), landcover, administrative regions, etc. 

In this work, we propose \textit{SeMAnD}, a multimodal and self-supervised anomaly detection strategy that exploits the semantic correspondence between different synchronized geospatial data modalities to detect anomalies in geospatial vector geometries by implicitly using a combination of acquired and derived modalities as reference. Augmentations to generate the self-supervision signal are performed on non-reference modalities. SeMAnD comprises of (i) a simple data augmentation strategy, called \textit{RandPolyAugment}, (\cref{subsection:randpolyaugment}) that generates augmentations of vector geometries, and (ii) a self-supervised training objective (\cref{ssl_with_rpa}) with three components that incentivize learning representations of multimodal data that are discriminative to local changes in one modality which are not corroborated by the other modalities. To demonstrate SeMAnD in this paper, satellite imagery, GPS-based mobility data, and road network graph are considered reference modalities for detecting errors in road casement polygons (RCP). However, SeMAnD can be easily extended by adding more/different reference modalities to detect errors in other/additional types of polygonal vector geometries. RCPs are geographically accurate representations of road shape and have been used for transport planning \cite{oflaherty2018, aus_road_casement}, cadastral mapping \cite{insprire_directive}, and traffic engineering \cite{esriuk_pavements_2020}.  

\begin{figure}
    \centering
    \begin{subfigure}{0.075\textwidth}
        \centering
        \includegraphics[width=\textwidth]{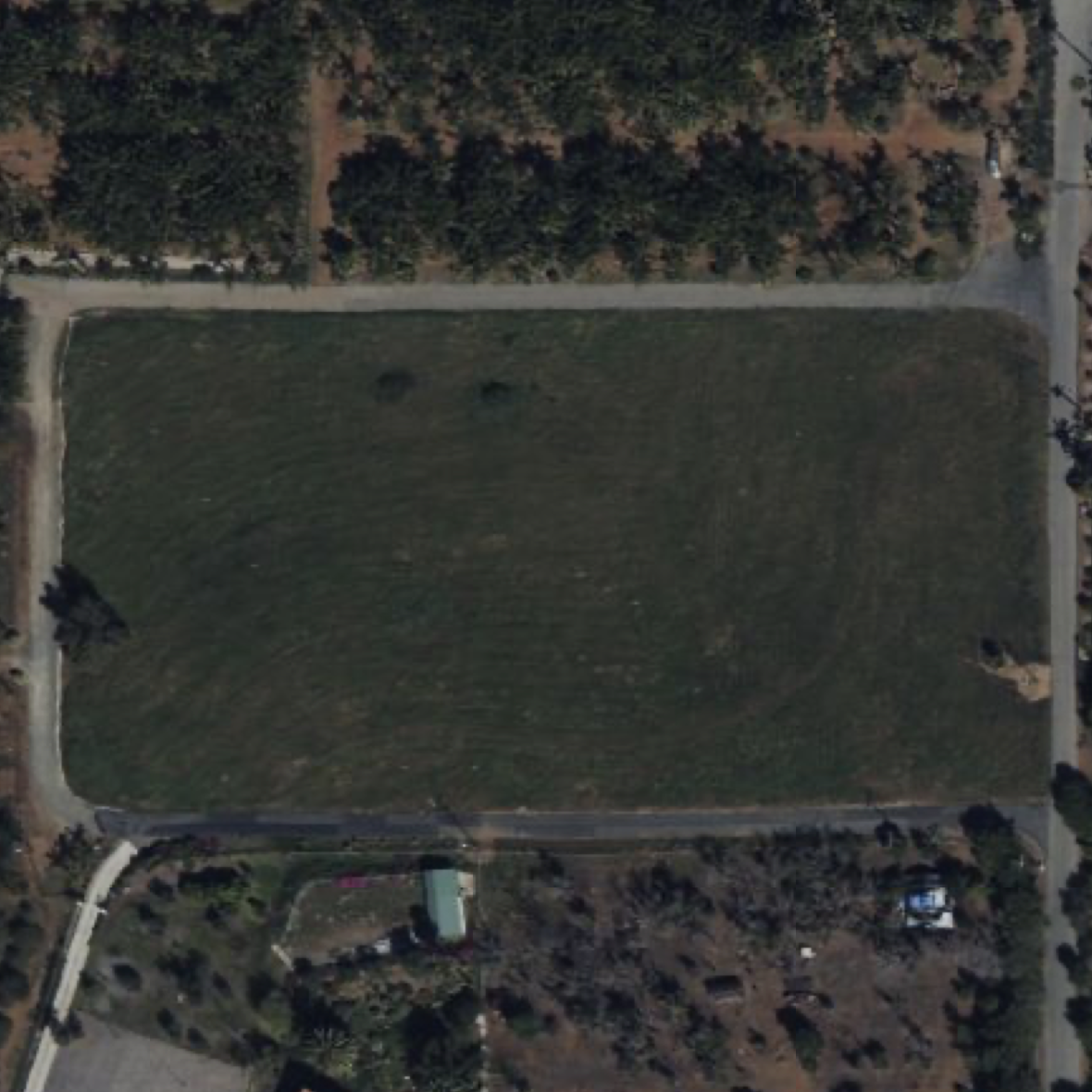}
     
        \includegraphics[width=\textwidth]{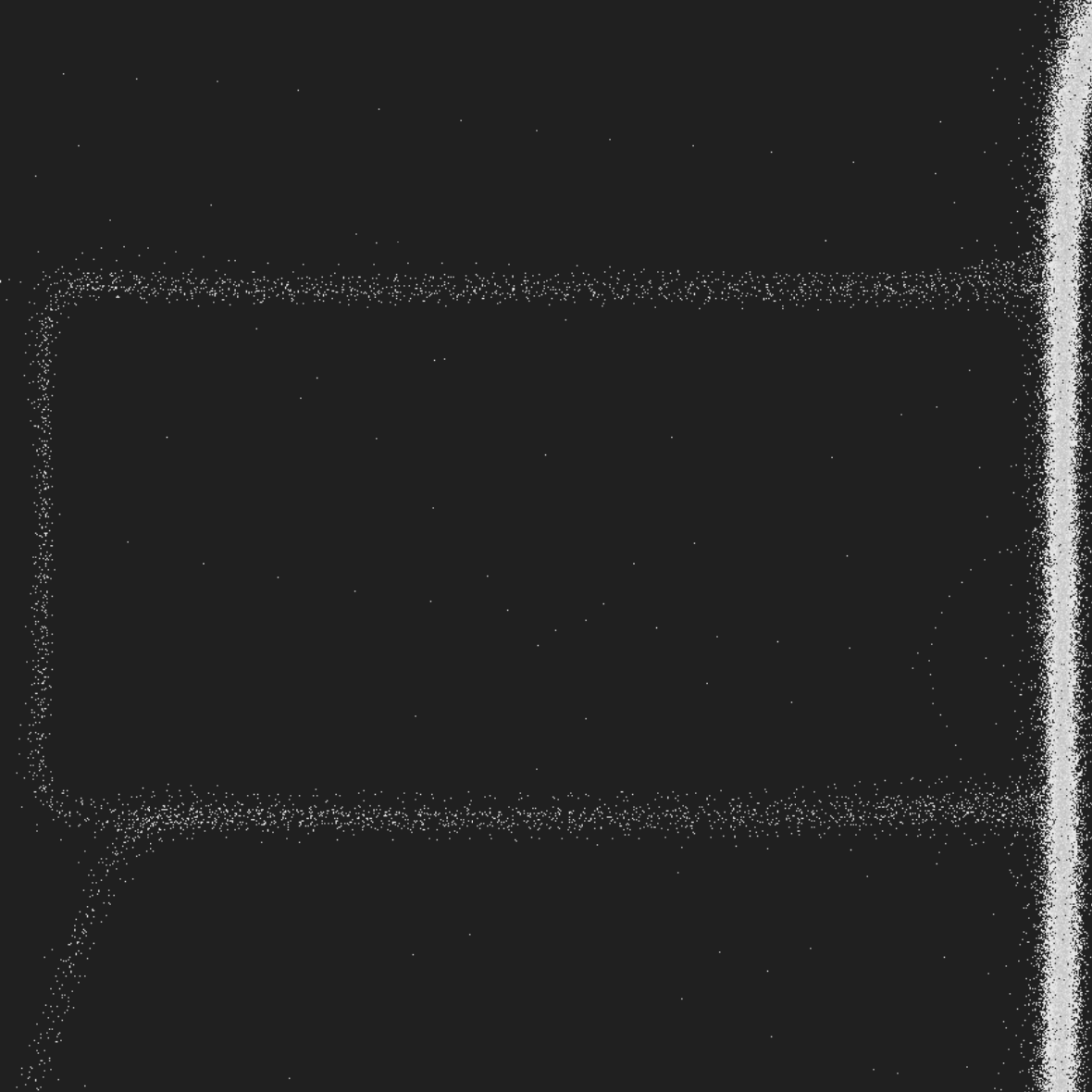}
        
        \includegraphics[width=\textwidth]{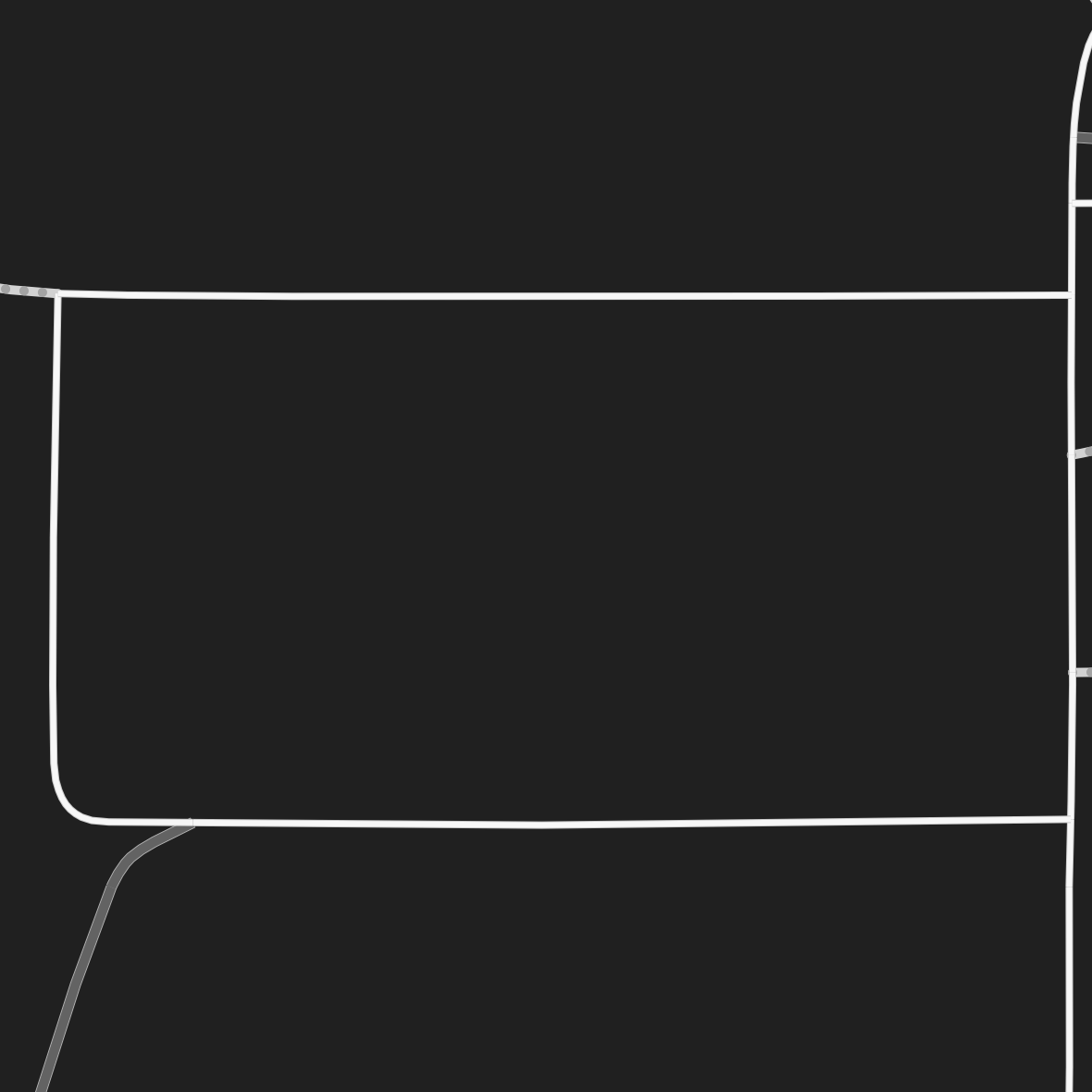}
        
        \includegraphics[width=\textwidth]{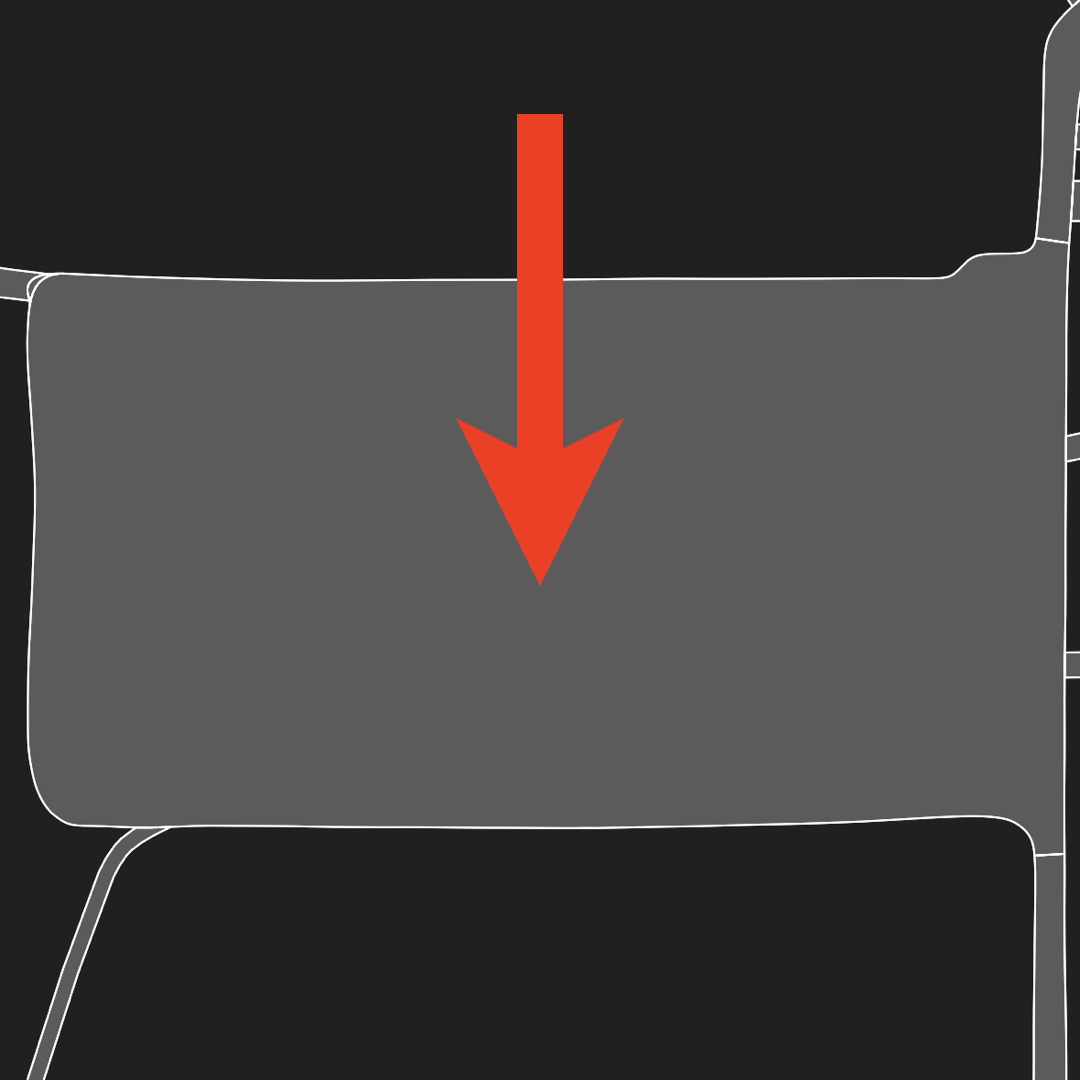}
        \caption{}
        \label{anomaly_1}
    \end{subfigure}
    \begin{subfigure}{0.075\textwidth}
        \centering
        \includegraphics[width=\textwidth]{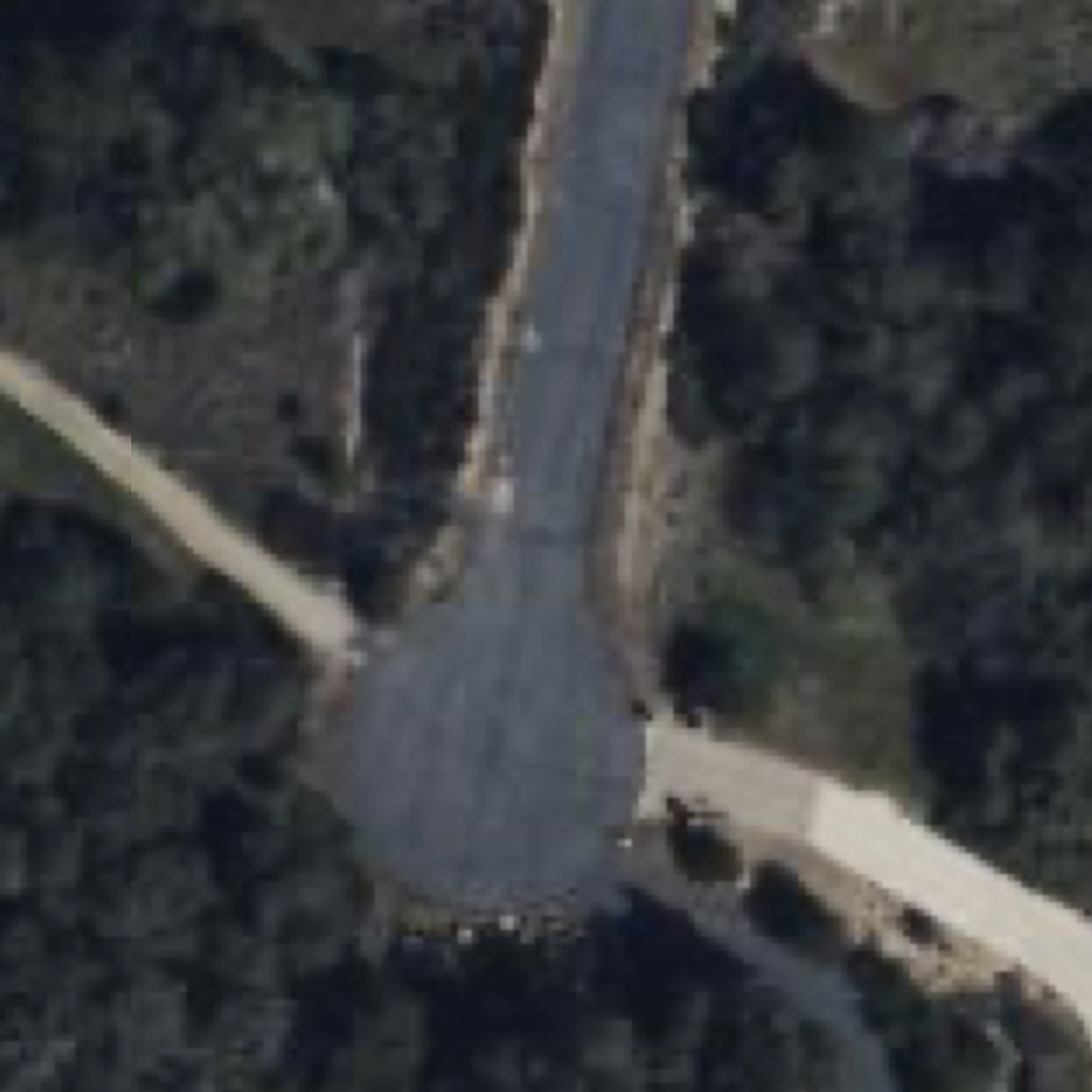}
        
        \includegraphics[width=\textwidth]{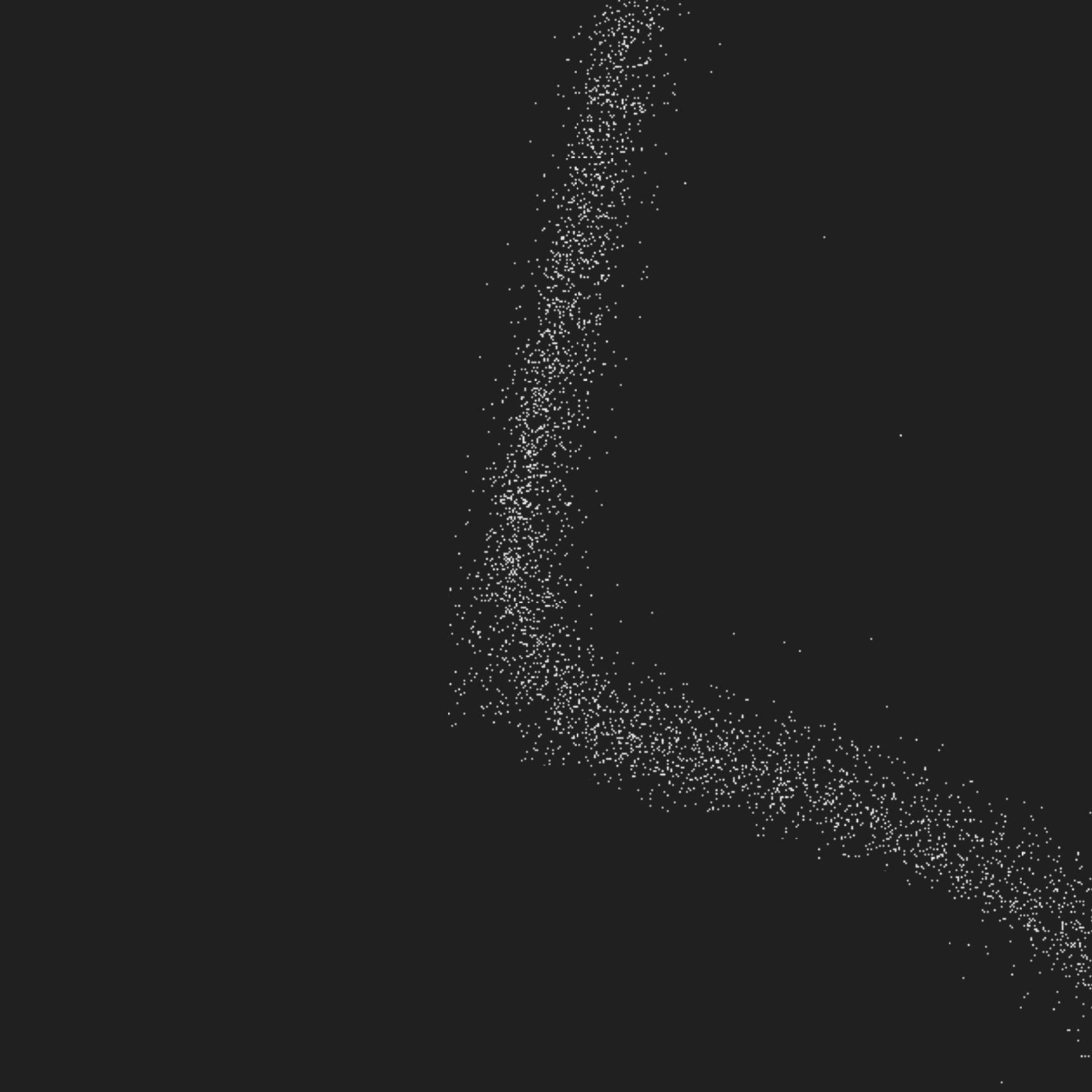}
        
        \includegraphics[width=\textwidth]{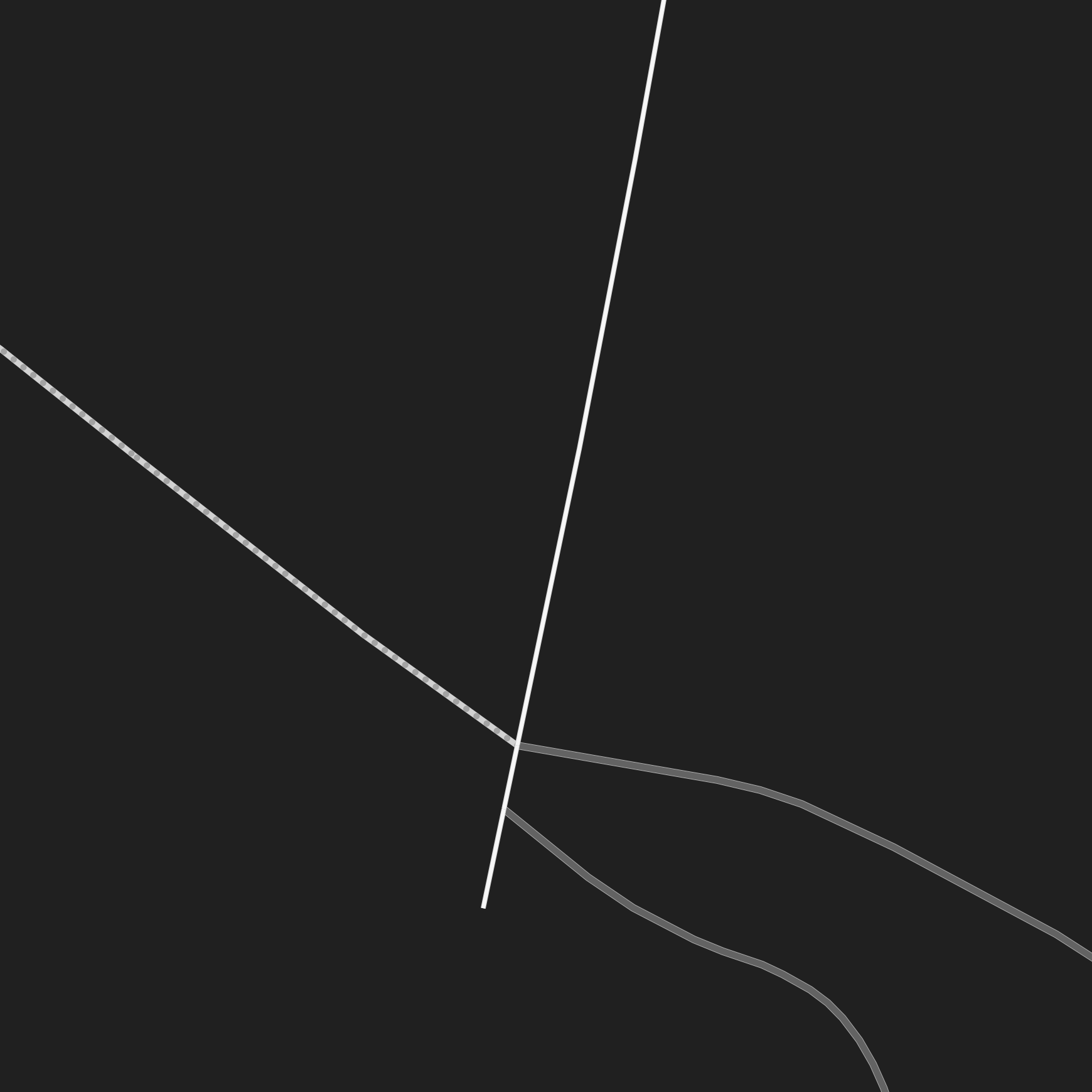}
        
        \includegraphics[width=\textwidth]{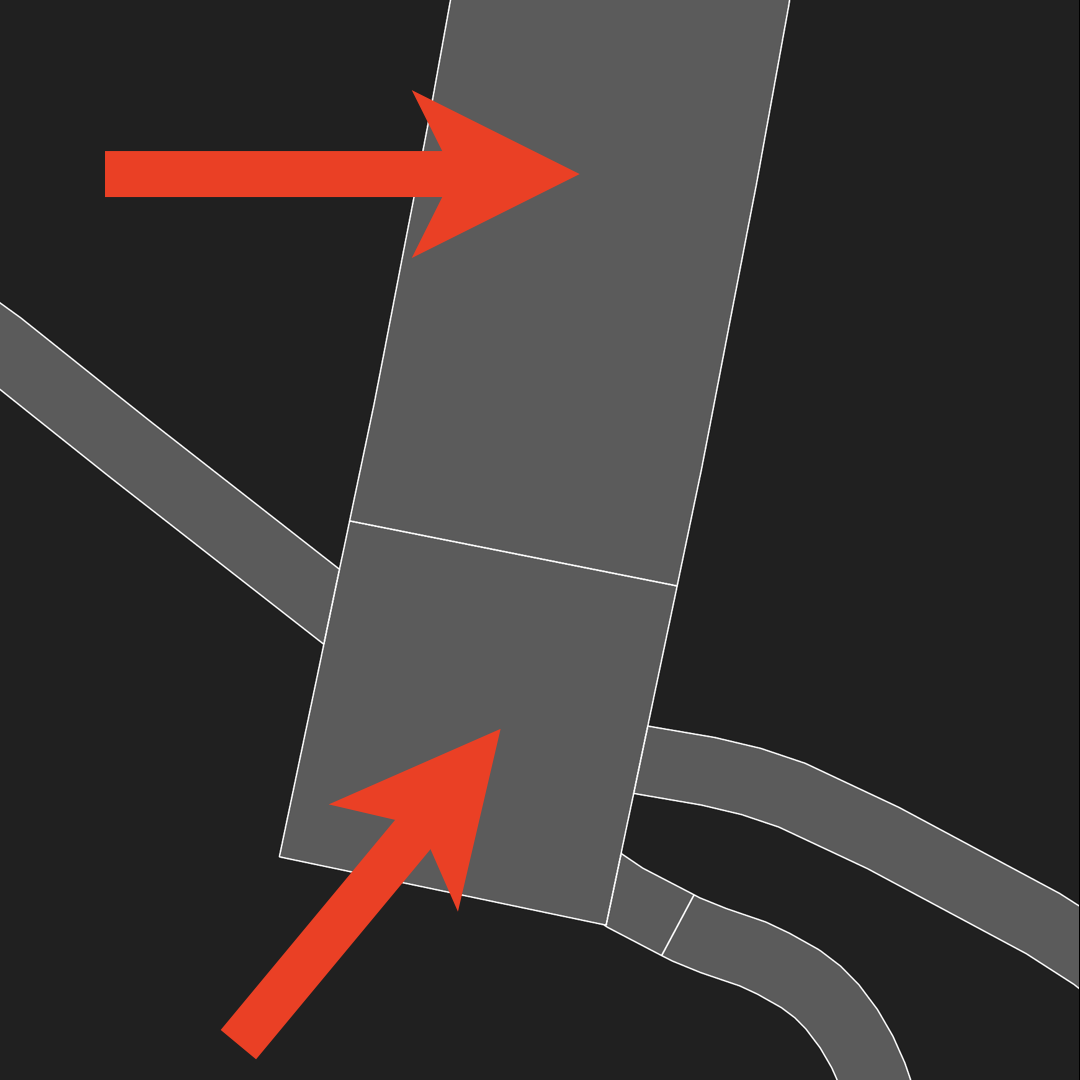}
        \caption{}
        \label{anomaly_2}
    \end{subfigure}
    \begin{subfigure}{0.075\textwidth}
        \centering
        \includegraphics[width=\textwidth]{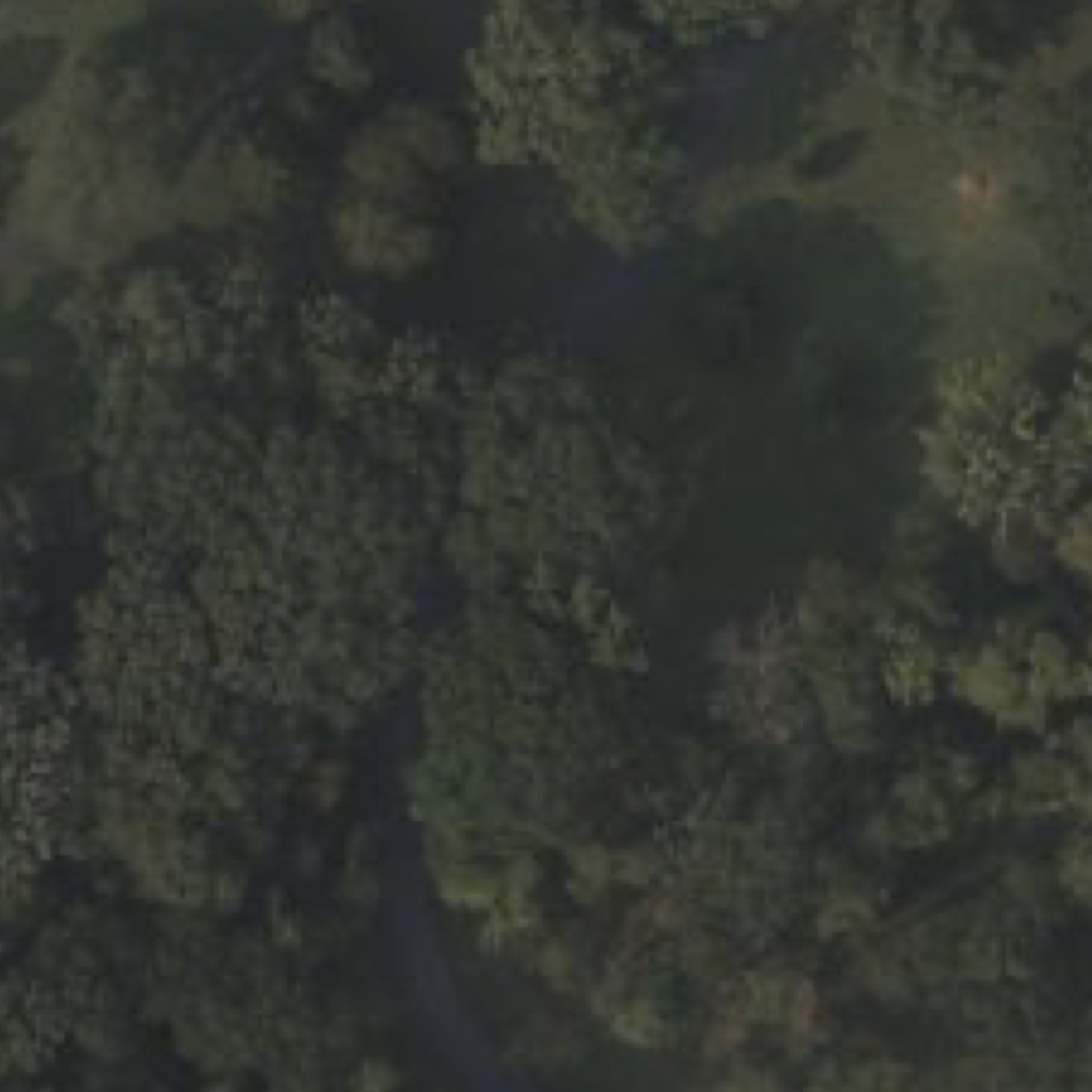}
        
        \includegraphics[width=\textwidth]{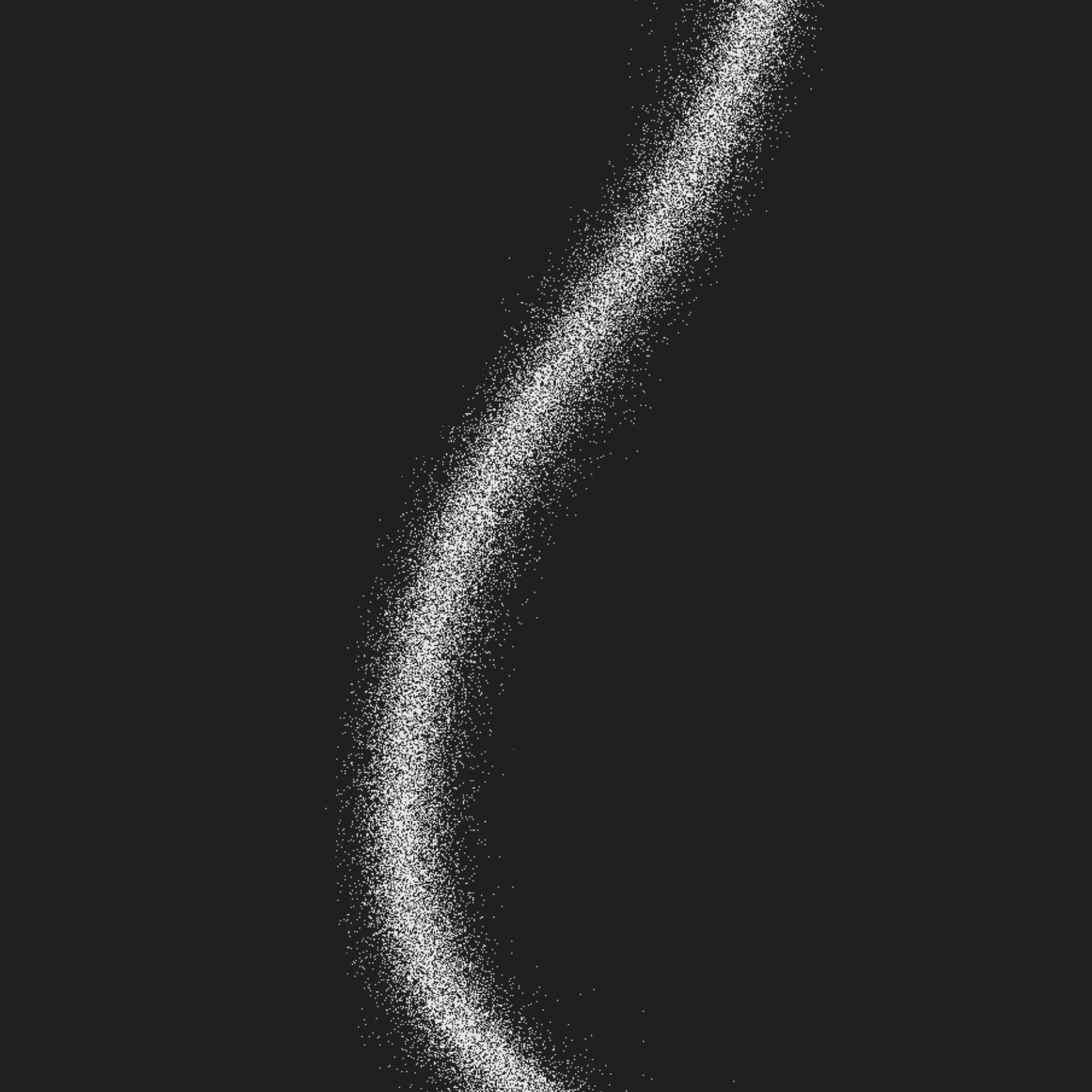}
        
        \includegraphics[width=\textwidth]{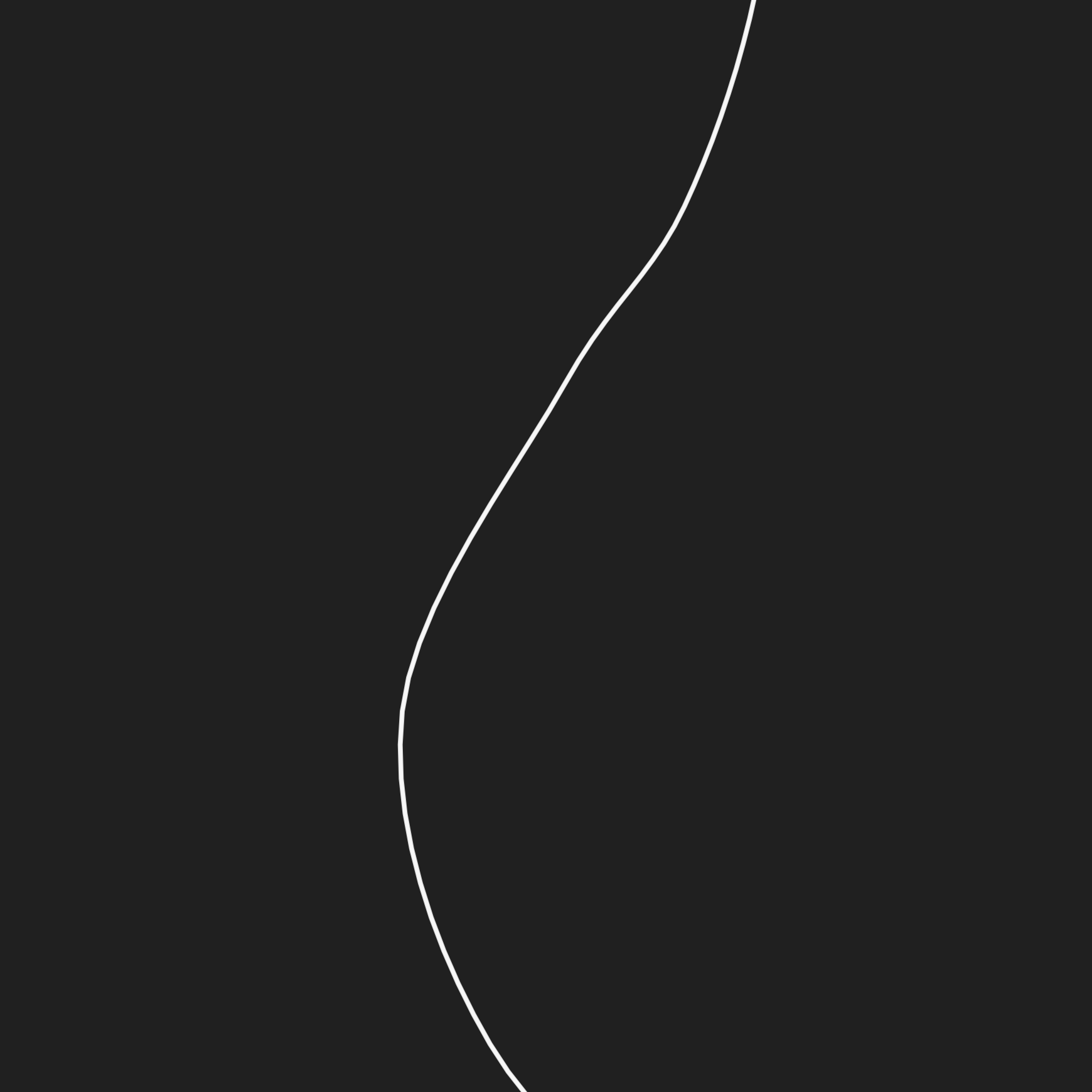}
        
        \includegraphics[width=\textwidth]{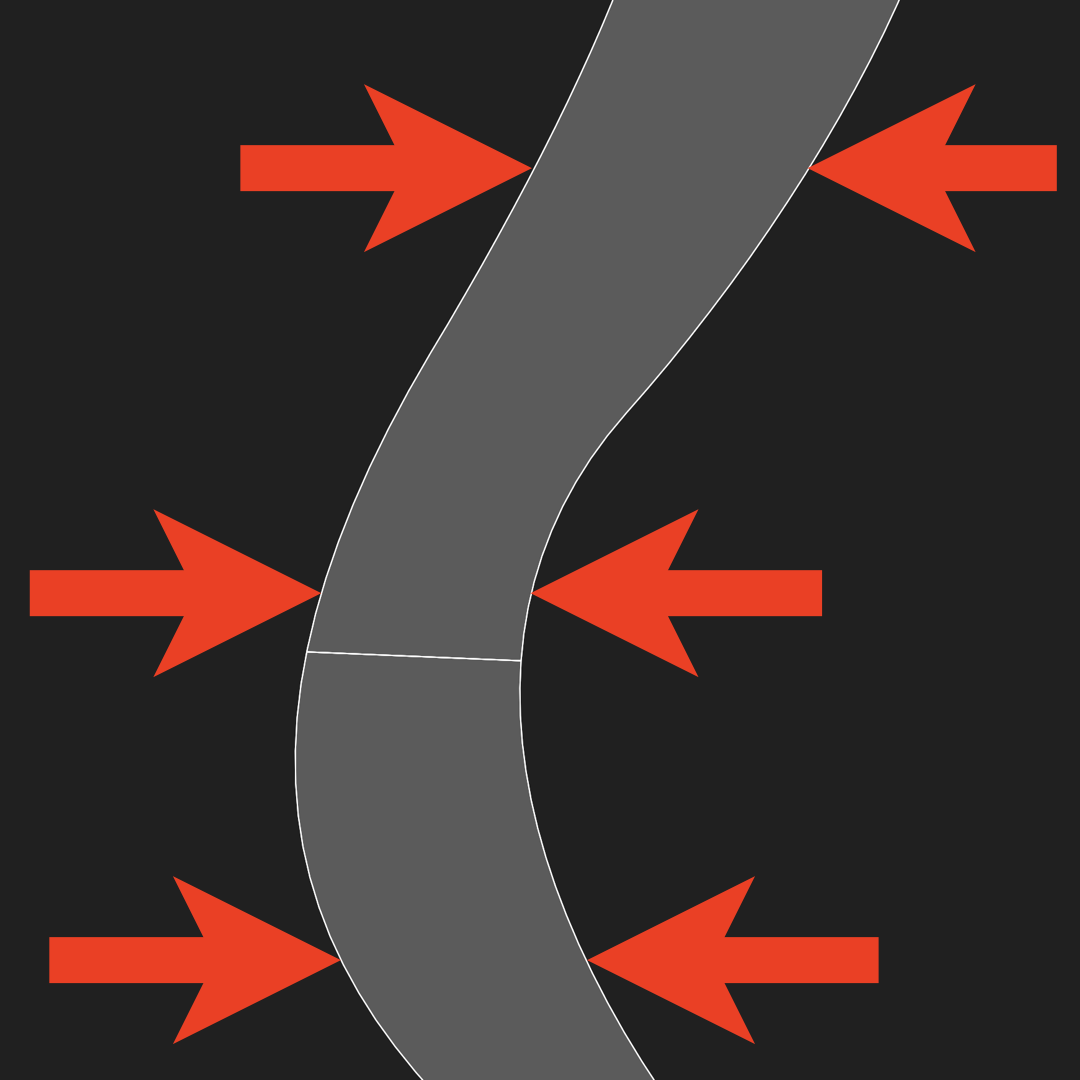}
        \caption{}
        \label{anomaly_3}
    \end{subfigure}
    \begin{subfigure}{0.075\textwidth}
        \centering
        \includegraphics[width=\textwidth]{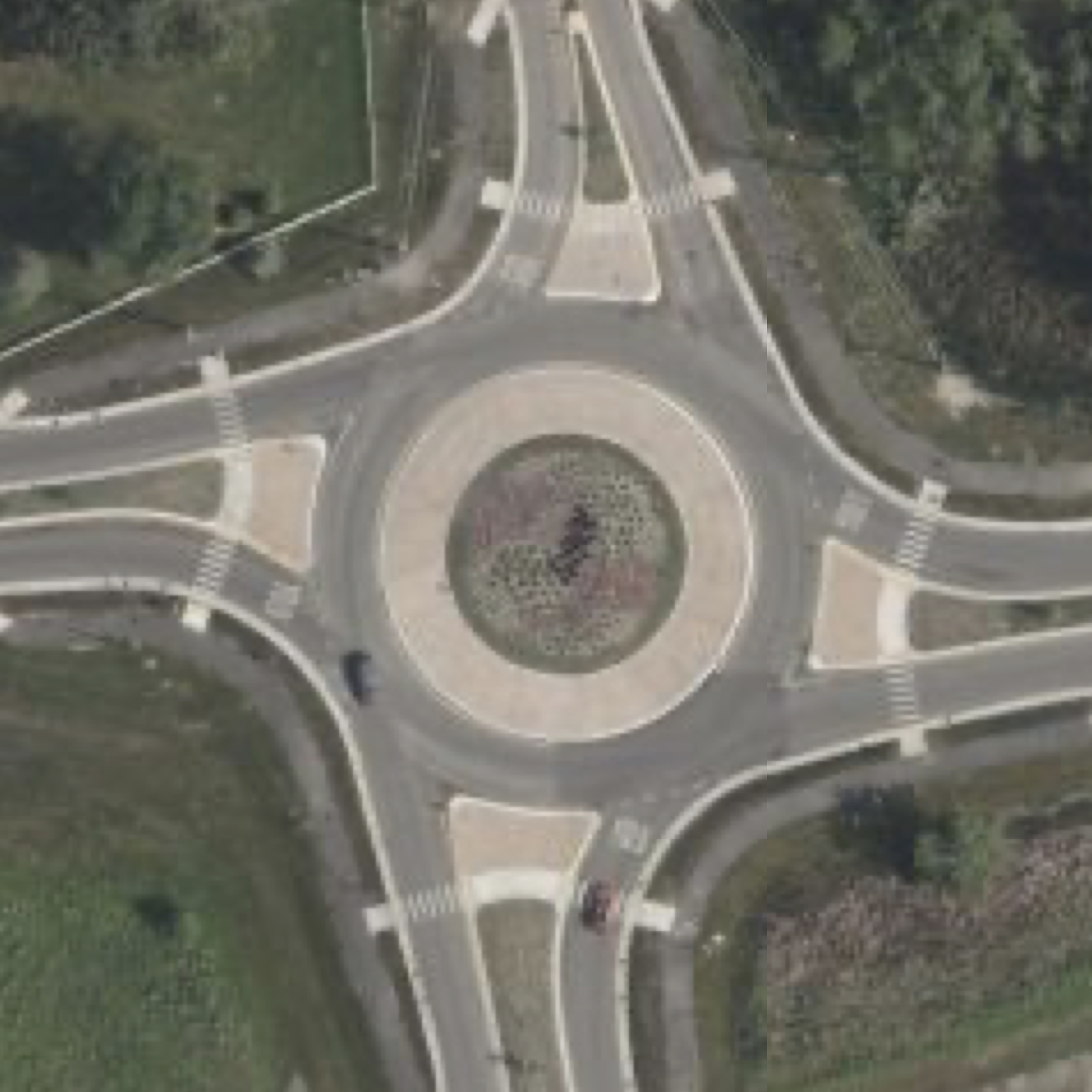}
        
        \includegraphics[width=\textwidth]{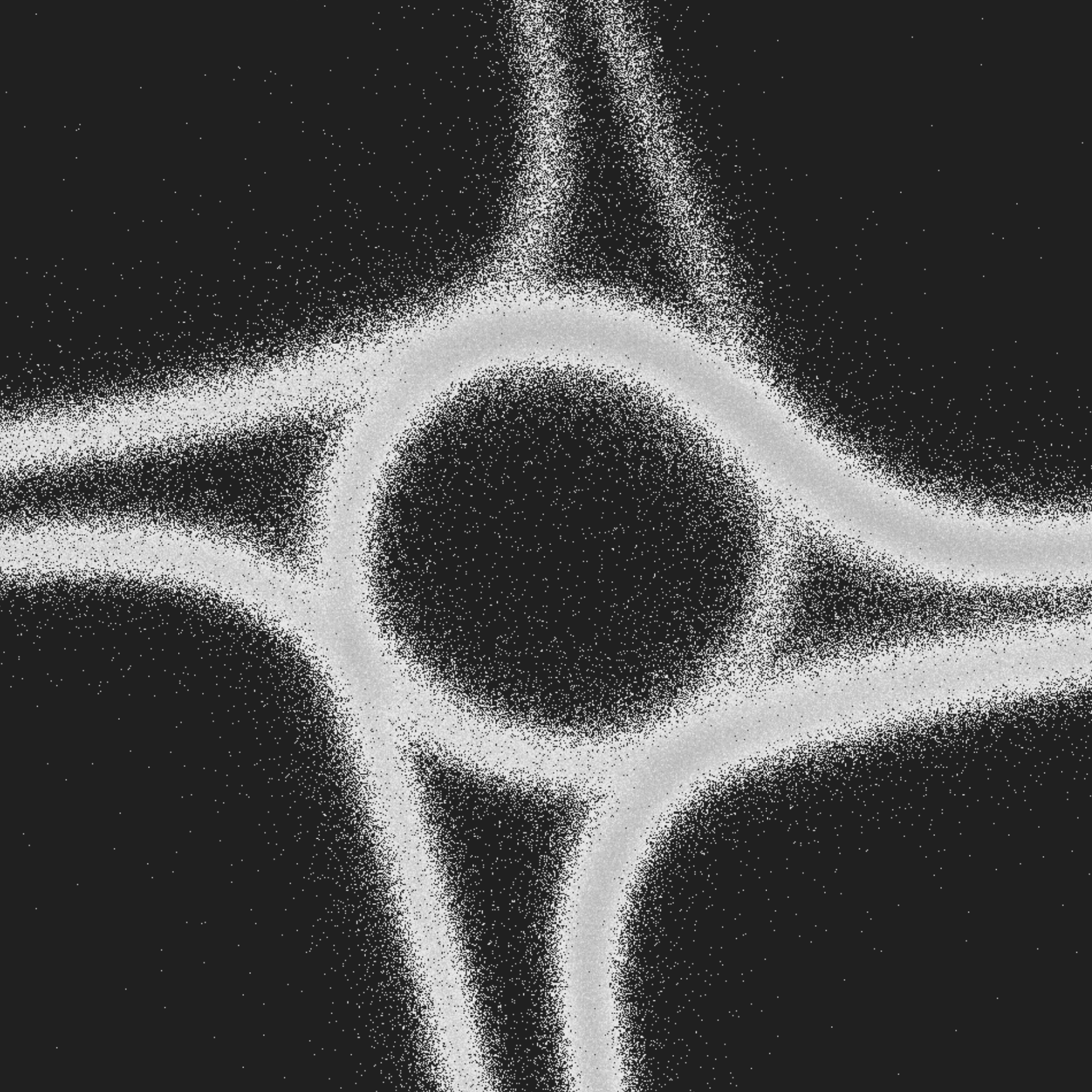}
        
        \includegraphics[width=\textwidth]{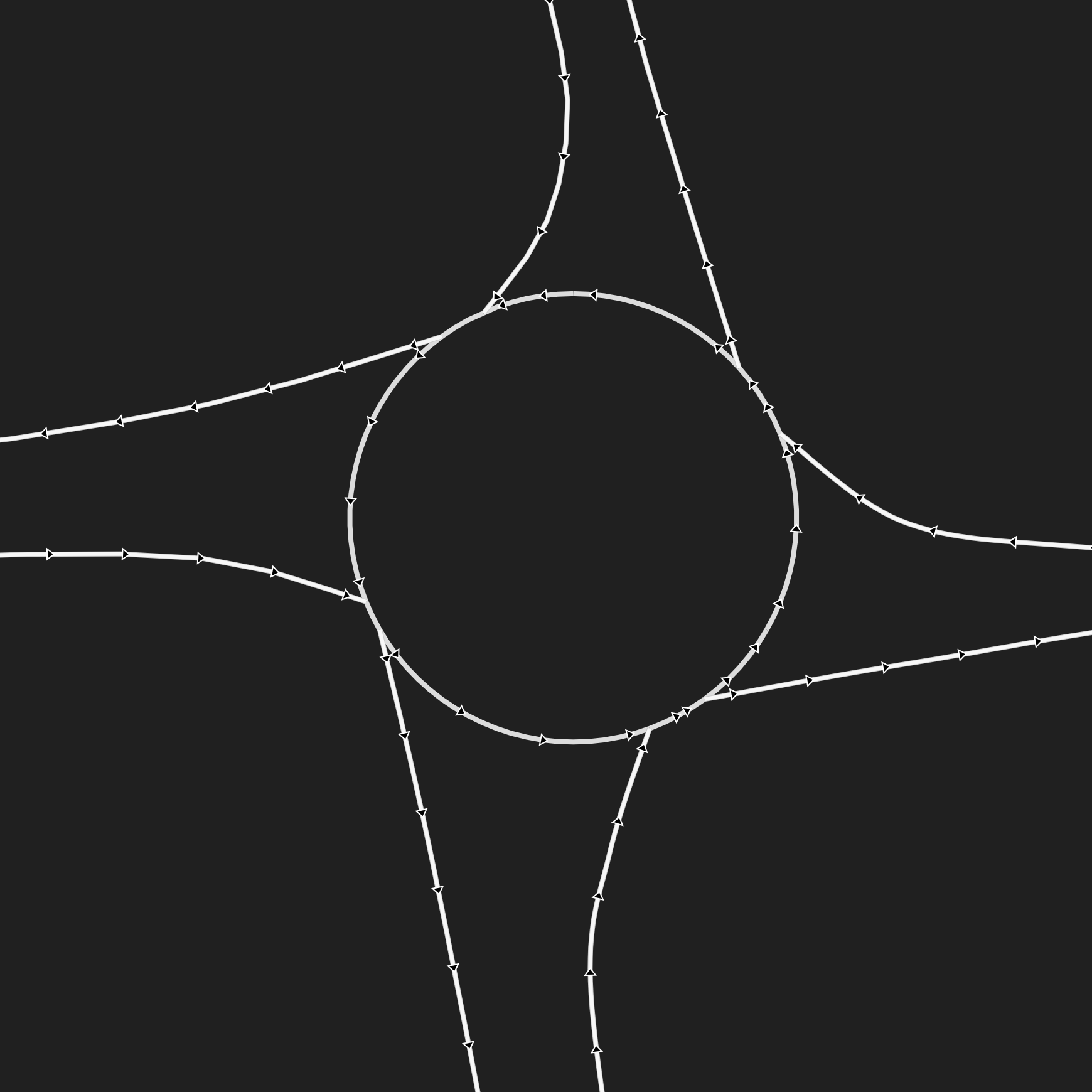}
        
        \includegraphics[width=\textwidth]{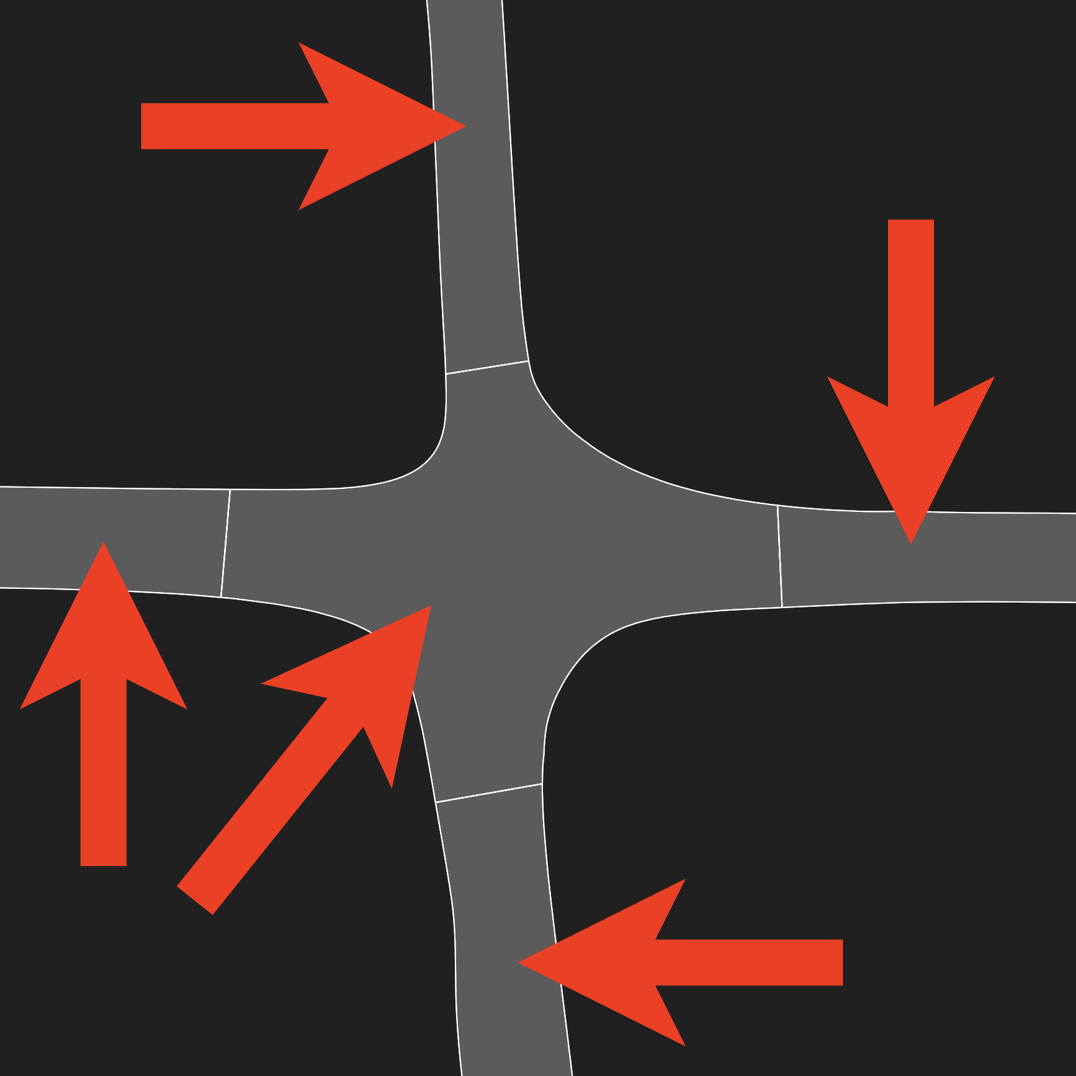}
        \caption{}
        \label{anomaly_4}
    \end{subfigure}
    \begin{subfigure}{0.075\textwidth}
        \centering
        \includegraphics[width=\textwidth]{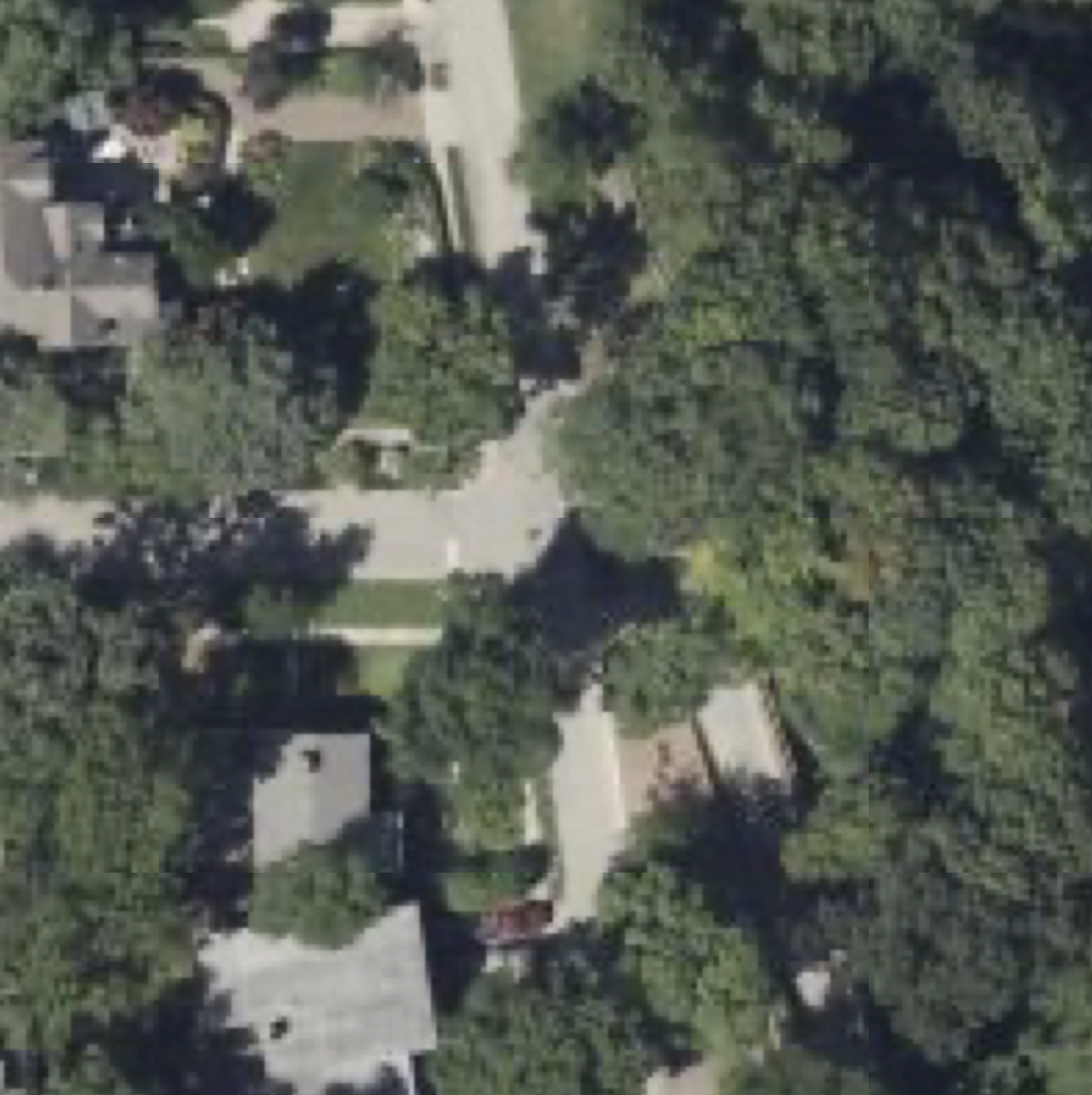}
        
        \includegraphics[width=\textwidth]{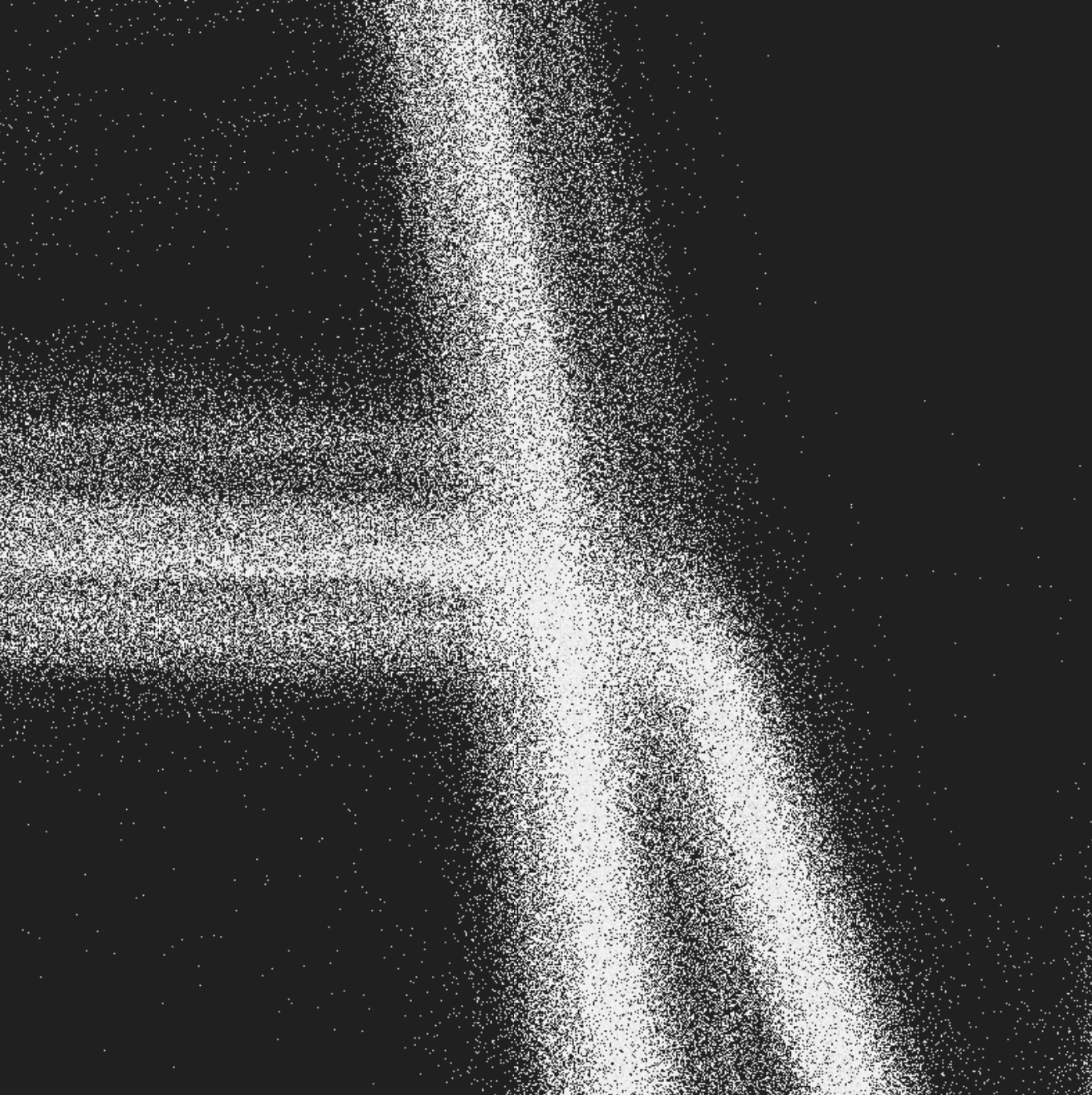}
        
        \includegraphics[width=\textwidth]{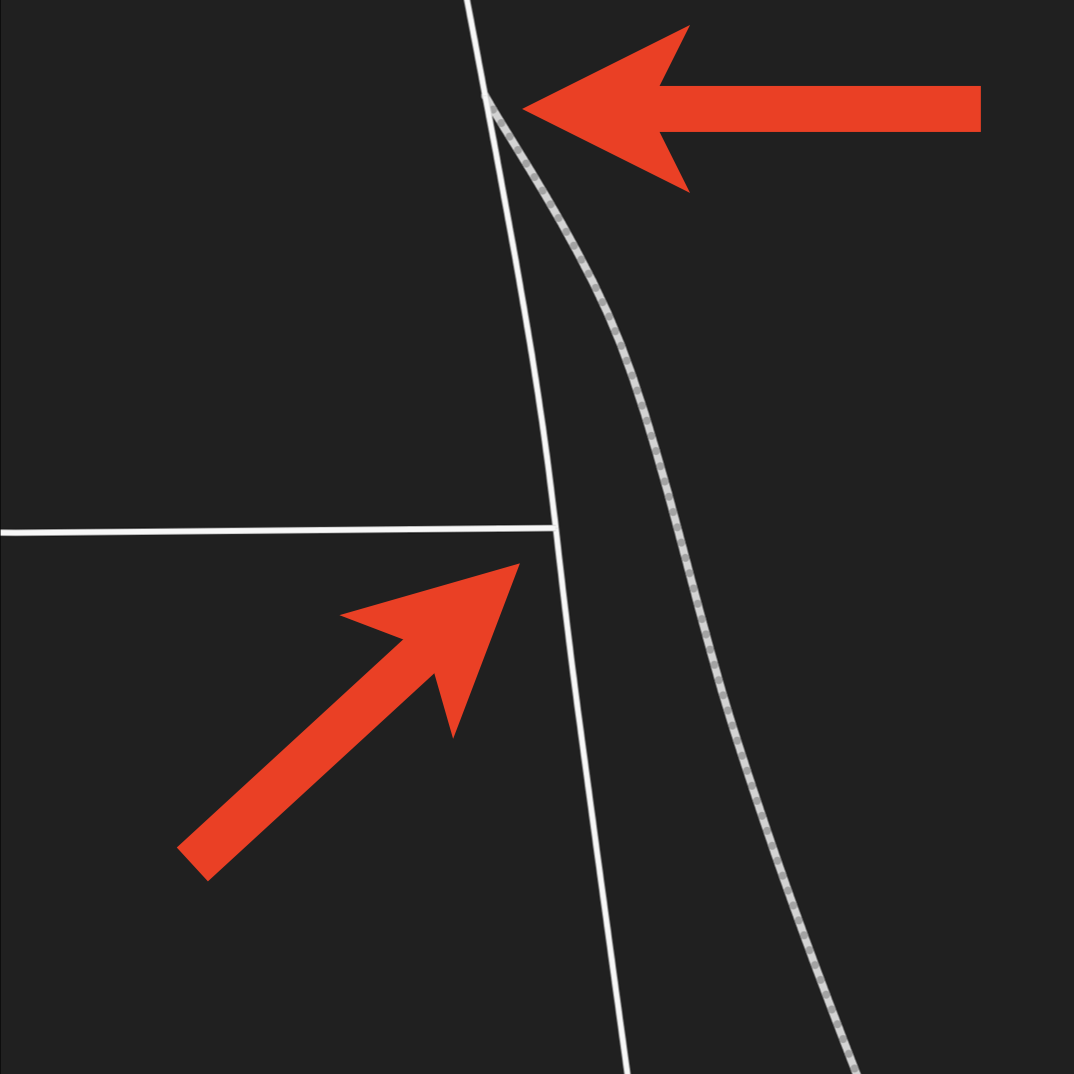}
        
        \includegraphics[width=\textwidth]{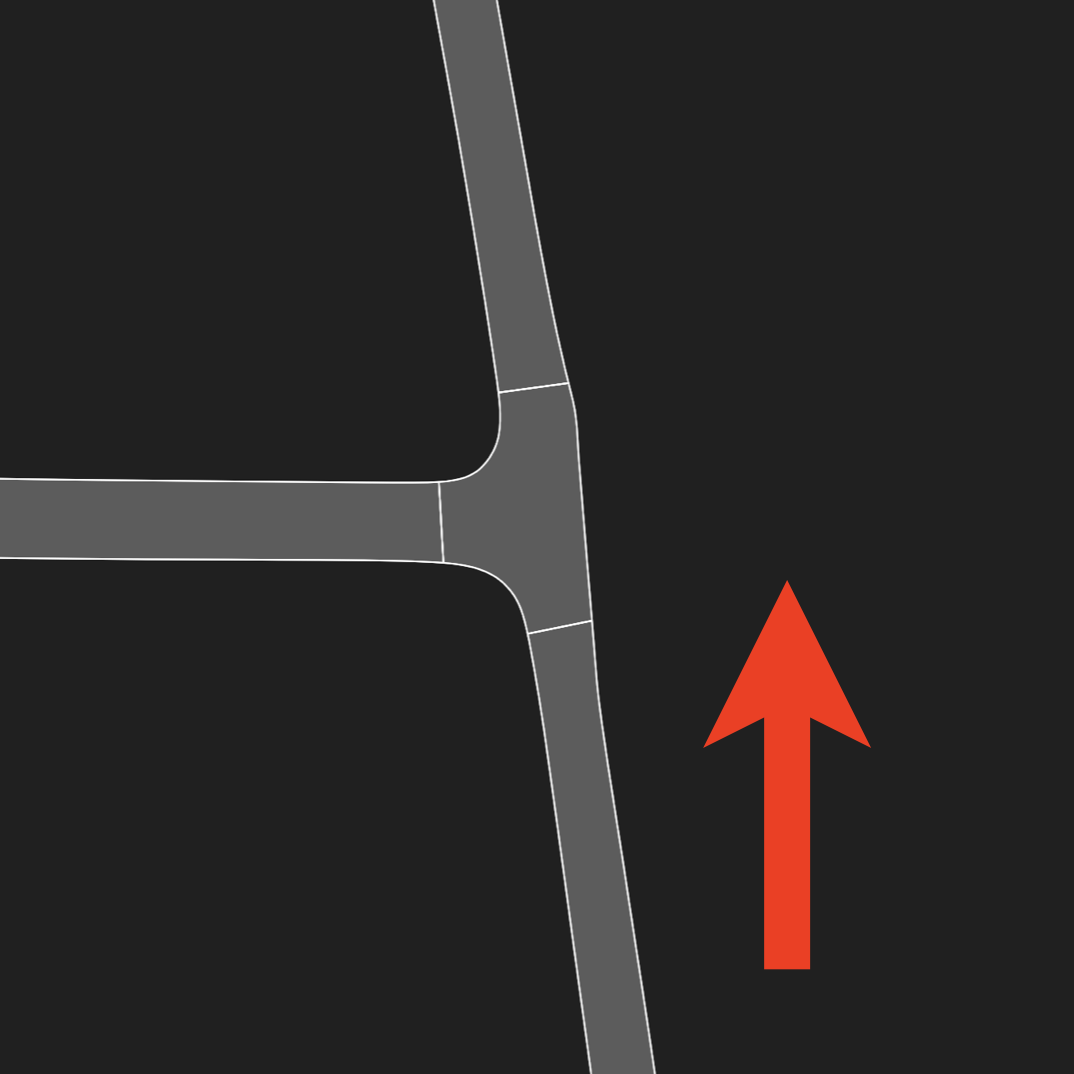}
        \caption{}
        \label{anomaly_5}
    \end{subfigure}
    \begin{subfigure}{0.075\textwidth}
        \centering
        \includegraphics[width=\textwidth]{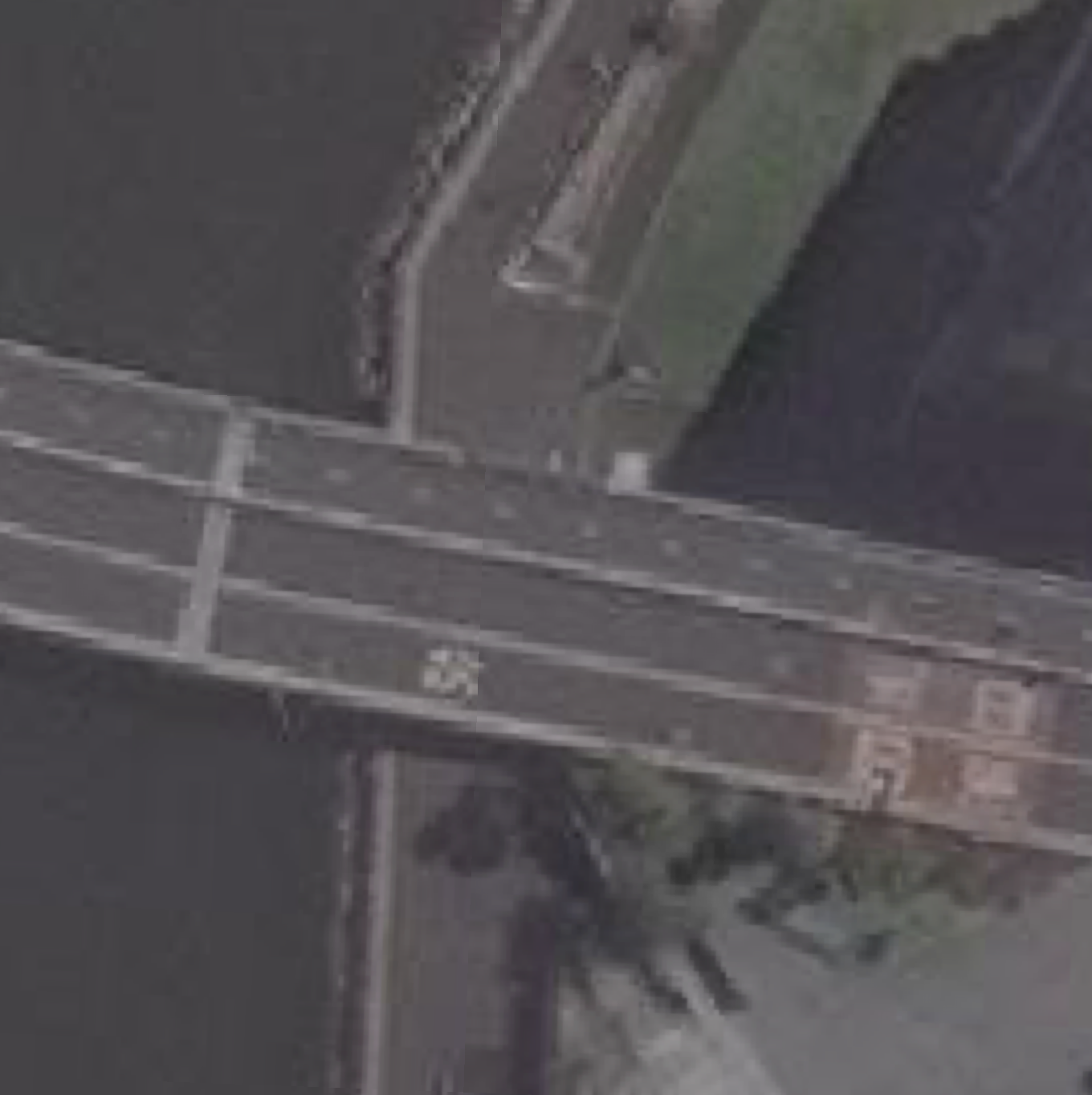}
        
        \includegraphics[width=\textwidth]{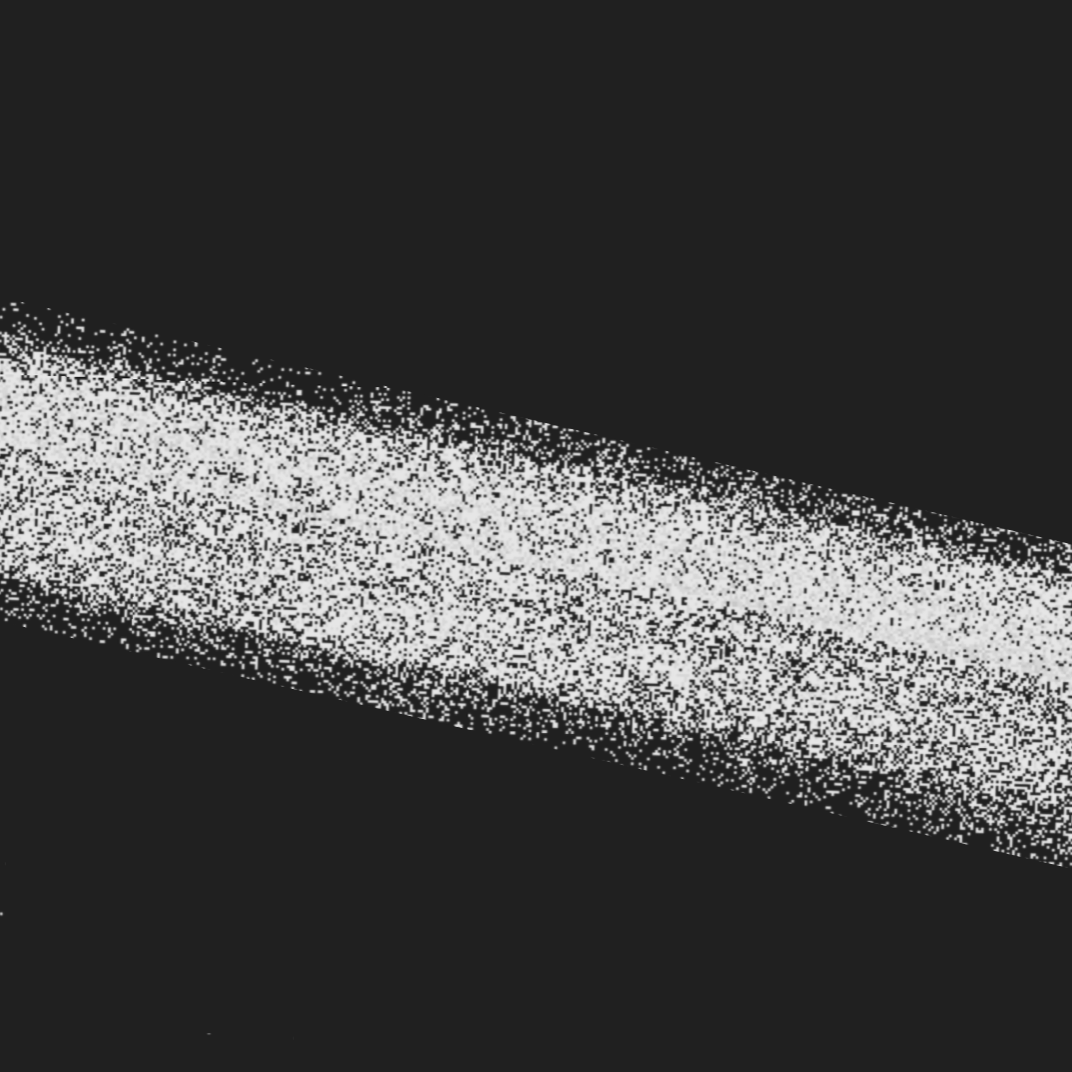}
        
        \includegraphics[width=\textwidth]{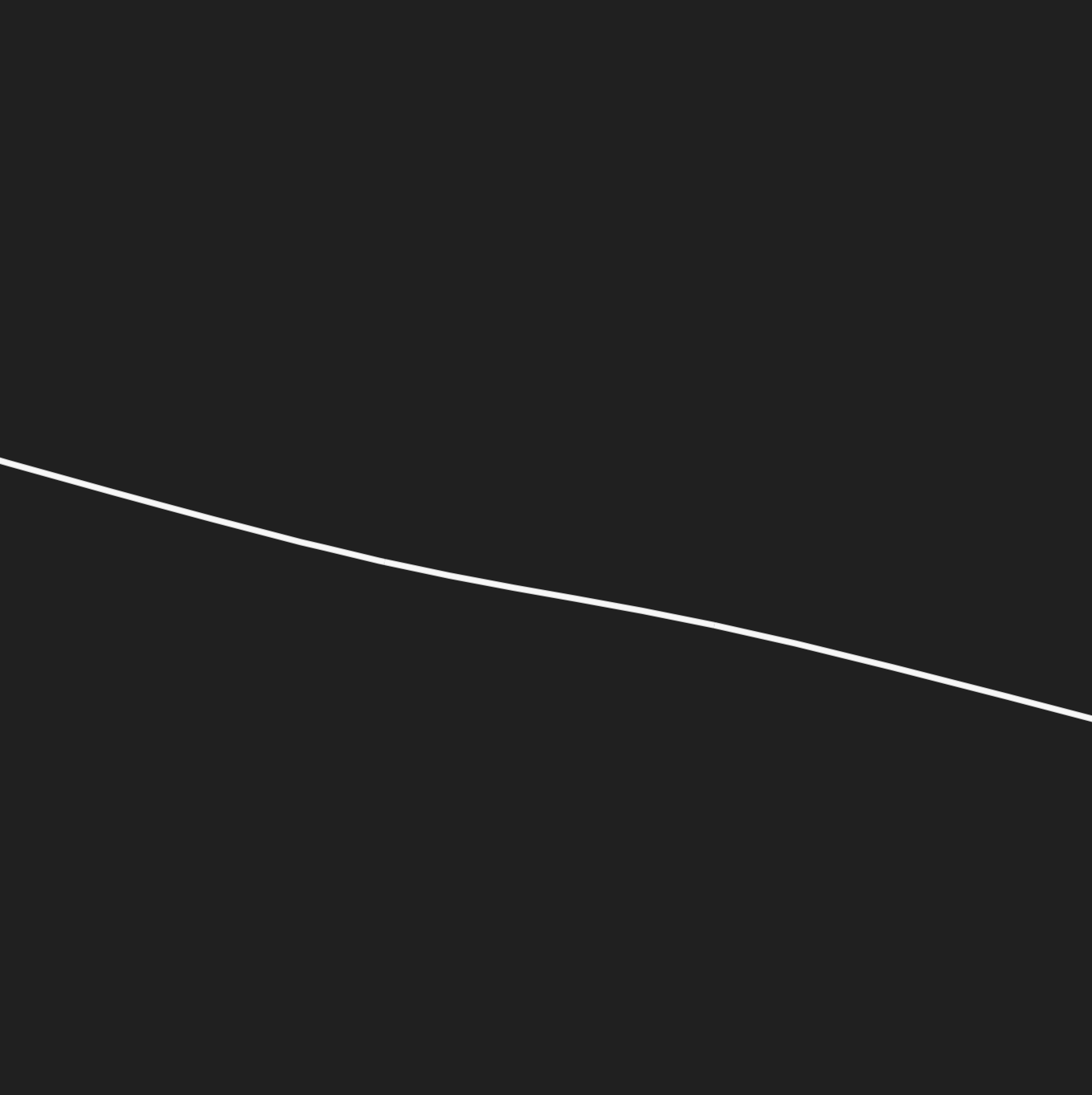}
        
        \includegraphics[width=\textwidth]{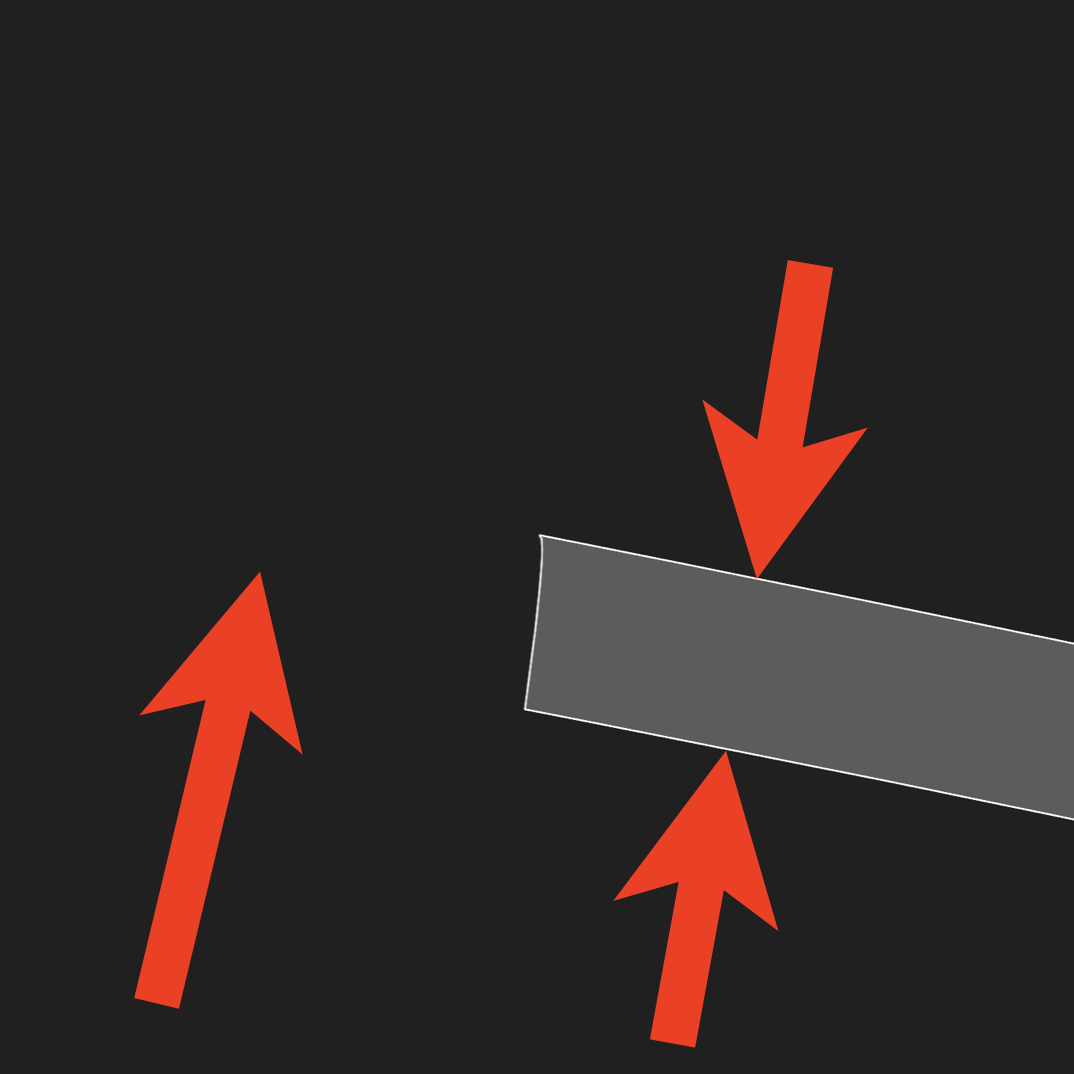}
        \caption{}
        \label{anomaly_6}
    \end{subfigure}
    \caption{Example geospatial anomalies (red arrows) explained in \cref{section:introduction} detected by SeMAnD. Rows from the top: satellite imagery, mobility heatmaps, road network, and road casement polygons.}
    \label{figure:anomalous_multimodal_data}
\end{figure}

\cref{figure:anomalous_multimodal_data} shows examples of real-world anomalies (highlighted with red arrows) detected by the trained (\cref{results}) SeMAnD model. \cref{anomaly_1} shows a playground misclassified by an over-inclusive RCP; \cref{anomaly_2} shows a malformed cul-de-sac and misshapen road; \cref{anomaly_3} shows a wider RCP compared to the width from mobility while satellite imagery is uninformative due to tree canopy; \cref{anomaly_4} shows RCPs at a junction which changed to a roundabout; \cref{anomaly_5} shows RCPs missing on a road and an incorrect road network; \cref{anomaly_6} shows an incomplete and narrower RCP. These examples demonstrate that SeMAnD's strategy for representation, alignment, and fusion of multimodal datasets (\cref{section:repn_algn_fusn}) combined with its training strategy is able to detect real-world changes resulting in data regressions (\cref{anomaly_4}) as well as various inaccuracies in derived data. SeMAnD can also be used to evaluate the accuracy and freshness of datasets. SeMAnD also detects errors in derived data modalities implicitly (non-augmented) used as reference (e.g., errors detected in the road network in \cref{anomaly_5}) which we hypothesize is a result of the inter-modality coordination and correlation learned by the model. Detecting these types of anomalies and flagging them for fixing is critical for keeping a geospatial mapping service up-to-date in real-time, tracking and fixing regressions in geospatial data due to real-world changes, improving the user experience, reducing errors in geospatial routing, search and recommendation services, and above all, improving user safety.

\section{Notation and Preliminaries}\label{section:notation_and_preliminaries}
This section introduces notation and terminology that helps address the challenges of representation and alignment of the different heterogeneous geospatial data modalities considered in this paper.
\begin{definition}\label{def_zoom_24_tile}
    A \textbf{zoom-q tile} is the cell or tile resulting from a $2^{q} \times 2^{q}$ raster representation of the spherical Mercator\footnote{Proposed SeMAnD method can be used with any choice of projection.} projection \cite{epsg_3857} of the earth's surface. 
\end{definition}
In this paper, (i) image pixels correspond to zoom-26 tiles\footnote{Corresponds to spatial resolution $\approx 0.6$m at the equator.}; (ii) image dimensions correspond to zoom-18 tiles with $256 \times 256$ pixels (contiguous neighboring zoom-26 tiles). 
\begin{definition}\label{def_gps_trajectory}
    A GPS trajectory, $\boldsymbol{T}_i \in \boldsymbol{\mathcal{T}}(t_0, \Delta t)$, or simply \textbf{trajectory}, is a discrete, sequential, and chronological representation of the spatiotemporal movement of an object \cite{zheng15}. $\boldsymbol{T}_i$ consists of ordered pairs, $\boldsymbol{T}_i = \{\boldsymbol{r}_1^i, \ldots, \boldsymbol{r}_{n_i}^i\}$, where pair, $\boldsymbol{r}^i_k = \{z_k^i, t_k^i \}$, for $k = 1, \ldots, n_i$. $\boldsymbol{r}^i_k$ is called a GPS record. $\boldsymbol{\mathcal{T}}(t_0,\Delta t)$ is the set of $|\boldsymbol{\mathcal{T}}(t_0, \Delta t)| = M$ trajectories occurring within observation time interval, $\Delta t$, starting at $t_0$, $n_i$ is length of trajectory $\boldsymbol{T}_i$, and $z^i_k$ is the zoom-26 tile\footnote{Finite area tiles mitigate handling real-valued latitude-longitude pairs. The tile represents all latitude-longitude pairs within itself.} where the object with trajectory $\boldsymbol{T}_i$ was present at time $\boldsymbol{t}_k^i$. Note: $t_1^i < \ldots < t_{n_i}^i$ for $i = 1, \ldots, M$.
\end{definition}
A pre-trained, LSTM-based, neural motion modality classifier is used to associate each GPS record with a motion modality, e.g., walking, driving, biking, etc. The full trajectory dataset, $\boldsymbol{\mathcal{T}}(t_0, \Delta t)$ is filtered to generate modality-specific trajectory datasets by dropping the GPS records not belonging to the chosen motion modality. In this paper, we restrict our attention to only walking and driving motion modalities. The driving and walking trajectory datasets are denoted as $\boldsymbol{\mathcal{T}}_{d}(t_0, \Delta t)$ and $\boldsymbol{\mathcal{T}}_{w}(t_0, \Delta t)$ respectively.   
\begin{definition}\label{def_road_casement_polygons}
    A road casement polygon (RCP) is a closed shape defined by a connected sequence of latitude-longitude coordinate pairs that encloses a portion of the casement or pavement of the road while conforming to the width of the road on the earth's surface.
\end{definition}
\begin{definition}\label{def_road_network}
    The road network is a graph, $\boldsymbol{G}_{r}(\boldsymbol{V}_{r}, \boldsymbol{E}_{r})$, that encodes the topology, spatial structure, and connectivity of roads. Vertices, $\boldsymbol{V}_{r}$, are represented with a latitude-longitude coordinate pair and denote intersections or the starting/ending of road segments. Edges, $\boldsymbol{E}_{r}$, denote the road segments connecting two vertices.
\end{definition}
In this paper, $\boldsymbol{E}_{r}$ consists only of road segments that are navigable (i.e., either drivable, walkable, or both).

\section{Representation, Alignment, and Fusion of Heterogeneous Geospatial Data Modalities}\label{section:repn_algn_fusn}
This work uses satellite imagery, road network graph, polygonal vector geometries, and GPS trajectories as data modalities. Representing heterogeneous data modalities, aligning these representations, and fusing aligned representations for multimodal prediction are three key challenges in multimodal machine learning \cite{baltruvsaitis2018}. Strategies addressing each of these tasks are tightly coupled with the following requirements \cite{baltruvsaitis2018}: (i) Representations must summarize the information contained in the different modalities that exploit their complementarity and redundancy; (ii) Alignment must identify the direct relationship between the subelements of the different modalities; (iii) Fusion must take advantage of the different natural structures, predictive power, and noise topologies of the modalities for the downstream multimodal prediction. In geospatial datasets, spatial locality and spatial correlations of different modalities is important \cite{iyer2023perspectives}. With this background, we outline below our proposed strategy \cite{xiao2020vae,iyer2021trinity,ganguli2022reachability} for representation, alignment, and fusion of the geospatial datasets considered in this work.

\begin{figure}
    \centering
    \includegraphics[width=0.40\textwidth]{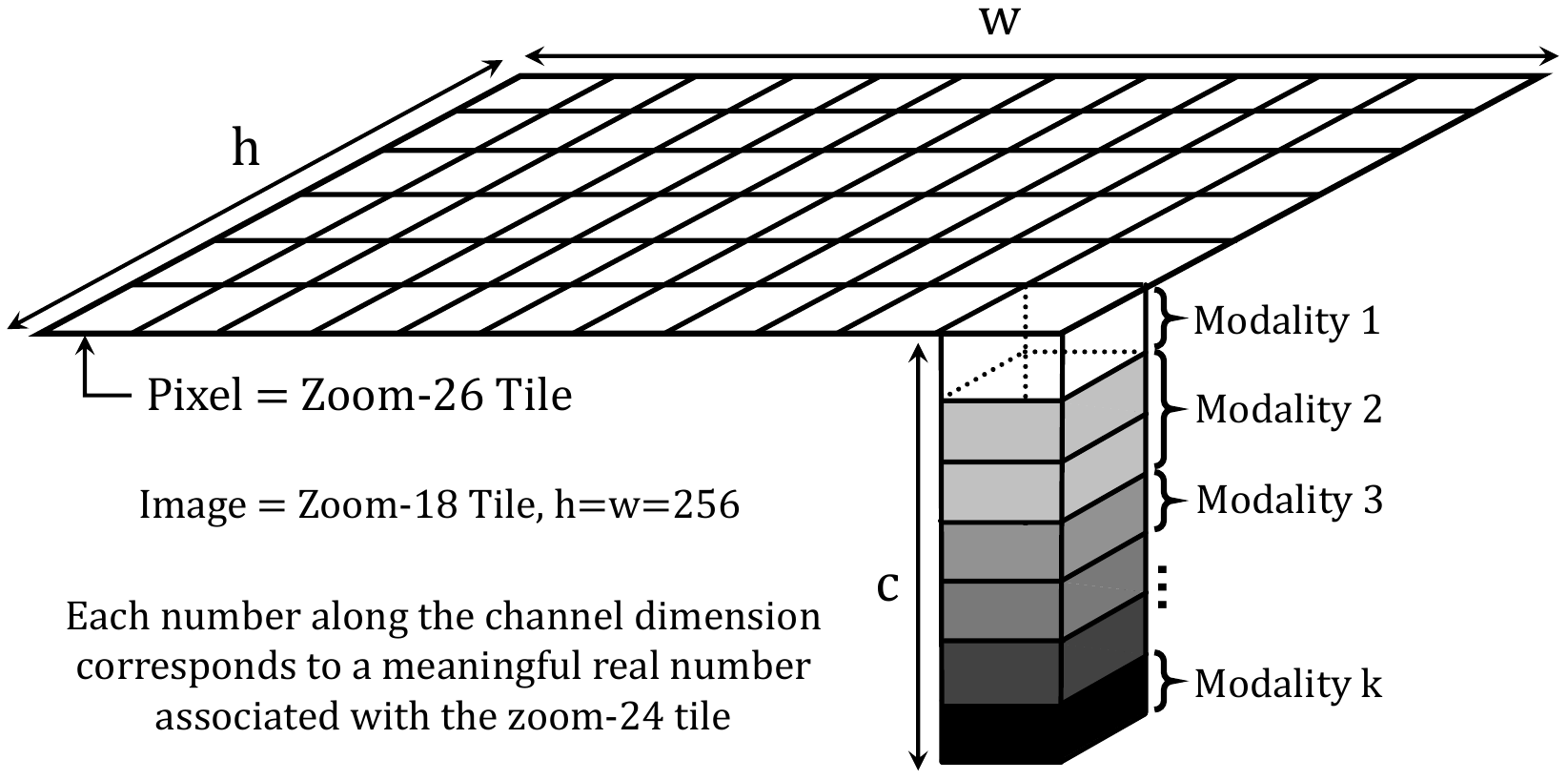}
    \caption{Schematic demonstrating fusion via channel concatenation of pixel-aligned (implying spatially aligned) geospatial data modality representations yielding a multimodal image-like tensor.}
    \label{figure:representation_alignment_and_fusion}
\end{figure}

\subsection{Representation}\label{section::representation}
Given the Cartesian grid structure implied by zoom-26 tiles (\cref{figure:representation_alignment_and_fusion}), associating $\boldsymbol{c}$ meaningful real numbers (represented as $\boldsymbol{c}$ channels) generated from a data modality to each zoom-26 tile (pixel) within any zoom-18 tile forms a $256 \times 256 \times \boldsymbol{c}$-dimensional image-like tensor representation of the data modality in the geographic location represented by the zoom-18 tile. Satellite imagery has $\boldsymbol{c}=3$ channels (RGB). Mobility data from GPS trajectories is converted to \textit{count-based raster maps} (CRM). A CRM is a single-channel representation ($\boldsymbol{c} = 1$) where the value of each pixel is the number of occurrences of GPS records in the zoom-26 tile corresponding to the pixel counted over all trajectories in $\boldsymbol{\mathcal{T}}(t_0, \Delta t)$. CRMs are generated separately for driving (DCRM) and walking (WCRM) motion modalities using $\boldsymbol{\mathcal{T}}_{d}(t_0, \Delta t)$ and $\boldsymbol{\mathcal{T}}_{w}(t_0, \Delta t)$ respectively. Trajectory-independent road geometry obtained from the road network graph is rasterized to yield the \textit{road network presence} (RNP) channel ($\boldsymbol{c}=1$), with a binary representation such that each pixel is assigned the value 1 if the corresponding zoom-26 tile has at least one road segment present and the value 0 otherwise. A similar rasterization process yields the image-like binary representation of road casement polygons, called the \textit{road casement polygon presence} (RCPP) channel ($\boldsymbol{c}=1$), where pixels with at least one road casement polygon present are assigned the value 1 and the value 0 otherwise.

\subsection{Alignment and Fusion}\label{section::alignment_and_fusion} 
The aforementioned representation strategy allows for explicit spatial alignment of the different modalities by simply aligning the image-like tensor representations at the pixel level (\cref{figure:representation_alignment_and_fusion}). Pixel-aligned representations are fused via concatenation along the channel dimension (early fusion \cite{baltruvsaitis2018}). With all four data modalities fused together, the multimodal image-like tensor representation has seven channels. 

Analogous to spectral imaging for remote sensing, where channels corresponding to different spectral bands capture additional information of a geographic location to complement their RGB (visible) counterparts, the WCRM and DCRM are image-like representations of mobility patterns deduced from trajectory data. RNP can be considered as an \textit{ontological} representation (where traffic \textit{should} exist) of mobility while WCRM and DCRM are the \textit{epistemological} representations (where traffic \textit{actually} exists) of mobility. In summary, each data modality is transformed into a semantically meaningful image-like tensor representation preserving their natural spatial locality leading to natural strategies for alignment and fusion of heterogeneous modalities. The fused multimodal artifacts are amenable to analysis using computer vision techniques such as convolutional neural networks with inductive biases that exploit the spatial locality of the input channels.

\section{Methodology}\label{section:methodology}
\begin{figure*}
    \centering
    
    \sbox{\bigpicturebox}{
        \begin{subfigure}[b]{0.6\textwidth}
            \scalebox{1}[1]{\includegraphics[width=\textwidth]{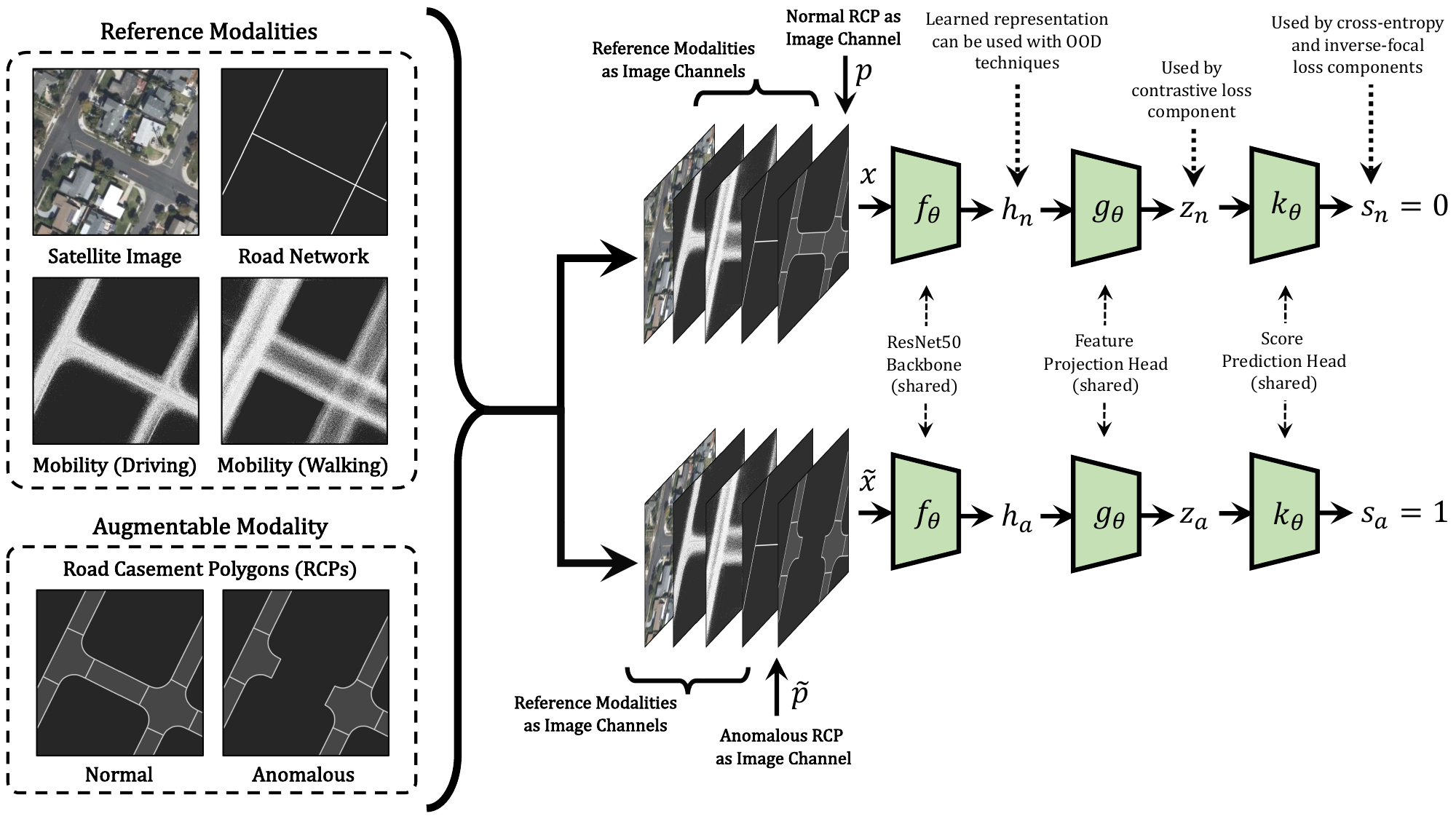}}
            \caption{Schematic of the RandPolyAugment self-supervised training strategy.}
            \label{figure:schematic}
        \end{subfigure}
    }
    \usebox{\bigpicturebox}\hfill
    \begin{minipage}[b][\ht\bigpicturebox][s]{.37\textwidth}
        \vspace{0.5cm}
        \centering
        \begin{subfigure}{0.30\textwidth}
            \includegraphics[width=\textwidth]{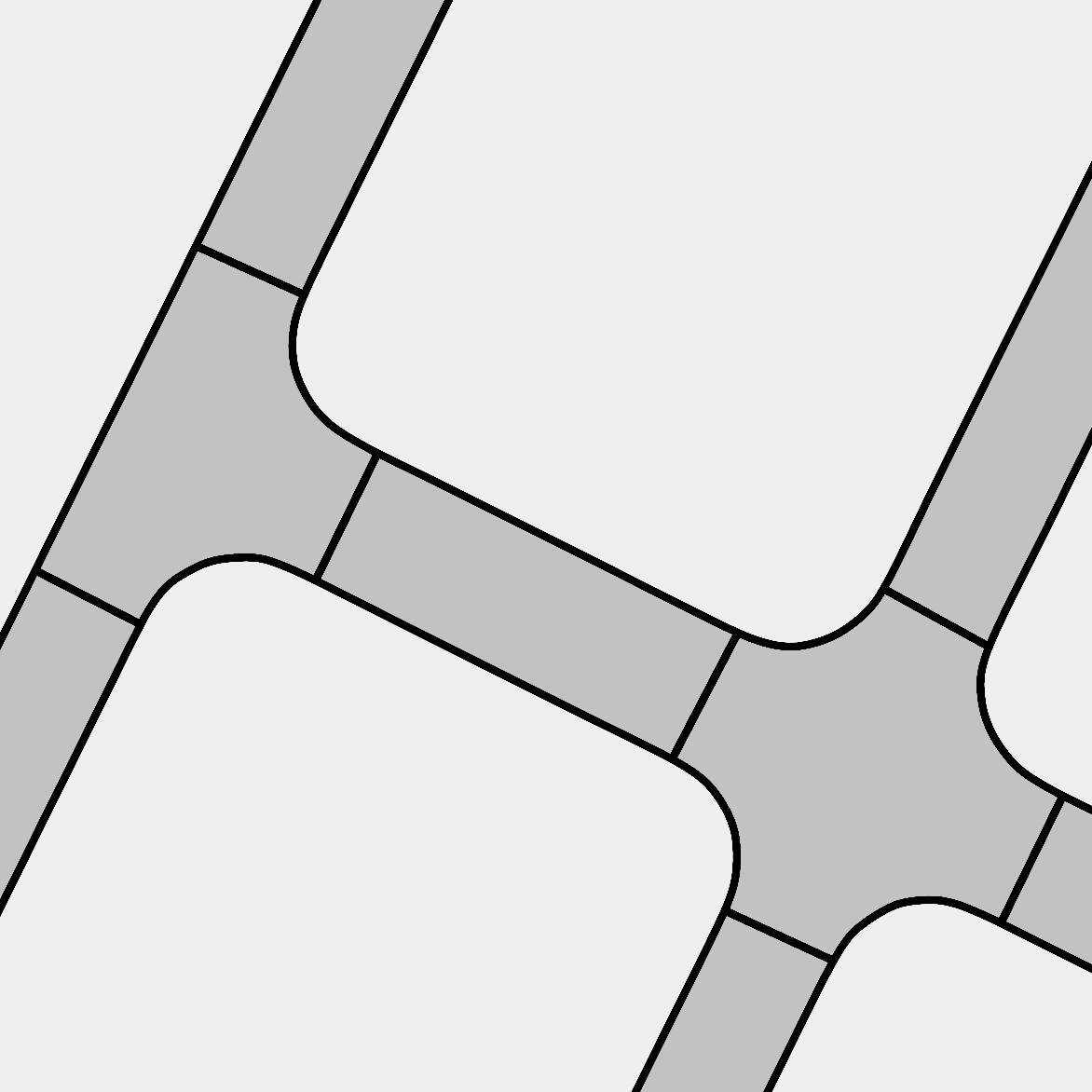}
            \caption{Normal}
            \label{figure:normal}
        \end{subfigure}\hfill
        \begin{subfigure}{0.30\textwidth}
            \includegraphics[width=\textwidth]{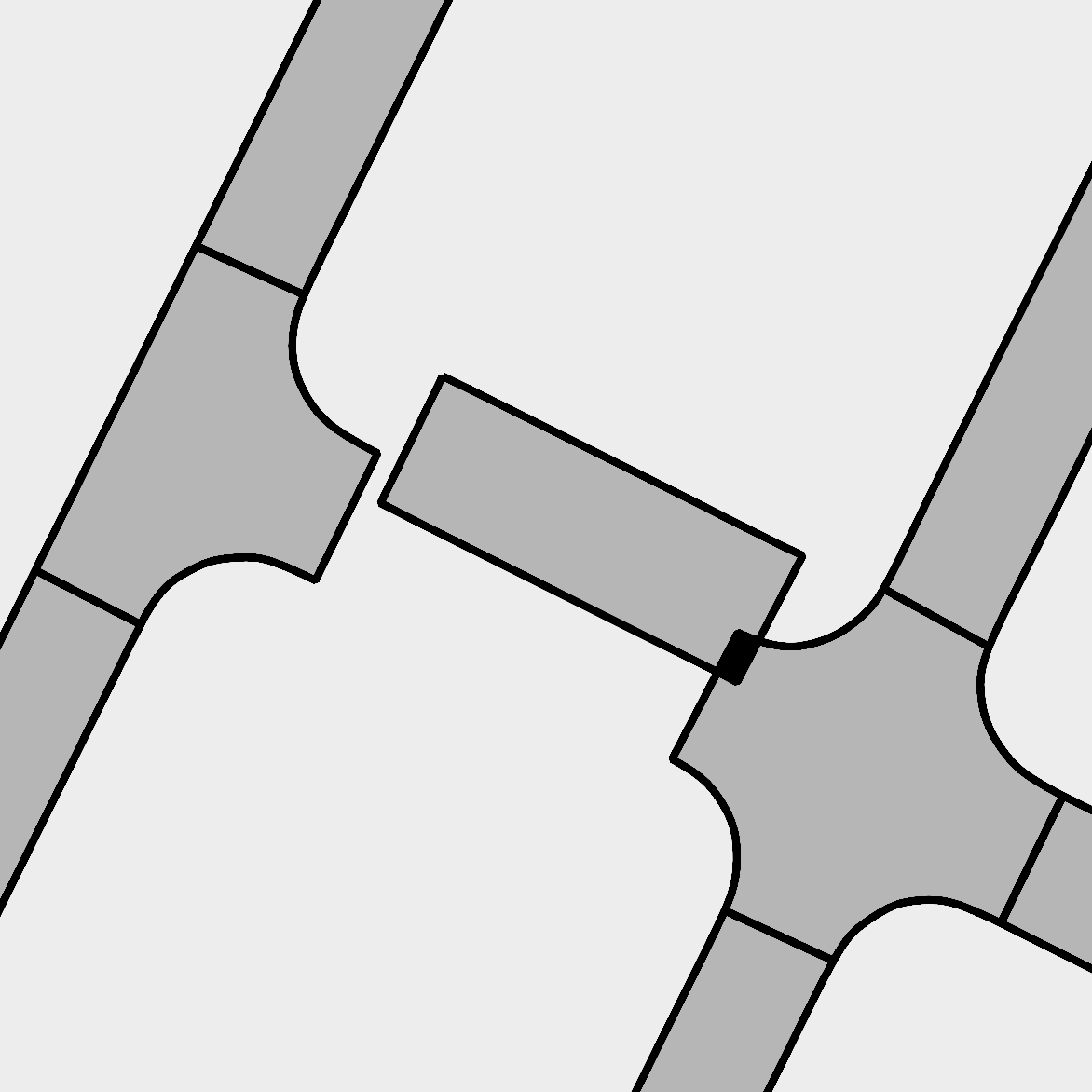}
            \caption{Translated}
            \label{figure:translated}
        \end{subfigure}\hfill
        \begin{subfigure}{0.30\textwidth}
            \includegraphics[width=\textwidth]{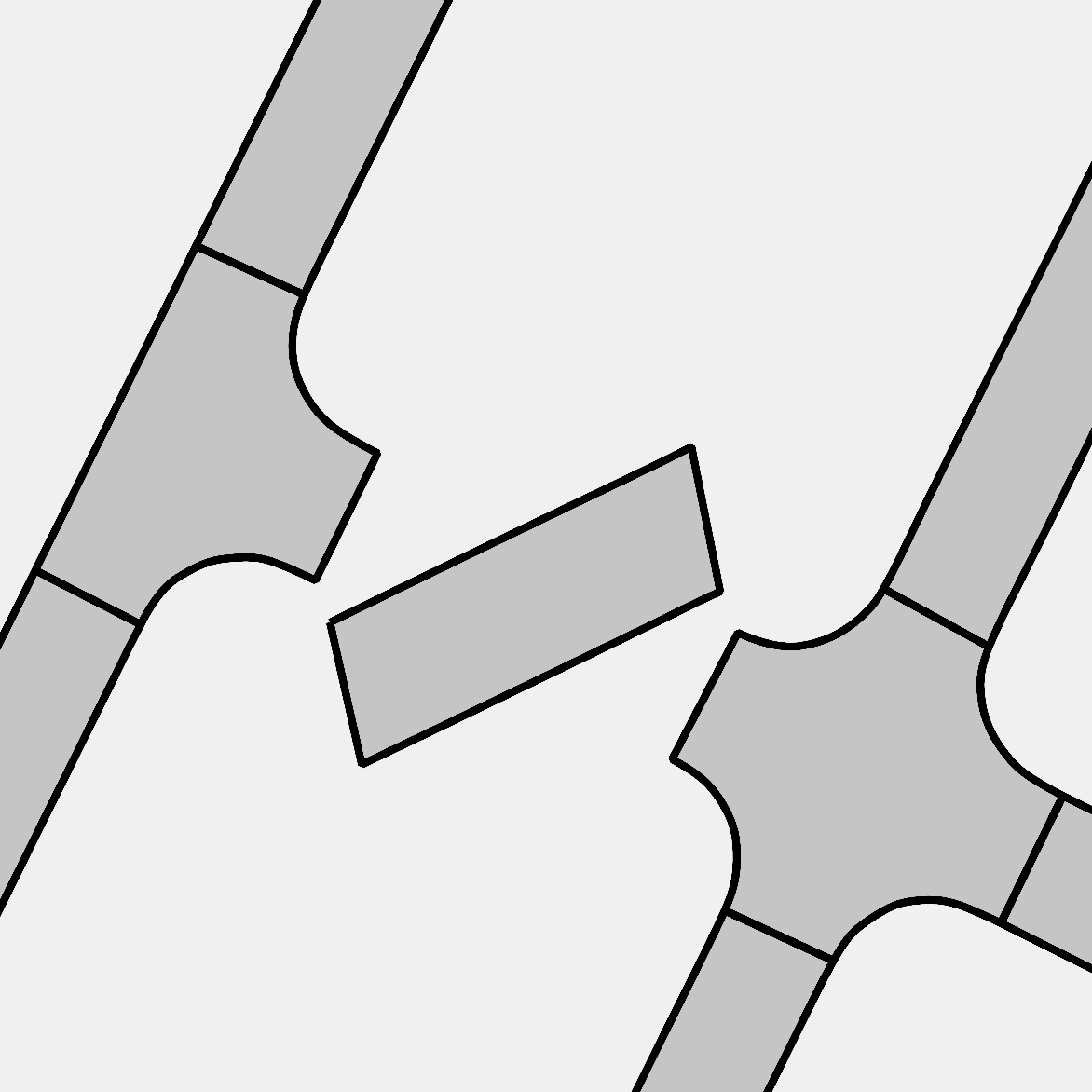}
            \caption{Rotated}
            \label{figure:rotated}
        \end{subfigure}

        \vspace{0.5cm}

        \begin{subfigure}{0.30\textwidth}
            \includegraphics[width=\textwidth]{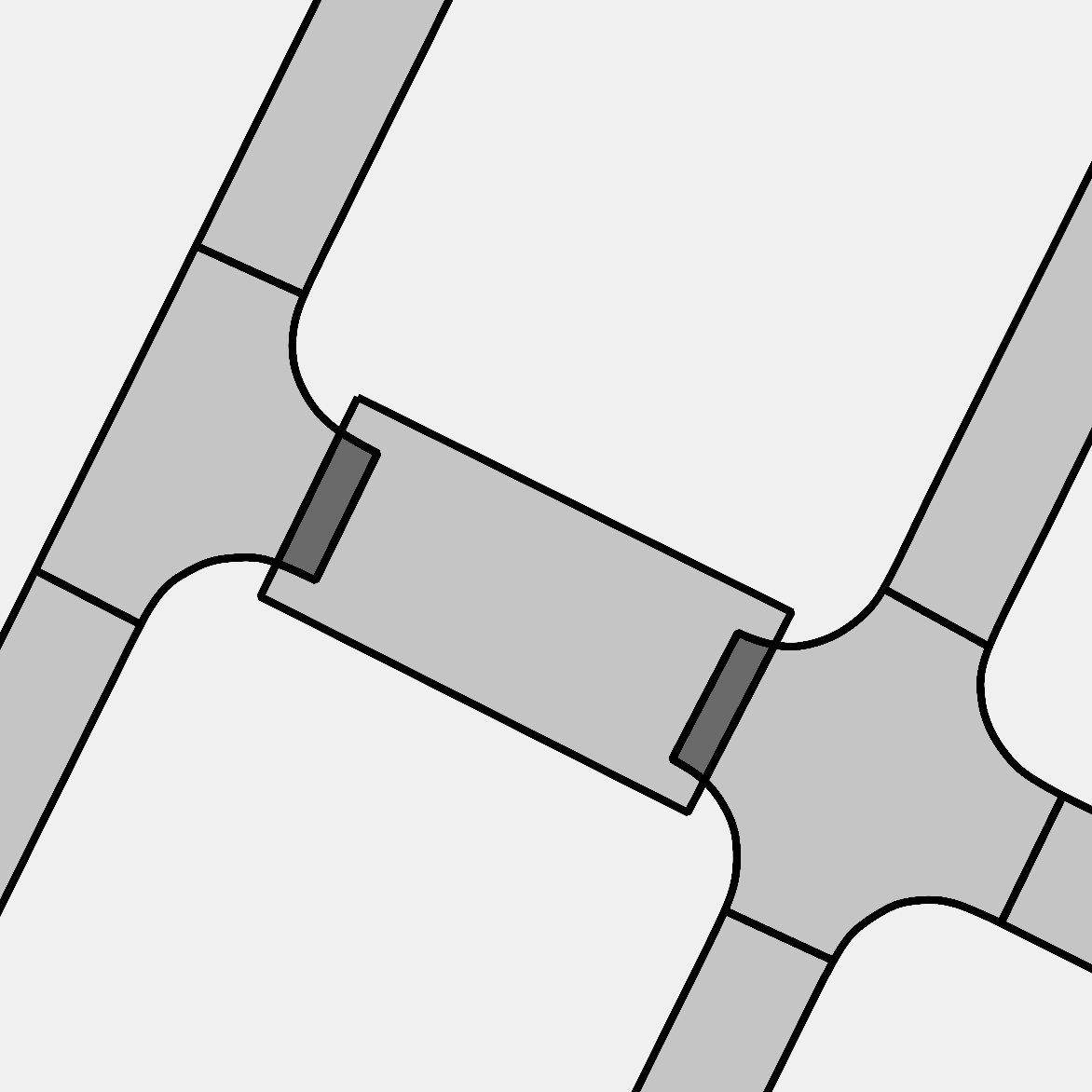}
            \caption{Scaled}
            \label{figure:scaled}
        \end{subfigure}
        \hspace{0.2cm}
        \begin{subfigure}{0.30\textwidth}
            \includegraphics[width=\textwidth]{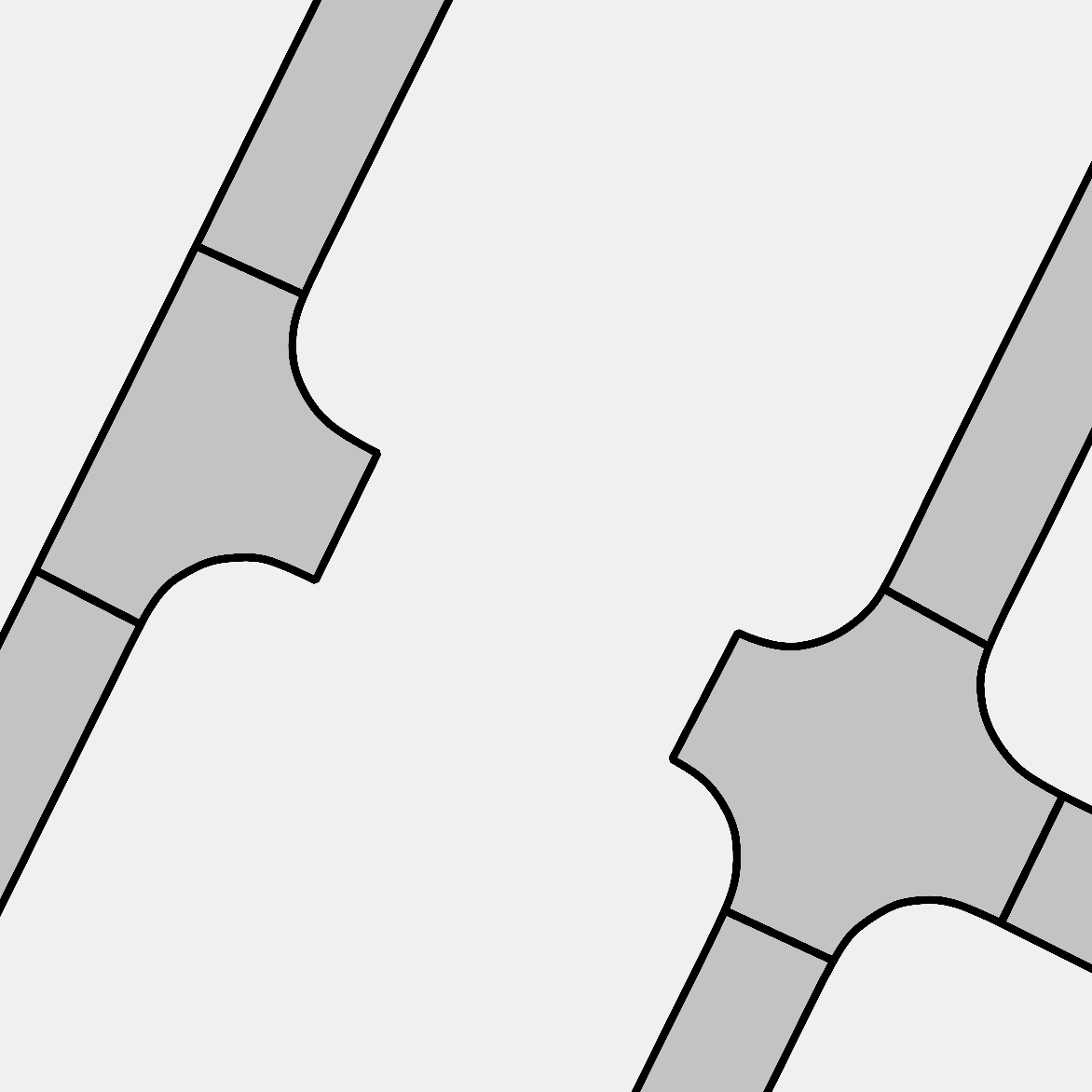}
            \caption{Deleted}
            \label{figure:deleted}
        \end{subfigure}
    \end{minipage}
    
    \caption{A schematic of the RandPolyAugment self-supervised training strategy is shown in Figure (a) demonstrating representation, alignment, and fusion of the different data modalities. The affine transformations used on the road casement polygons are shown in Figures (c)--(f) by comparing their impact to arrangement in a normal instance shown in Figure (b).}
    \label{figure:schematic_and_affine_transforms}
\end{figure*}

Defining good pretext tasks and training objectives is essential for self-supervised anomaly detection \cite{li2021,reiss2023anomaly,hojjati2022self}. This section outlines SeMAnD's self-supervised training strategy, schematically shown in \cref{figure:schematic}, that treats satellite imagery, WCRM, DCRM, and RNP implicitly as reference modalities (by not augmenting them) to learn separate representations (targeting anomaly detection) of the multimodal data for original and augmented versions of RCPP. \cref{subsection:randpolyaugment} and \cref{ssl_with_rpa} describe the two components of SeMAnD viz., the RandPolyAugment data augmentation strategy and the self-supervision training objective respectively.

\subsection{RandPolyAugment Augmentation Strategy}\label{subsection:randpolyaugment}
The key intuition behind the RandPolyAugment (RPA) strategy, implemented in \cref{algorithm:algorithm_randpolyaugment}, is that complex augmentations of RCPs can be composed by successively applying a random sequence of simple affine transformations (with parameters chosen uniformly at random from a specified interval) on individual polygons\footnote{Analogous to the motivation of integral transforms where complex functions are decomposed into the sum of many simpler functions.}. An RCP can span multiple tiles. RandPolyAugment is applied to the RCPP channel. The set of allowed transformations ($\boldsymbol{A}$) are translation (\cref{figure:translated}), rotation (\cref{figure:rotated}), scaling (\cref{figure:scaled}), and deletion (\cref{figure:deleted}).

\begin{algorithm}
    \small
    \caption{RandPolyAugment (RPA) Strategy}\label{algorithm:algorithm_randpolyaugment}
    \begin{algorithmic}[1]
        \INPUT RCPP of a zoom-18 tile $\boldsymbol{\psi}$ composed of $n_{\psi}$ RCP polygons, $\boldsymbol{P}_{\psi} = \{\boldsymbol{p}_{\psi}^{1}, \ldots, \boldsymbol{p}_{\psi}^{n_{\psi}}\}$; Set of affine transformation actions, $\boldsymbol{A} =$  \{\texttt{rotate}, \texttt{translate}, \texttt{scale}, \texttt{delete}\}
        \OUTPUT $\widetilde{\boldsymbol{\psi}} \leftarrow$ Augmented RCPP of the tile
        \PARAMETERS $0<$ $\theta_{0}$ $<$ $\theta_{1}$, $0<$ $\delta_{0}$ $<$ $\delta_{1}$, $0<$ $\beta^{-}$ $<$ $\beta_{0}^{-}$ $<1<$ $\beta_{0}^{+}$ $<$ $\beta^{+}$, $0 < p_s < 1$, Posedness = $\rho$
        \INITIALIZE Random seeds, Set $\boldsymbol{S}_{\widetilde{\psi}}=\emptyset$
        \Function{randPolyAugment}{$\boldsymbol{\psi}$}
            \State Set $\boldsymbol{S}_{\psi} \leftarrow$ Choose $\boldsymbol{p}_{\psi}^j$ with prob. $p_s$ $\forall$ $j$ $\in$ 1,$\ldots$,$n_{\psi}$
            \LineComment{If $\boldsymbol{S}_{\psi}=\emptyset$, sample 1 polygon uniformly at random}
            \LineComment{Polygons in $\boldsymbol{P}_{\psi}\setminus\boldsymbol{S}_{\psi}$ are left unchanged}
            \For{polygon $\boldsymbol{p} \in \boldsymbol{S}_{\psi}$}
                \State Set $\boldsymbol{S}_{a} \leftarrow$ Choose $a$ with prob. $\frac{1}{2}$ $\forall$ $a$ $\in$ $\boldsymbol{A}$
                \LineComment{If $\boldsymbol{S}_{a}=\emptyset$, randomly sample 1 action from $\boldsymbol{A}$}
                \State Ordered Sequence $\overrightarrow{\boldsymbol{S}}_{a} \leftarrow$ Randomly shuffle $\boldsymbol{S}_{a}$
                \For{action $\boldsymbol{a} \in \overrightarrow{\boldsymbol{S}}_{a}$}
                    \If{$\boldsymbol{a} ==$ \texttt{rotate}}
                        \State Sample $\theta \sim \mathbf{U}([-\theta_{1},-\theta_{0}] \cup [\theta_{0},\theta_{1}])$
                        \State Rotate $\boldsymbol{p}$ by $\theta$ around its centroid
                    \EndIf
                    \If{$\boldsymbol{a} ==$ \texttt{translate}}
                        \State Sample $\delta_x, \delta_y \sim \mathbf{U}([-\delta_{1},-\delta_{0}] \cup [\delta_{0},\delta_{1}])$
                        \State Translate $\boldsymbol{p}$ by $\delta_x$ in $x$ and $\delta_y$ in $y$ from centroid
                    \EndIf
                    \If{$\boldsymbol{a} ==$ \texttt{scale}}
                        \State Sample $\beta_{x}, \beta_{y} \sim \mathbf{U}([\beta^{-},\beta_{0}^{-}] \cup [\beta_{0}^{+},\beta^{+}])$
                        \State Scale $\boldsymbol{p}$ by $\beta_x$ in $x$ and  $\beta_y$ in $y$ from centroid
                    \EndIf
                    \If{$\boldsymbol{a} ==$ \texttt{delete}}
                        \State Delete $\boldsymbol{p}$; \textbf{break}
                    \EndIf
                \EndFor
                \LineComment{Cut p to make sure it is within tile $\psi$'s spatial extent}
                \State $\boldsymbol{S}_{\widetilde{\psi}}.\text{insert}(\boldsymbol{p})$ \Comment{\textit{Assuming} $\boldsymbol{p}$ \textit{is not deleted}}
            \EndFor
            \State $\widetilde{\boldsymbol{\psi}} \leftarrow$ Tile with polygons $\boldsymbol{S}_{\widetilde{\psi}} \cup \{\boldsymbol{P}_{\psi} \setminus \boldsymbol{S}_{\psi}\}$
            \State \Return $\widetilde{\boldsymbol{\psi}}$
        \EndFunction
        \LineComment{Wrapper function for acceptance-rejection method}
        \Function{augmentWithPosedness}{$\boldsymbol{\psi}$, $\rho$}
            \While{\texttt{True}}
                \State $\boldsymbol{\widetilde{\psi}} \leftarrow \texttt{randPolyAugment}(\boldsymbol{\psi})$
                \If{$\|\widetilde{\boldsymbol{\psi}} - \boldsymbol{\psi}\|_{F}/\|\boldsymbol{\psi}\|_{F} > \rho$}
                    \State \Return $\widetilde{\boldsymbol{\psi}}$
                \EndIf
            \EndWhile
        \EndFunction
    \end{algorithmic}
\end{algorithm}

Consider the RCPP channel of a zoom-18 tile, $\boldsymbol{\psi}$, which intersects a set of $n_{\psi}$ RCPs, $\boldsymbol{P}_{\psi} = \{\boldsymbol{p}_{\psi}^{1}, \ldots, \boldsymbol{p}_{\psi}^{n_{\psi}}\}$. The first step (steps 2--4) is to choose the set of polygons that will be augmented, denoted as $\boldsymbol{S}_{\psi}$, by adding each polygon in $\boldsymbol{P}_{\psi}$ to $\boldsymbol{S}_{\psi}$ with probability $p_s$. We found the optimal value of $p_s=0.5$. For each polygon in $\boldsymbol{S}_{\psi}$: (i) a random set of affine transformations ($\boldsymbol{S}_a$) are chosen (step 6--7) from $\boldsymbol{A}$ and shuffled (step 8) to create an ordered sequence of transformations, $\overrightarrow{\boldsymbol{S}}_{a}$; (ii) each action in $\overrightarrow{\boldsymbol{S}}_{a}$ is applied (steps 9--22) to generate the augmented polygon. Replacing polygons in $\boldsymbol{S}_{\psi}$ by their augmented counterparts yields the augmented RCPP channel of the tile, $\widetilde{\boldsymbol{\psi}}$.

The translation, rotation, and scaling transformations are performed by treating the centroid of the polygon as the origin. The rotation transformation rotates the polygon around the centroid by an angle $\theta$ sampled from $\mathbf{U}([-\theta_{1},-\theta_{0}] \cup [\theta_{0},\theta_{1}])$. The translation transformation translates the polygon from the centroid in the $x$ (longitude) and $y$ (latitude) directions by $\delta_x$ and $\delta_y$ respectively, which are sampled from $\mathbf{U}([-\delta_{1},-\delta_{0}] \cup [\delta_{0},\delta_{1}])$. The scaling transformation scales the polygon by $\beta_x$ along the $x$ direction and $\beta_y$ along the $y$ direction which are sampled from $\mathbf{U}([\beta^{-},\beta_{0}^{-}] \cup [\beta_{0}^{+},\beta^{+}])$. Here, $0<$ $\theta_{0}$ $<$ $\theta_{1}$, $0<$ $\delta_{0}$ $<$ $\delta_{1}$, and $0<$ $\beta^{-}$ $<$ $\beta_{0}^{-}$ $<1<$ $\beta_{0}^{+}$ $<$ $\beta^{+}$ are input parameters provided by the user. The intervals $[-\theta_{0},\theta_{0}]$, $[-\delta_{0},\delta_{0}]$, and $[\beta_0^-,\beta_0^+]$ are excluded from the sampling intervals for rotation, translation, and scaling respectively, to avoid small augmentations. In our experiments, the optimal values for these parameters were found to be $\theta_0=\frac{\pi}{6}$, $\theta_1=\frac{\pi}{2}$, $\delta_0=0.00006$ decimal degrees, $\delta_1=0.00012$ decimal degrees, $\beta^{-}=0.2$, $\beta_{0}^{-}=0.7$, $\beta_{0}^{+}=1.4$, and $\beta^{+}=3$. 

A scalable and distributed implementation of \cref{algorithm:algorithm_randpolyaugment} using Apache Spark \cite{spark} is used to generate RCPP augmentations for each zoom-18 tile in the training dataset. The number of augmentations per tile is equal to the number of training epochs. A critical parameter influencing the performance of trained models (\cref{results}) is the minimum anomalous area in the generated augmentations. We define a parameter $\rho$, called \textit{posedness}, as the normalized ratio of the Frobenius norm of the pixel-wise difference of $\boldsymbol{\psi}$ and $\widetilde{\boldsymbol{\psi}}$, i.e., $\rho = \|\widetilde{\boldsymbol{\psi}}-\boldsymbol{\psi}\|_F/\|\boldsymbol{\psi}\|_F$. As $\rho$ increases, the size of the anomalous regions increases leading the classification (\cref{cross_entropy_loss} and \cref{inverse_focal_loss}) and contrastive (\cref{contrastive_loss}) tasks defined over the learned feature representations to be better posed, while also influencing the model's sensitivity. \cref{algorithm:algorithm_randpolyaugment} uses the acceptance-rejection method (steps 26--30) to generate augmentations with posedness above the user-specified threshold, $\rho$.

\subsection{SeMAnD Self-Supervision Training Objectives}\label{ssl_with_rpa}
For each zoom-18 tile in the training set, the image-like tensor representations (\cref{section:repn_algn_fusn}) of the reference modalities are pixel-aligned and concatenated along the channel dimension to the rasterized normal ($\boldsymbol{\psi}$) and augmented ($\widetilde{\boldsymbol{\psi}}$) RCPP channel to create the normal and augmented multimodal instances for the tile, $\boldsymbol{x}$ and $\widetilde{\boldsymbol{x}}$ respectively (\cref{figure:schematic}). Early fusion preserves the spatial locality of the different data modalities. All input channels are normalized so that pixel values are in $[0,1]$; WCRM and DCRM (which have large dynamic ranges) are log-normalized and then linearly normalized while other channels are linearly normalized. Training data (600,928 normal zoom-18 tiles) is collected from diverse geographies (in terms of tile distribution based on urban-rural stratification, landcover, population density, and activity stratas) using stratified sampling \cite{cao2023self} to ensure proportional representation of tiles from the different stratas.

The neural architecture of the model (\cref{figure:schematic}) is inspired from SimCLR \cite{chen2020}. The model consists of three main components: (1) a \textit{backbone} neural network, $\boldsymbol{f}_{\theta}$, to generate a representation vector of the input; (2) a small \textit{contrastive projection head}, $\boldsymbol{g}_{\theta}$, to map the learned representation into a space where the contrastive loss is applied; (3) a small \textit{score prediction head}, $\boldsymbol{k}_{\theta}$, which maps the representation learned by the contrastive projection head to the probabilities for binary classification of the input being normal or augmented (anomalous). We model $\boldsymbol{f}_{\theta}$ as a ResNet50 \cite{he2016}, so that $\boldsymbol{h}_{n}=\boldsymbol{f}_{\theta}(\boldsymbol{x})$ and $\boldsymbol{h}_{a}=\boldsymbol{f}_{\theta}(\widetilde{\boldsymbol{x}})$. $\boldsymbol{g}_{\theta}$ is a MLP with two hidden layers followed by ReLU leading to contrastive features with non-negative components, $\boldsymbol{z}_{n}=\boldsymbol{g}_{\theta}(\boldsymbol{h}_{n})$, $\boldsymbol{z}_{a}=\boldsymbol{g}_{\theta}(\boldsymbol{h}_{a})$. $\boldsymbol{k}_{\theta}$ is an MLP with two hidden layers followed by the softmax activation yielding scores $\boldsymbol{s}_{n}=\boldsymbol{k}_{\theta}(\boldsymbol{z}_{n})$, $\boldsymbol{s}_{a}=\boldsymbol{k}_{\theta}(\boldsymbol{z}_{a})$. The dimensions of the vectors $\boldsymbol{h}$, $\boldsymbol{z}$, and $\boldsymbol{s}$ are 2048, 128, and 2 respectively. The individual components of $\boldsymbol{s}$ are denoted as $\boldsymbol{s}=[\boldsymbol{s}^{0}, \boldsymbol{s}^{1}]$.

In each training iteration, a minibatch of $N$ normal and their corresponding augmented examples are randomly sampled. The training objective has three components. Following the idea of CutPaste \cite{li2021}, the first component ($\mathcal{L}_{BC}$) is \textit{cross-entropy loss} computed for the binary classification task of associating the normal example, $\boldsymbol{x}$, to the label 0 and the corresponding augmentation, $\widetilde{\boldsymbol{x}}$, to the label 1.
\begin{equation}\label{cross_entropy_loss}
    \mathcal{L}_{BC} = -\mathbf{E}_{\boldsymbol{x}}[\log{(\boldsymbol{s}_{n}^{0})}] -\mathbf{E}_{\widetilde{\boldsymbol{x}}}[\log{(\boldsymbol{s}_{a}^{1})}].
\end{equation}

\begin{figure}
    \centering
    \includegraphics[width=0.30\textwidth]{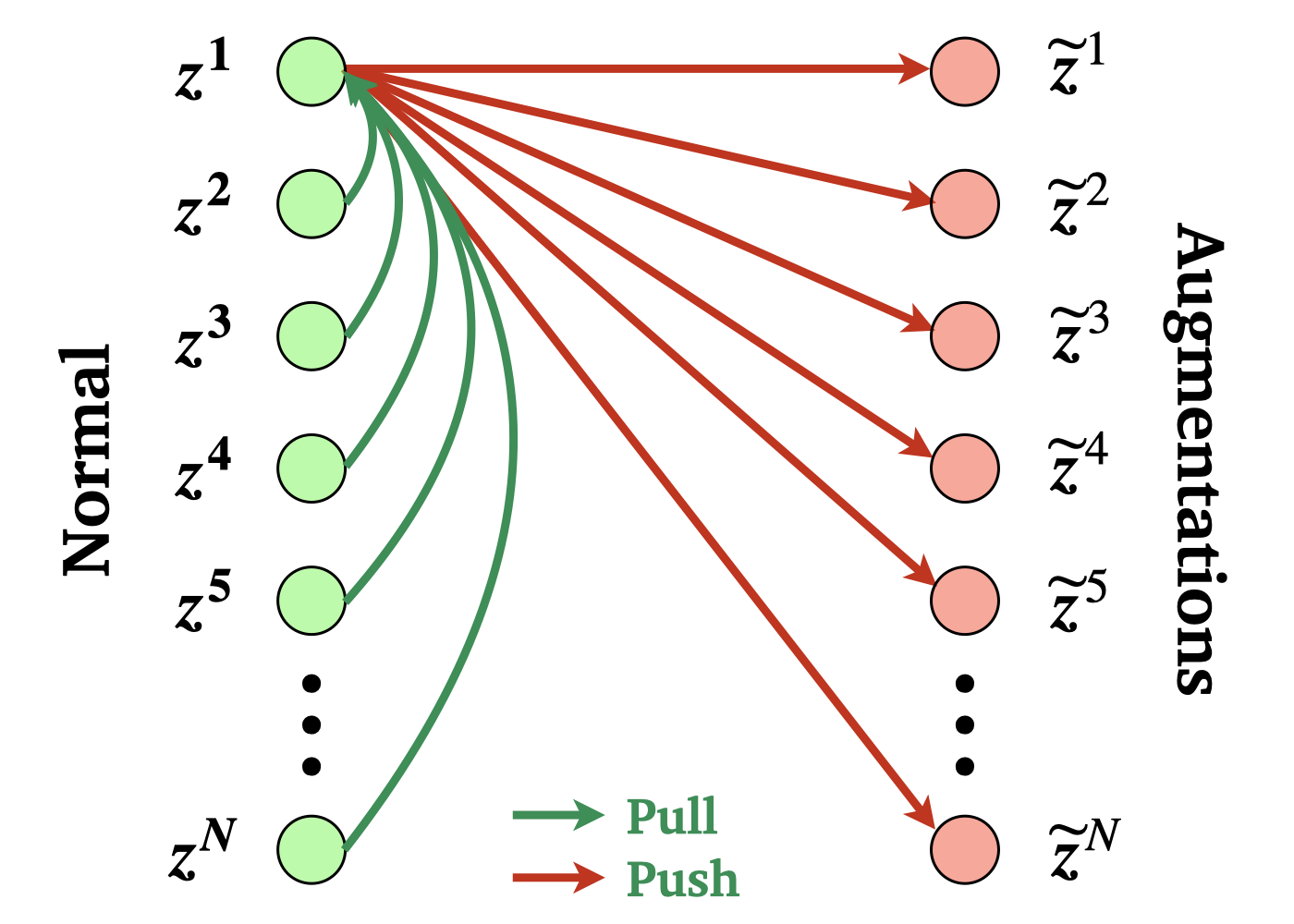}
    \caption{Schematic describing the SeMAnD contrastive loss component, $\mathcal{L}_{CL}$, from the perspective of the normal example with contrastive feature $\boldsymbol{z}^1$ in a minibatch (batch size $2N$) with $N$ normal instances and $N$ augmented instances.}
    \label{figure:sentinel_contrastive_loss_schematic_diagram}
\end{figure}

Following the idea of SimCLR \cite{chen2020}, the second component is the contrastive loss ($\mathcal{L}_{CL}$) that treats features of any two normal examples or any two augmented examples from the minibatch as a positive pair (pull closer), while treating the features from a pair of normal and augmented examples as a negative pair (push apart). $\mathcal{L}_{CL}$ is designed (following \cite{wang2020}) to minimize agreement (representations pushed away on hypersphere) between learned features of normal and augmented instances of the same tile while implicitly incentivizing clustering (representations pulled together on hypersphere) of all the normal and anomalous features into separate clusters. Each minibatch has $N$ normal instances, denoted by $\boldsymbol{X} = \{\boldsymbol{x}^{1}, \boldsymbol{x}^{2}, \ldots, \boldsymbol{x}^{N}\}$, and $N$ corresponding anomalous instances from RandPolyAugment, denoted by $\widetilde{\boldsymbol{X}} = \{\widetilde{\boldsymbol{x}}^{1}, \widetilde{\boldsymbol{x}}^{2}, \ldots, \widetilde{\boldsymbol{x}}^{N}\}$. The computed contrastive features of the instances are $\{\boldsymbol{z}^{1}, \boldsymbol{z}^{2}, \ldots, \boldsymbol{z}^{N}\}$ and $\{\widetilde{\boldsymbol{z}}^{1}, \widetilde{\boldsymbol{z}}^{2}, \ldots, \widetilde{\boldsymbol{z}}^{N}\}$. For each normal example in $\boldsymbol{X}$, all examples in $\widetilde{\boldsymbol{X}}$ are treated as negative examples while for each anomalous example in $\widetilde{\boldsymbol{X}}$, all examples in $\boldsymbol{X}$ are treated as negative examples. Let $\langle\boldsymbol{u}, \boldsymbol{v}\rangle_{\tau} = \boldsymbol{u}^{\top}\boldsymbol{v}/\tau\|\boldsymbol{u}\|\|\boldsymbol{v}\|$ denote the dot product between $\ell_2$-normalized $\boldsymbol{u}$ and $\boldsymbol{v}$ divided by a temperature parameter $\tau$. Then, $\mathcal{L}_{CL}$ is computed as
\begin{subequations}
    \begin{gather}
        \ell_{ij} = -\ln\frac{\exp(\langle\boldsymbol{z}^{i},\boldsymbol{z}^{j}\rangle_{\tau})}{\exp(\langle\boldsymbol{z}^{i},\boldsymbol{z}^{j}\rangle_{\tau}) + \sum_{k=1}^{N}\exp(\langle\boldsymbol{z}^{i},\widetilde{\boldsymbol{z}}^{k}\rangle_{\tau})}\\
        \widetilde{\ell}_{ij} = -\ln\frac{\exp(\langle\widetilde{\boldsymbol{z}}^{i},\widetilde{\boldsymbol{z}}^{j}\rangle_{\tau})}{\exp(\langle\widetilde{\boldsymbol{z}}^{i},\widetilde{\boldsymbol{z}}^{j}\rangle_{\tau}) + \sum_{k=1}^{N}\exp(\langle\widetilde{\boldsymbol{z}}^{i},\boldsymbol{z}^{k}\rangle_{\tau})}\\
        \mathcal{L}_{CL} = \frac{1}{N}\sum_{i=1}^{N}\frac{1}{2N}\sum_{j=1, j \neq i}^{N}(\ell_{ij} + \widetilde{\ell}_{ij}).
        \label{contrastive_loss}
    \end{gather}
\end{subequations}
A schematic diagram describing the impact of $\mathcal{L}_{CL}$ on the learned contrastive feature representations from the perspective of the normal example with contrastive feature $\boldsymbol{z}^{1}$ in a minibatch with $N$ normal instances is shown in \cref{figure:sentinel_contrastive_loss_schematic_diagram}. 

The third component is \textit{inverse focal loss} ($\mathcal{L}_{IF}$). For anomaly detection, there is a risk of spurious false positives when correctly classified examples are close to the decision boundary. The inverse focal loss \cite{long2017} introduces an exponential weighting to penalize the model's indecisiveness by increasing the relative loss (compared to cross-entropy) on these examples; in contrast to focal loss \cite{lin2017} that reduces the relative loss to focus on hard, misclassified examples.
\begin{equation}\label{inverse_focal_loss}
    \mathcal{L}_{IF} = -\mathbf{E}_{\boldsymbol{x}}[e^{\gamma \boldsymbol{s}_{n}^{0}}\log{(\boldsymbol{s}_{n}^{0})}] -\mathbf{E}_{\widetilde{\boldsymbol{x}}}[e^{\gamma \boldsymbol{s}_{a}^{1}}\log{(\boldsymbol{s}_{a}^{1})}].
\end{equation}
The total loss combining all the three components is 
\begin{equation}\label{total_loss}
    \mathcal{L}_{\text{Total}} = (1/\hat{\lambda})(\lambda_{BC}\mathcal{L}_{BC} + \lambda_{CL}\mathcal{L}_{CL} + \lambda_{IF}\mathcal{L}_{IF}),
\end{equation}
where $\hat{\lambda} = \lambda_{BC} + \lambda_{IF} + \lambda_{CL}$. Exhaustive ablation studies (discussed in \cref{results} and \cref{table:ablation_study_loss_component_weights}) reveal optimal loss hyperparameters to be $\gamma = 1$, $\tau=0.5$, $\lambda_{BC} = \lambda_{CL} = 1.0$, and $\lambda_{IF} = 1.5$. Specifically, if $\mathcal{L}_{\text{Total}}=\mathcal{L}_{BC}$ as in \cite{li2021}, model performance drops by 2.8\%.

To the best of our knowledge, no prior self-supervised anomaly detection work uses the inverse focal loss and the proposed version of contrastive loss as a self-supervision objective. While $\mathcal{L}_{CL}$ is inspired from SimCLR's NT-Xent loss \cite{chen2020} (among others \cite{sohn2016,oord2018,wu2018}), the SimCLR objective tries to learn feature representations that are \textit{invariant} to augmentations while $\mathcal{L}_{CL}$ (and $\mathcal{L}_{\text{Total}}$) tries to incentivize representations that are \textit{discriminative} to augmentations.

Models are implemented in TensorFlow \cite{abadi2016tensorflow} and trained on 8 NVidia V100 GPUs with batch size of 1440 for 100 epochs. The learning rate schedule consists of a linear warm-up for 1 epoch up to a learning rate of $10^{-2}$. Thereafter, the learning rate is reduced with a cosine decay of 0.001 over the next 99 epochs. The AdamW optimizer \cite{loshchilov2017decoupled} with momentum of 0.9 is used. We found that initializing the backbone (except \texttt{conv1} stage to handle 7 channel inputs) with pretrained weights (transfer learning) from models trained on natural RGB imagery and then training with SeMAnD performs better than random initialization. We tried both supervised (ImageNet \cite{kolesnikov2020big}) and self-supervised (\cite{gidaris2018,noroozi2016,doersch2015,chen2020}) pretrained weight initializations that produced gains of 0.9\%, 6.2\%, 2.7\%, 2.1\%, and 2.9\% respectively over models trained with random initialization. Accordingly, SeMAnD's backbone uses SimCLR's bias-free ResNet50 variant and is initialized with pretrained SimCLR weights before training. For the first 10 epochs, the backbone weights (except \texttt{conv1} stage) are frozen and only the projection heads are trained. Thereafter, all layers of the model are unfrozen and jointly trained.

\subsection{Anomaly Score and Localization}\label{subsection:anomaly_score_and_localization}
The proposed model design allows for two different ways of computing the anomaly score. The first is to interpret $\boldsymbol{s}^{1} \in [0,1]$ as the anomaly score. The second is the two-stage approach which uses the representation learned by the contrastive projection head, $\boldsymbol{z}$, in conjunction with out-of-distribution detection (OOD) methods \cite{yang2021} such as one-class classifiers, parametric and non-parametric density-based methods, and distance-based methods that use a geographical region-specific prototype vector (centroid of learned representations) to compute a \textit{relative} anomaly score. In this work, we experimented with multiple techniques including one-class SVM \cite{tax2002}, SVDD \cite{ruff2018}, Gaussian density estimation \cite{rippel2021}, and distance-based methods such as cosine similarity, Euclidean distance, and Mahalanobis distance. Across all test geographies, for OOD methods, we found all methods to be comparable with cosine similarity having a slightly better performance in detecting real anomalies compared to other methods. This superiority of cosine similarity score is hypothesized to be due to its use in the contrastive loss (\cref{contrastive_loss}) as the distance proxy. Results presented in \cref{results} use the cosine similarity score. The computation of an anomaly score, $\boldsymbol{s}$, by the model allows for using a visual explanation tool like GradCAM \cite{selvaraju2017} to identify the regions in the image that lead to higher scores in anomalous inputs. We find (\cref{results}, \cref{figure:examples_anomaly_localization}) that the localization map computed by GradCAM correlates well with locations of anomalies.

\subsection{Dataset Description}\label{subsection::dataset_description}
Small, publicly available GPS trajectory datasets (e.g., \cite{ucigpsdataset13,osmgpstraces21,msrtaxi11}) (i) have varying sampling rates with incomplete trajectories, (ii) are geographically incomplete, (iii) have inconsistent observation times and intervals, and (iv) lack geographical diversity. To the best of our knowledge, there is no sufficiently large, geographically diverse, publicly available, privacy-preserving GPS trajectory dataset covering an observation time interval which can be matched to other data modalities like satellite imagery, road casement polygons, and road network graph collected in the same geographical regions. To overcome this difficulty, we use \textit{probe data} \cite{tcrunch18}, a privacy-preserving proprietary GPS trajectory dataset, to generate WCRM and DCRM channels. Similarly, matching proprietary sources of satellite imagery, road network graph, and vector geometry data are used for generating the other input channels.

\section{Related Work}\label{section:related_work}
Various approaches \cite{ruff2021unifying,salehi2021unified} to deep anomaly detection \cite{chalapathy2019, pang2021} such as one-class classification \cite{ruff2018,ghafoori2020deep,erfani2016high,chalapathy2018anomaly,yang2021}, probabilistic density modeling \cite{sohn2020learning,rippel2021modeling,an2015variational,zhai2016deep,zenati2018adversarially}, and reconstruction \cite{zhou2017anomaly,zong2018deep,chen2017outlier,aytekin2018clustering} have been widely studied revealing that anomaly detection performance is mostly determined by the quality of learned representations \cite{reiss2023anomaly} of the data. Improvements in self-supervised representation learning \cite{gidaris2018,doersch2015,noroozi2016,chen2020,grill2020,tian2020,caron2021emerging,he2022masked} have driven significant improvements in anomaly detection \cite{hojjati2022self,reiss2023anomaly}. Studies of both self-predictive \cite{li2021,golan2018deep,bergman2020classification,reiss2021panda} and contrastive \cite{li2021,tack2020csi,cho2021masked,sehwag2021ssd} self-supervised anomaly detection methods have shown that model performance depends on the quality of the pretext task and the definition of anomaly score; different types of anomalies (high-level semantic or low-level local defects) may benefit from different pretext tasks \cite{hojjati2022self,li2021}.

Multimodal machine learning has gained increasing attention \cite{bayoudh2021,guo2019,baltruvsaitis2018} mostly focusing around visual, natural language, and audio signals due their immense potential for real-world applications. Common learning strategies are: (i) separate feature extractors for different data modalities with varying architectural choices for alignment and fusion of learned representations \cite{alayrac2020,akbari2021vatt,alwassel2020,owens2018}; (ii) deep fusion Transformer encoders with cross-modal attention across different data modalities \cite{lu2019vilbert,tan2019lxmert,chen2020uniter,kim2021vilt}; and (iii) unification of learning objectives and neural architectures by systematically identifying and removing domain-specific inductive biases \cite{baevski2022data2vec,jaegle2021perceiver}. Self-supervised multimodal learning has various domain-specific applications such as medical imaging \cite{taleb2021} and robotics \cite{lee2019,meyer2020}. In contrast to aforementioned SSL techniques, multimodal methods, and augmentation strategies like \cite{devries2017,zhong2020,cubuk2020randaugment} that learn representations \textit{invariant} to augmentations, SeMAnD operates on fused image-like tensor representations of multimodal geospatial data that are not object-centric \cite{purushwalkam2020demystifying} to learn representations \textit{discriminative} to augmentations.

\section{Results}\label{results}
\textbf{Performance on Real Anomalies:} We obtain three manually labeled test sets (\cref{subsection::dataset_description}) of real geometry-based geospatial anomalies (i.e., both road network and/or RCPs could be anomalous) from different geographical regions distinct (in terms of distribution of tiles based on urban-rural stratification, landcover, population density, and activity stratas) from the geographies in the training dataset using stratified sampling \cite{cao2023self} to evaluate performance of trained anomaly detection models measured using AUC. Geometric anomalies are not expected to drastically differ across geographies and careful stratified sampling of training and test data helps to test the validity of this intuition. The test set size in regions 1, 2, and 3 are 927, 1065, and 930 respectively. \cref{table:accuracy_synthetic_and_real_anomalies} shows the model performance on real anomalies as the posedness ($\rho$) of RandPolyAugment generated training augmentations is varied with $\rho=0.10$ being optimal; model performance improves as $\rho$ increases until $\rho=0.10$, beyond which it degrades. All results reported use models trained with all data modalities and $\rho=0.10$.

\textbf{Ablation Study over Loss Components:} Exhaustive ablation studies on the weights of the loss components in \cref{total_loss} confirm the importance of the inverse-focal and contrastive losses. Using Bayesian hyperparameter optimization with 1000 trials and with the weight of each of the 3 components in \cref{total_loss} allowed to continuously vary in the range $[0,5]$, we obtained the optimal values $\lambda_{BC} = \lambda_{CL} = 1.0$, and $\lambda_{IF} = 1.5$. \cref{table:ablation_study_loss_component_weights} compares the anomaly detection performance (AUC averaged over all test regions; all data modalities used) of the optimal model with models trained with only one loss component. The optimal model gains 1.7--10.5\% compared to the models trained with only the individual components thereby demonstrating the effectiveness of the SeMAnD training objective.

\begin{table}
    \small
    \centering
    \begin{tabular}{X Y Y Y Y}
        \toprule
        \multirow{2}{1.8cm}{\centering Geographic Region} & \multicolumn{4}{c}{Anomaly Classification AUC} \\
        \cmidrule(lr){2-5}
                        &  $\rho=0.03$ & $\rho=0.05$ & $\rho=0.10$     & $\rho=0.15$                  \\
        \midrule
        Test Region 1   & 96.2\%       & 98.0\%      & \textbf{99.4\%} & 89.4\%                       \\
        Test Region 2   & 96.8\%       & 97.3\%      & \textbf{99.1\%} & 85.1\%                       \\
        Test Region 3   & 98.0\%       & 98.8\%      & \textbf{98.7\%} & 86.7\%                       \\
        \bottomrule
    \end{tabular}
    \bigskip
    \caption{Anomaly detection performance of models trained with all modalities with varying posedness ($\rho$) in 3 distinct test geographies. The value of $\rho=0.10$ is found to be optimal.}
    \label{table:accuracy_synthetic_and_real_anomalies}
\end{table}

\begin{table}
    \small
    \centering
    \begin{tabular}{E E E E E}
        \toprule
        $\lambda_{BC}$ & 1.0    & 0.0    & 0.0    & 1.0             \\
        $\lambda_{IF}$ & 0.0    & 1.0    & 0.0    & 1.5             \\
        $\lambda_{CL}$ & 0.0    & 0.0    & 1.0    & 1.0             \\
        \midrule
        Avg. AUC       & 95.8\% & 97.4\% & 88.6\% & \textbf{99.1\%} \\
        \bottomrule
    \end{tabular}
    \bigskip
    \caption{Selected experiments from an exhaustive ablation study over loss component weights in \cref{total_loss} to maximize anomaly detection performance. All data modalities are used. Reported AUC is averaged over all three test regions.}
    \label{table:ablation_study_loss_component_weights}
\end{table}

\begin{figure}[h]
    \centering
    \includegraphics[width=0.40\textwidth]{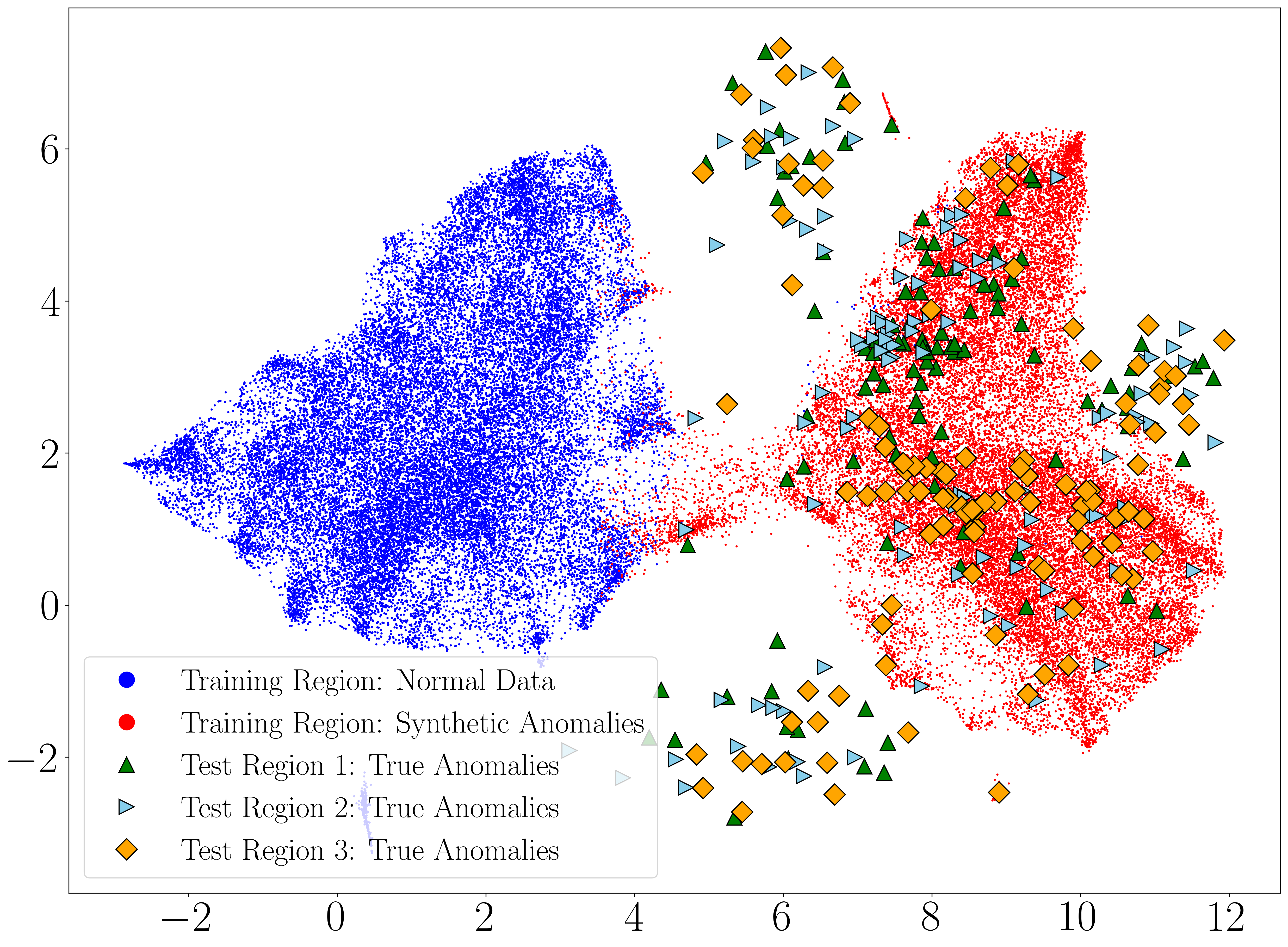}
    \caption{UMAP embeddings of normal (blue) and augmented (red) training instances superimposed with embeddings of real anomalies from the 3 test geographic regions.}
    \label{figure:umap}
\end{figure}

\textbf{Visualizing Learned Representations:} To explore the representations learned by the model, we plot in \cref{figure:umap} the 2D UMAP embeddings of the normal data in the training dataset (blue), the embeddings of one RandPolyAugment augmentation for each normal example (red), and the embeddings of 100 top scoring real anomalies from the three test geographies. While some of the anomalies fall within the space spanned by the augmentations, a significant number of anomalies are non-overlapping and far away from this space. This suggests that while RandPolyAugment produces augmentations that resemble some real anomalies, it does not cover all the possible ways in which unforeseen anomalies can occur. SeMAnD learns representations of normal data by clustering them into a smaller region in the latent space so that representations of augmentations and real anomalies lie outside of it. RandPolyAugment generates augmentations during training (similar to outlier exposure \cite{hendrycks2018}) in the same domain as the normal instance (similar to CutPaste \cite{li2021}) which enables the model to learn useful representations, $\boldsymbol{h}$, that also help the model generalize to unseen anomalies. The road network anomaly detected by SeMAnD in \cref{anomaly_5} or the junction changing to a roundabout anomaly detected by SeMAnD in \cref{anomaly_4} are examples of unseen anomaly types that RandPolyAugment cannot produce.

\textbf{Adding Data Modalities Improves Performance:} \cref{table:impact_of_adding_data_modalities} reports the AUC of trained models ($\rho=0.10$) on the test datasets of real anomalies when the models are trained with different combinations (all possible subsets) of the reference modalities (mobility data, satellite imagery, road network graph). \cref{table:impact_of_adding_data_modalities} quantitatively demonstrates that adding more data modalities (that provide complementary and supplementary information) to the input and jointly using them to learn semantically meaningful representations of a concept of normality (multimodal learning) enhances anomaly detection performance.

\begin{table}
    \small
    \centering
    \begin{tabular}{K J J J}
        \toprule
        \multirow{2}{1.3cm}{\centering Data Modalities} & \multicolumn{3}{c}{Anomaly Classification AUC}  \\
        \cmidrule(lr){2-4}
                   & Test Region 1   & Test Region 2   & Test Region 3   \\
        \midrule
        RNP        & 79.0\%          & 86.0\%          & 87.0\%          \\
        M          & 83.7\%          & 89.2\%          & 90.5\%          \\ 
        SI         & 85.7\%          & 90.5\%          & 91.9\%          \\ 
        RNP, M     & 87.2\%          & 91.5\%          & 93.0\%          \\
        RNP, SI    & 88.5\%          & 92.3\%          & 94.0\%          \\ 
        M, SI      & 89.6\%          & 93.1\%          & 94.8\%          \\ 
        RNP,M,SI & \textbf{99.4\%} & \textbf{99.1\%} & \textbf{98.7\%} \\
        \bottomrule
    \end{tabular}
    \bigskip
    \caption{Impact of different combinations of reference data modalities on model performance. M denotes mobility (both WCRM and DCRM), RNP denotes road network presence channel, and SI denotes satellite imagery.}
    \label{table:impact_of_adding_data_modalities}
\end{table}

\textbf{Diverse Augmentations Improve Performance:} Recent SSL approaches \cite{chen2020,grill2020} have shown that stronger and diverse augmentations lead to better learned representations. \cref{table:more_affine_transformations_are_better} shows a similar increase of performance of SeMAnD models ($\rho=0.10$, using all input modalities) trained with increasing cardinality ($|\boldsymbol{A}|$) of the allowed affine transformations set ($\boldsymbol{A}$) in \cref{algorithm:algorithm_randpolyaugment}. The cardinality of the subset ($|\boldsymbol{A}|$) varies from 1 to 4. For a given value of $|\boldsymbol{A}|$, the reported AUC is averaged over all $4 \choose |\boldsymbol{A}|$ combinations of translation, rotation, scaling, and deletion and over all three test regions.

\begin{table}[h]
    \small
    \centering
    \begin{tabular}{E E E E E}
        \toprule
                    & $|\boldsymbol{A}|=1$ & $|\boldsymbol{A}|=2$ & $|\boldsymbol{A}|=3$ & $|\boldsymbol{A}|=4$ \\
        \midrule
        Avg. AUC    & 76.2\%               & 85.8\%               & 92.4\%               & \textbf{99.1\%}      \\
        \bottomrule
    \end{tabular}
    \bigskip
    \caption{AUC of models trained with varying cardinality of the allowed affine transformations set ($\boldsymbol{A}$) in \cref{algorithm:algorithm_randpolyaugment}.}
    \label{table:more_affine_transformations_are_better}
\end{table}

\textbf{Comparison with Other Self-Supervision Methods}: \cref{table:randpolyaugment_comparison_with_other_baselines} presents anomaly detection performance comparison (AUC averaged over all three test regions) of SeMAnD (trained with RandPolyAugment) with multimodal models trained with other pretext tasks applied on the RCPP channel that learn representations by classifying rotation \cite{gidaris2018}, CutOut \cite{devries2017} , RandomErase \cite{zhong2020}, and CutPaste \cite{li2021}. The rotation strategy randomly rotates the RCPP channel by 0, $\pi/2, \pi$, or $3\pi/2$ with equal probability to generate the augmentation $\widetilde{\boldsymbol{\psi}}$ that is concatenated to the reference channels to generate $\widetilde{\boldsymbol{x}}$. Alternatively, the augmented RCPP channel, $\widetilde{\boldsymbol{\psi}}$, can be generated using the methodology outlined in CutOut \cite{devries2017}, RandomErase \cite{zhong2020}, or CutPaste \cite{li2021} which when concatenated to the reference channels yields the augmentated instance, $\widetilde{\boldsymbol{x}}$. Posedness ($\rho$) can be defined for each of the patch-based augmentation strategies. Results for the best choice of $\rho$ are reported for each strategy. SeMAnD outperforms the baseline approaches by 4.8--19.7 AUCs.

\begin{table}[h]
    \small
    \centering
    \begin{tabular}{V U}
        \toprule
        SSL Training Strategy            & Anomaly Classification AUC \\
        \midrule
        Rotation (Gidaris, et al, 2018)  & 79.4\%                     \\
        CutOut (Devries, et al, 2017)    & 83.1\%                     \\
        RandomErase (Zhong, et al, 2020) & 82.3\%                     \\
        CutPaste (Li, et al, 2021)       & 94.3\%                     \\
        SeMAnD (Ours)                    & \textbf{99.1\%}            \\
        \bottomrule
    \end{tabular}
    \bigskip
    \caption{Anomaly detection performance of SeMAnD and other SSL training strategies averaged over the three test sets.}
    \label{table:randpolyaugment_comparison_with_other_baselines}
\end{table}

While SeMAnD satisfactorily outperforms other methods, CutPaste remains a competitive self-supervised strategy. To understand why SeMAnD does better than CutPaste on geospatial anomalies, we create two new test sets with 30,000 augmentations each. Augmentations are generated with RandPolyAugment in one set and CutPaste in the other. \cref{table:comparison_synthetic_datasets} reports the anomaly detection AUC of models trained with CutPaste and SeMAnD on both these test sets. Training and test sets all use $\rho=0.10$. It is observed that both models are quite successful in detecting test anomalies generated with the same strategy as their training datasets. However, SeMAnD is more successful at detecting CutPaste augmentations (AUC of 92.9\%) compared to CutPaste's performance in detecting RandPolyAugment augmentations (AUC of 84.6\%). This suggests that the RandPolyAugment strategy, which is a geospatial domain-specific augmentation strategy, generalizes better for geospatial anomalies than CutPaste. Nevertheless, CutPaste performs remarkably well as an out-of-the-box, domain-agnostic, self-supervised anomaly detection strategy applicable to visual datasets in any domain.

\begin{table}[h]
    \small
    \centering
    \begin{tabular}{L M N}
        \toprule
        \multirow{2}{2.6cm}{\centering SSL Training Strategy} & \multicolumn{2}{c}{Anomaly Generation Strategy}\\
        \cmidrule(lr){2-3}
                                                              & {CutPaste} & {RandPolyAugment}                 \\
        \midrule
        CutPaste                                              & 99.5\%     & 84.6\%                            \\
        SeMAnD                                                & 92.9\%     & 99.7\%                            \\
        \bottomrule 
    \end{tabular}
    \bigskip
    \caption{Comparison of SeMAnD and CutPaste on augmentations generated using CutPaste and RandPolyAugment.}
    \label{table:comparison_synthetic_datasets}
\end{table}

\begin{figure}[h]
    \centering
    \begin{subfigure}{0.075\textwidth}
        \centering
        \includegraphics[width=\textwidth]{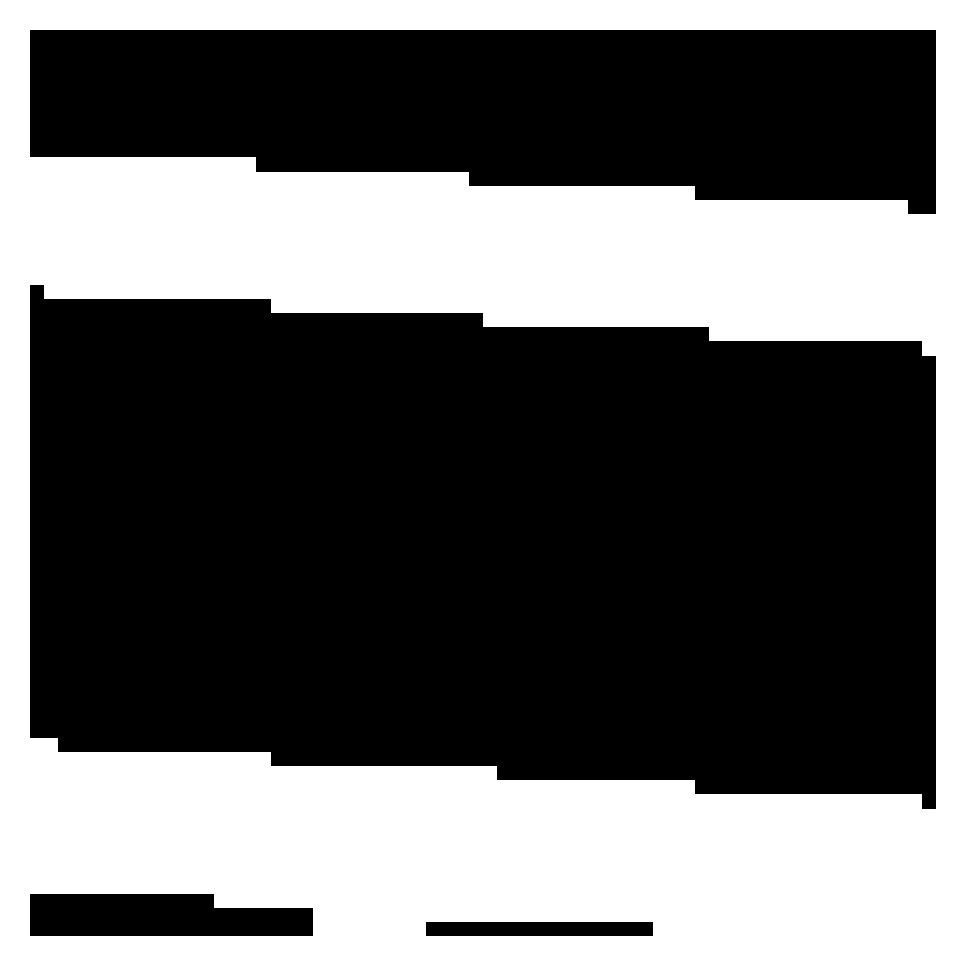}

        \includegraphics[width=\textwidth]{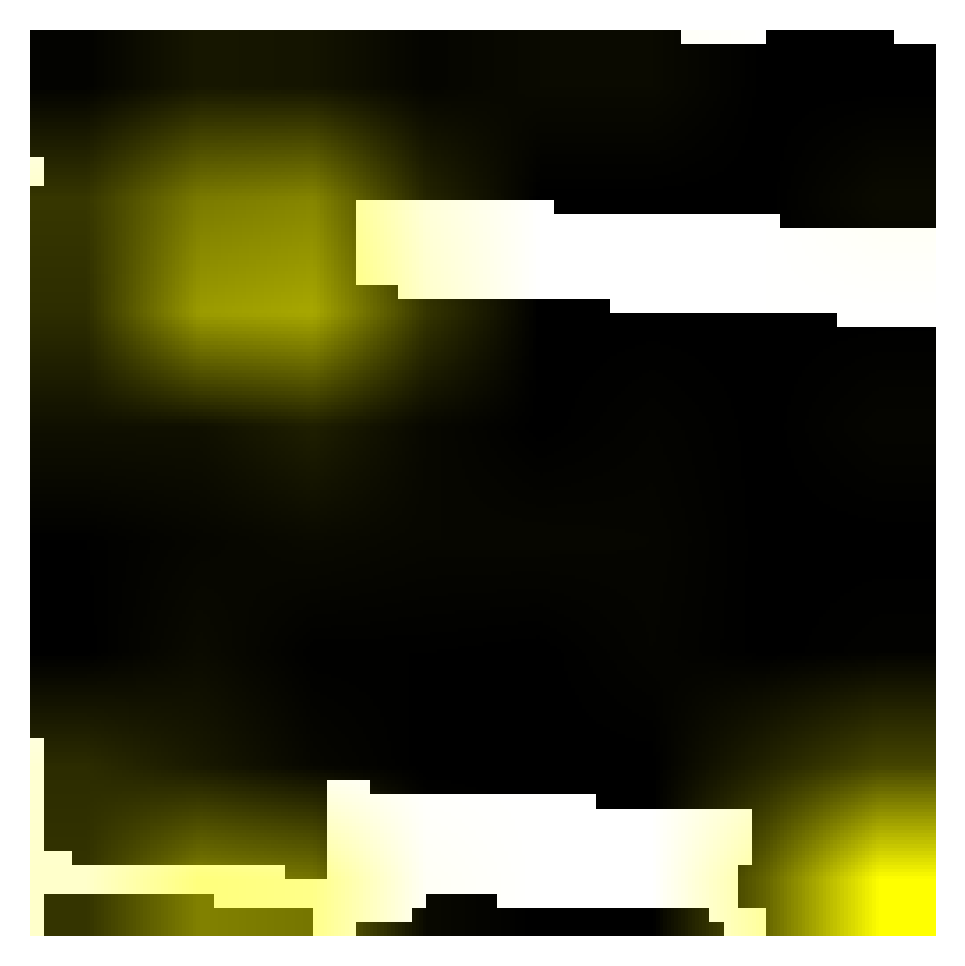}
        \caption{}
        \label{figure:gradcam_01}
    \end{subfigure}
    \begin{subfigure}{0.075\textwidth}
        \centering
        \includegraphics[width=\textwidth]{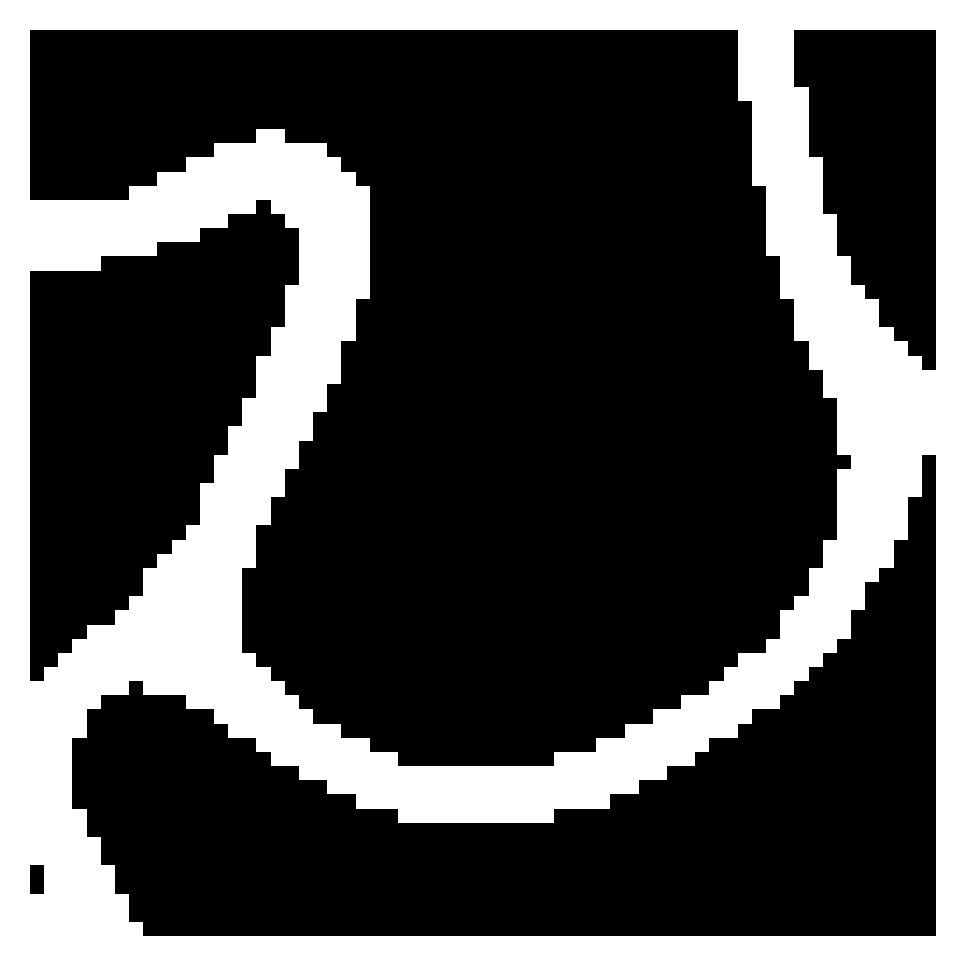}

        \includegraphics[width=\textwidth]{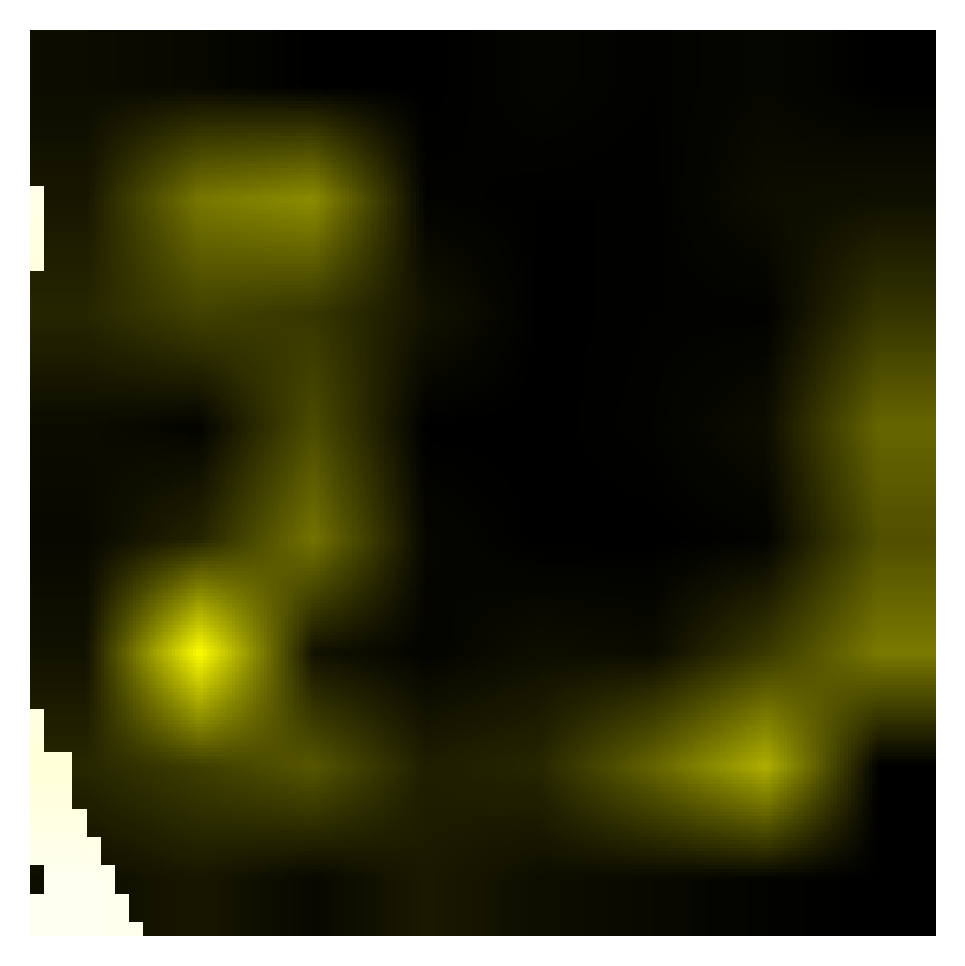}
        \caption{}
        \label{figure:gradcam_02}
    \end{subfigure}
    \begin{subfigure}{0.075\textwidth}
        \centering
        \includegraphics[width=\textwidth]{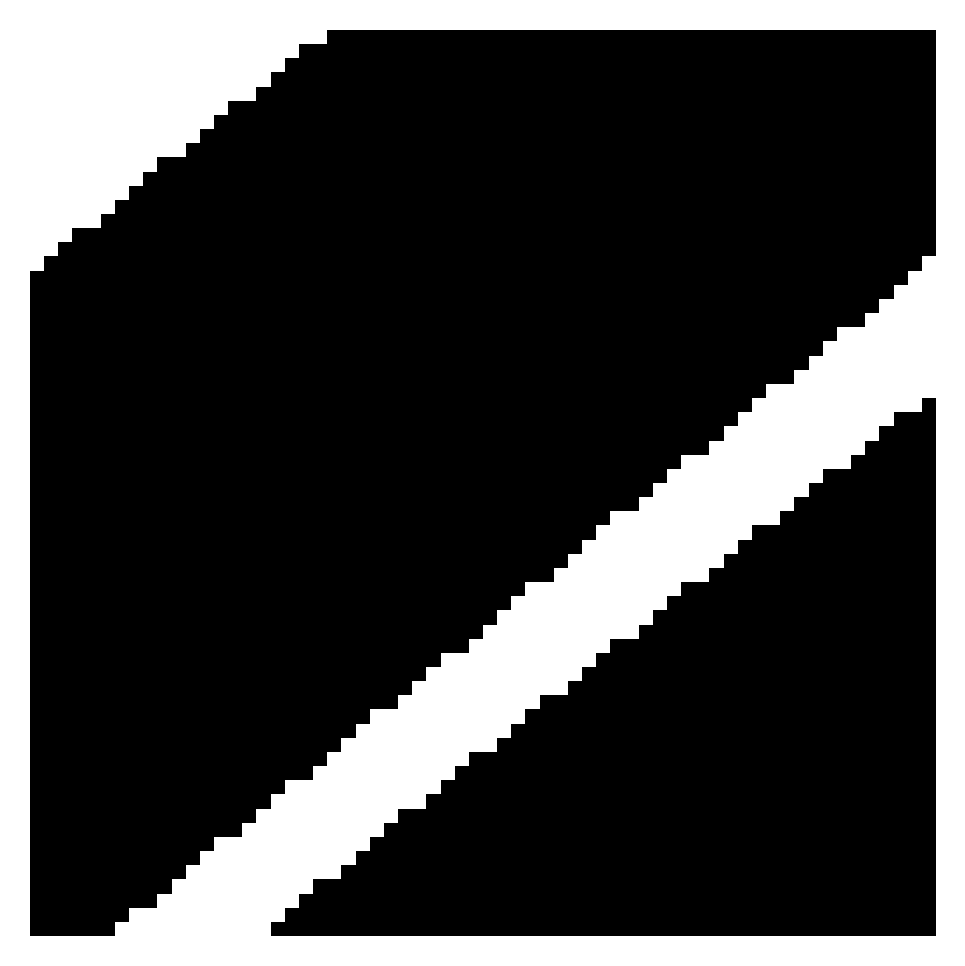}

        \includegraphics[width=\textwidth]{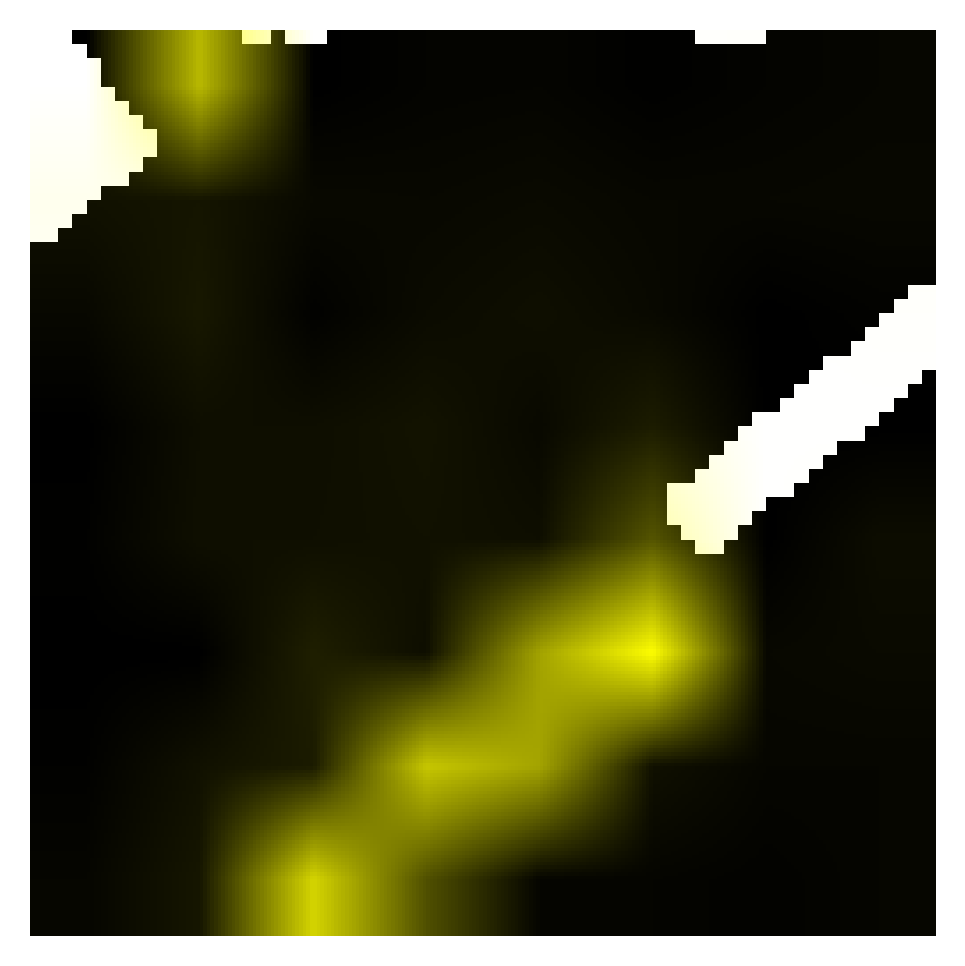}
        \caption{}
        \label{figure:gradcam_03}
    \end{subfigure}
    \begin{subfigure}{0.075\textwidth}
        \centering
        \includegraphics[width=\textwidth]{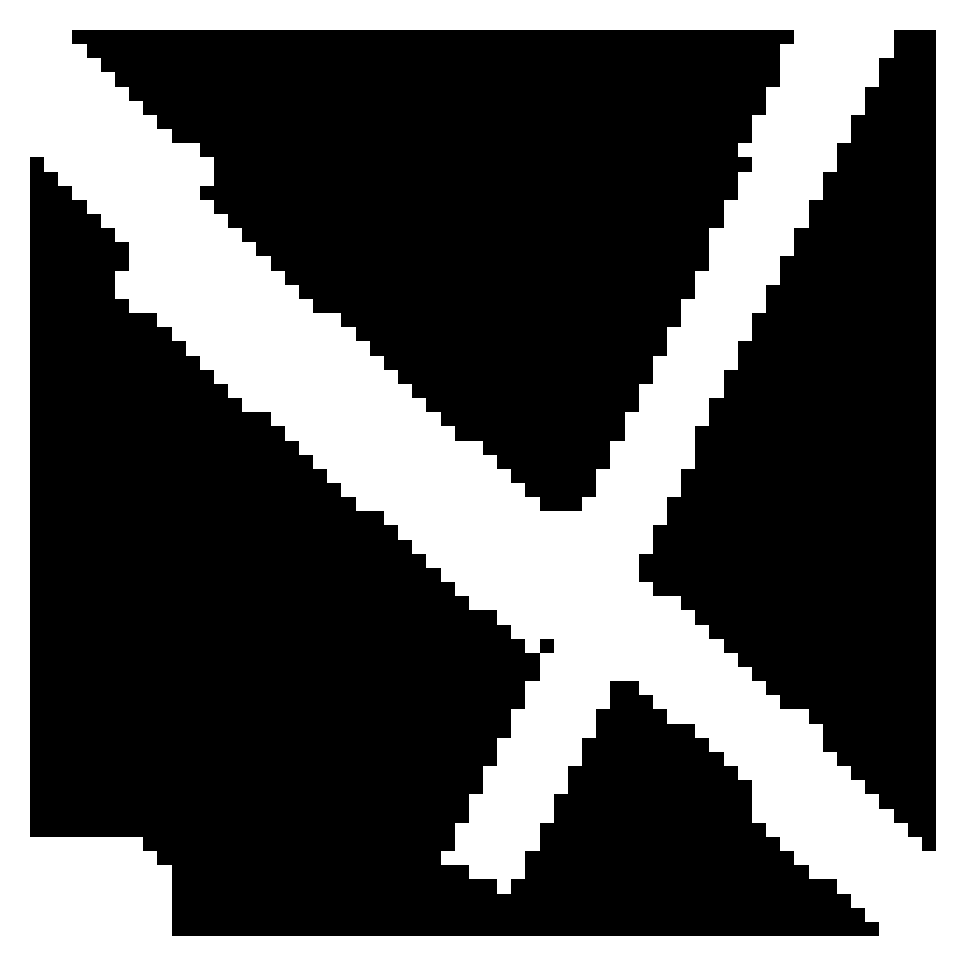}

        \includegraphics[width=\textwidth]{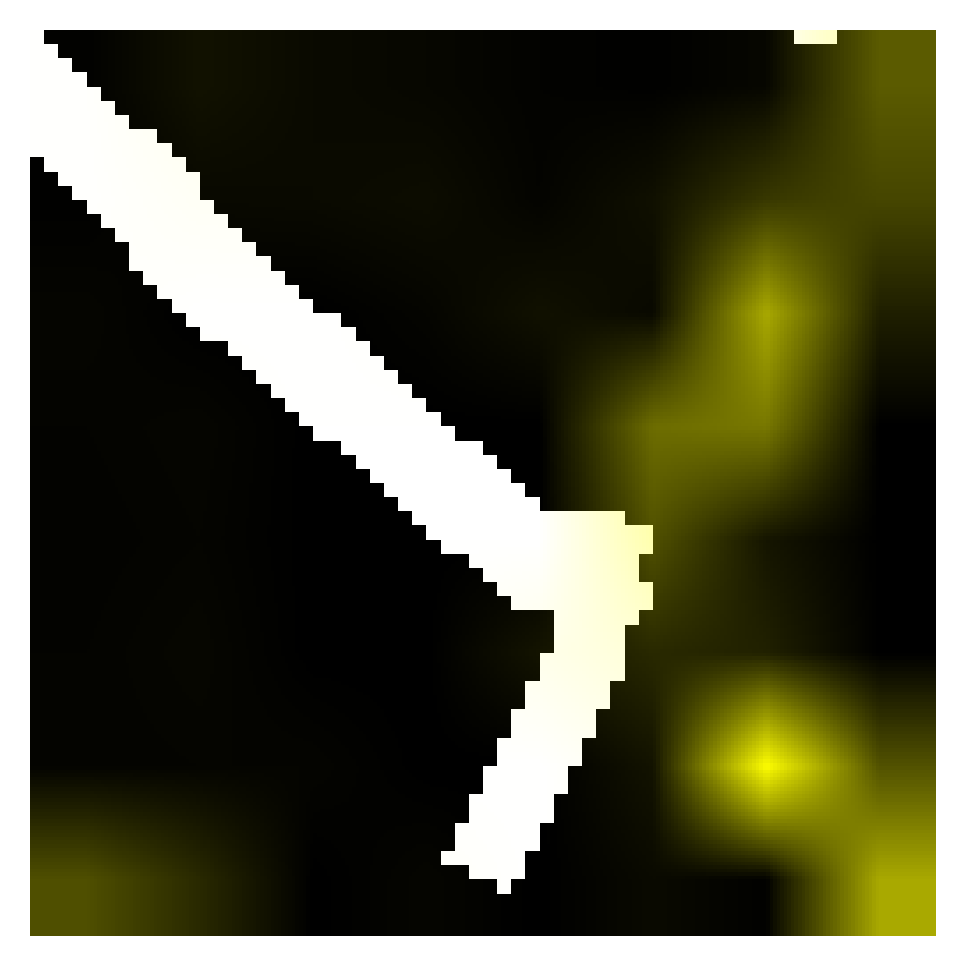}
        \caption{}
        \label{figure:gradcam_04}
    \end{subfigure}
    \begin{subfigure}{0.075\textwidth}
        \centering
        \includegraphics[width=\textwidth]{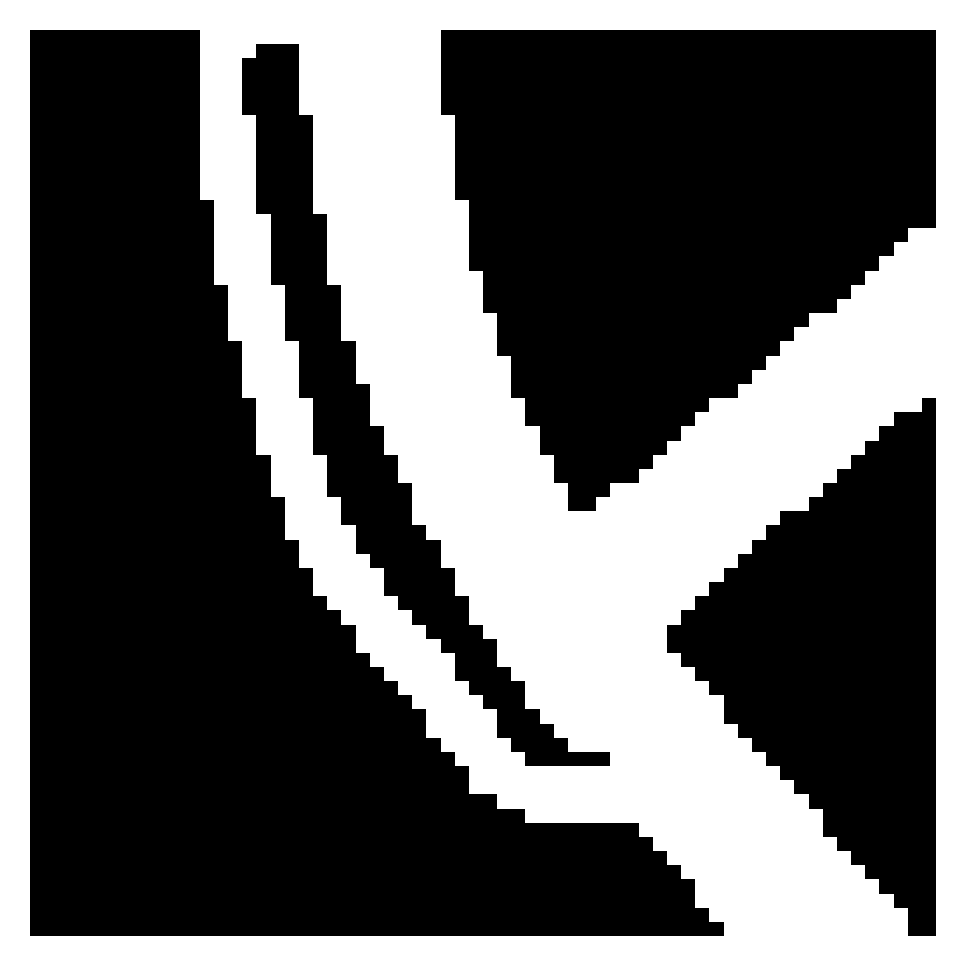}

        \includegraphics[width=\textwidth]{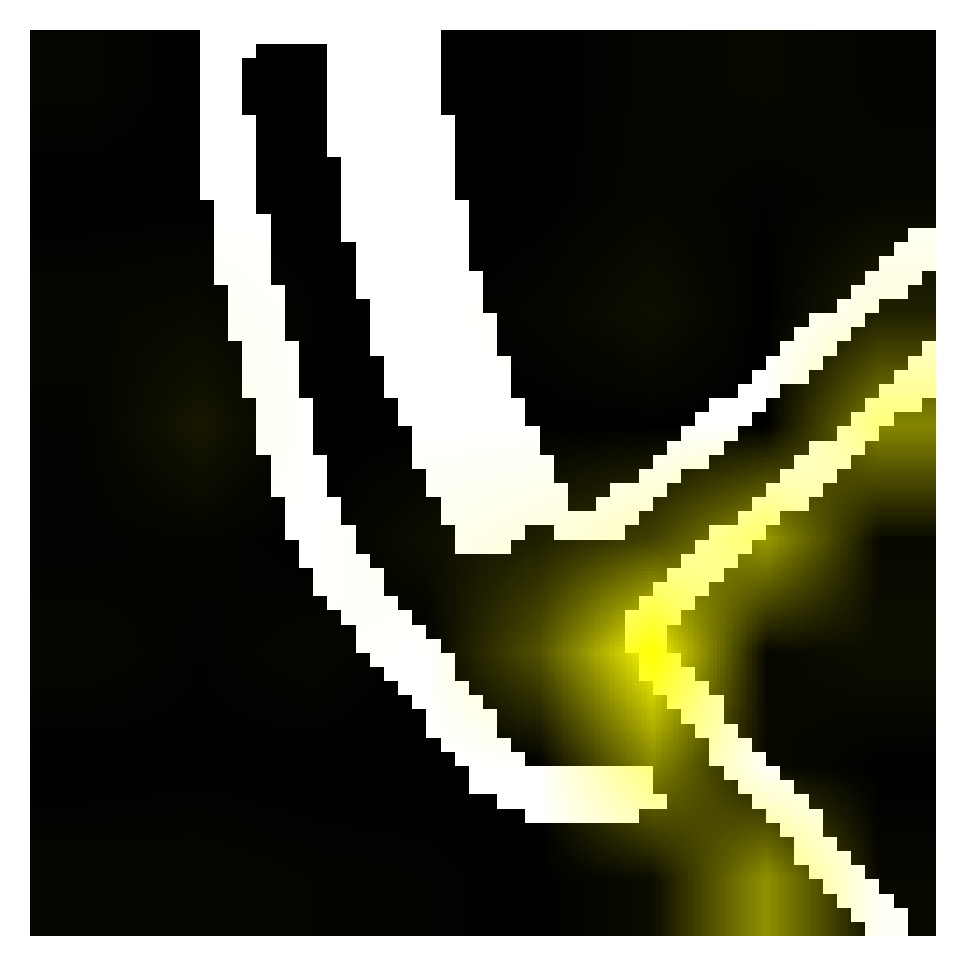}
        \caption{}
        \label{figure:gradcam_05}
    \end{subfigure}
    \begin{subfigure}{0.075\textwidth}
        \centering
        \includegraphics[width=\textwidth]{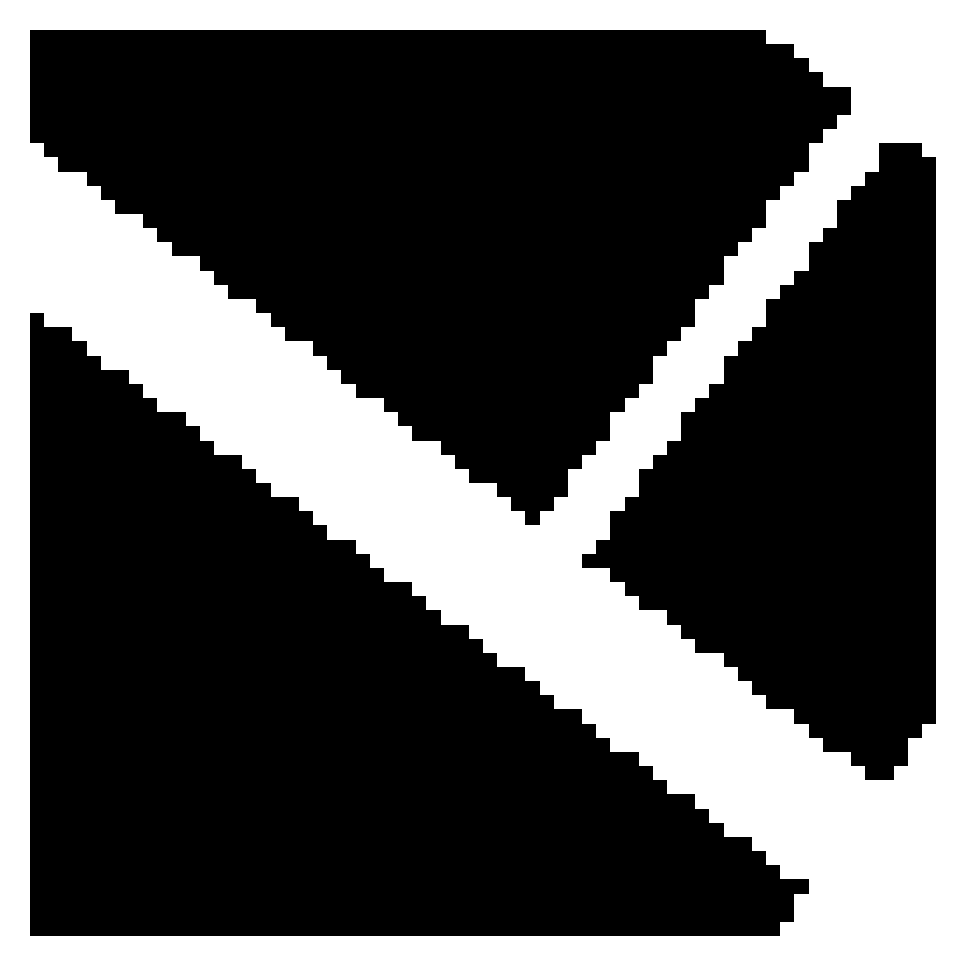}

        \includegraphics[width=\textwidth]{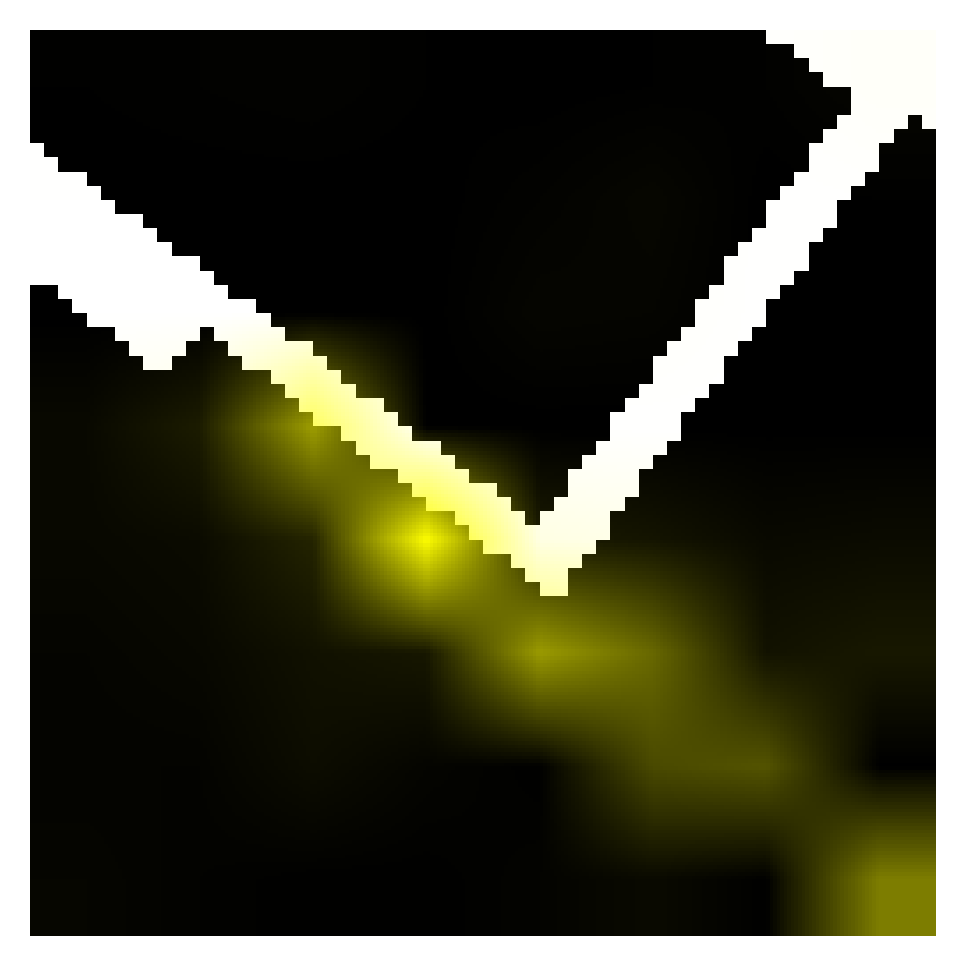}
        \caption{}
        \label{figure:gradcam_06}
    \end{subfigure}
    
    \caption{Examples of defect localization using GradCAM. Top row: normal RCPP. Bottom row: anomalous RCPP with superposed GradCAM heatmap (yellow).}
    \label{figure:examples_anomaly_localization}
\end{figure}

\textbf{Anomaly Localization:} As mentioned in \cref{subsection:anomaly_score_and_localization}, GradCAM is found to be effective in localizing anomalous regions in the input as exemplified in \cref{figure:examples_anomaly_localization}. In some cases where multiple anomalies are present, GradCAM does not always identify all of the anomalies but only some of them. There was no obvious preference pattern observed in the anomalies that GradCAM identified. It is hypothesized that when multiple anomalies are present, only a few are sufficient for the model to make a classification decision which is reflected in the learned representation and hence in the visual attention map generated by GradCAM.

\section{Anomaly Detection and Data Health Monitoring}
Quantitatively measuring and tracking regressions in multimodal geospatial datasets due to real-world changes is challenging. SeMAnD provides a comprehensive solution to address this challenge. SeMAnD models score every zoom-18 tile\footnote{Zoom-18 tiles have an edge length of approximately 152.83 m and cover approximately $2.34\times 10^{-2}$ sq. km. area on the equator.} in a given region (e.g., service regions or markets) and assigns a score between $[0,1]$, where 0 denotes normal and 1 denotes anomalous. The output of deployed SeMAnD models can therefore be utilized in two ways: (i) automated anomaly detection, and (ii) data health monitoring. In approach (i), all tiles that are associated with a score above a chosen threshold are flagged as anomalous for downstream review and fixing which facilitates tracking and fixing regressions in data due to real-world changes. In other words, SeMAnD provides an automated, low volume, high precision, and high recall framework for anomaly detection. In approach (ii), histograms of scores can be produced for the whole region or individual region stratas (e.g., downtown areas, residential areas, landmarks, high population density or activity areas) using frameworks such as \cite{cao2023self} on-demand or at a pre-defined cadence. This approach facilitates quantifying data quality, data freshness, and health across different regions or the same region at different times and at different granularities. SeMAnD is more cost-effective and time-effective compared to random sampling-based approaches since it allows targeted querying of anomalies and data regressions. SeMAnD is also a more holistic approach compared to rule-based checks due to its multimodal nature and can therefore complement or replace rule-based checks. When changes are made to existing data or new data is obtained from internal or external vendors, SeMAnD can be used as a quality assessment tool before the new data is ingested.

\begin{figure}[h]
    \centering
    \includegraphics[width=0.48\textwidth]{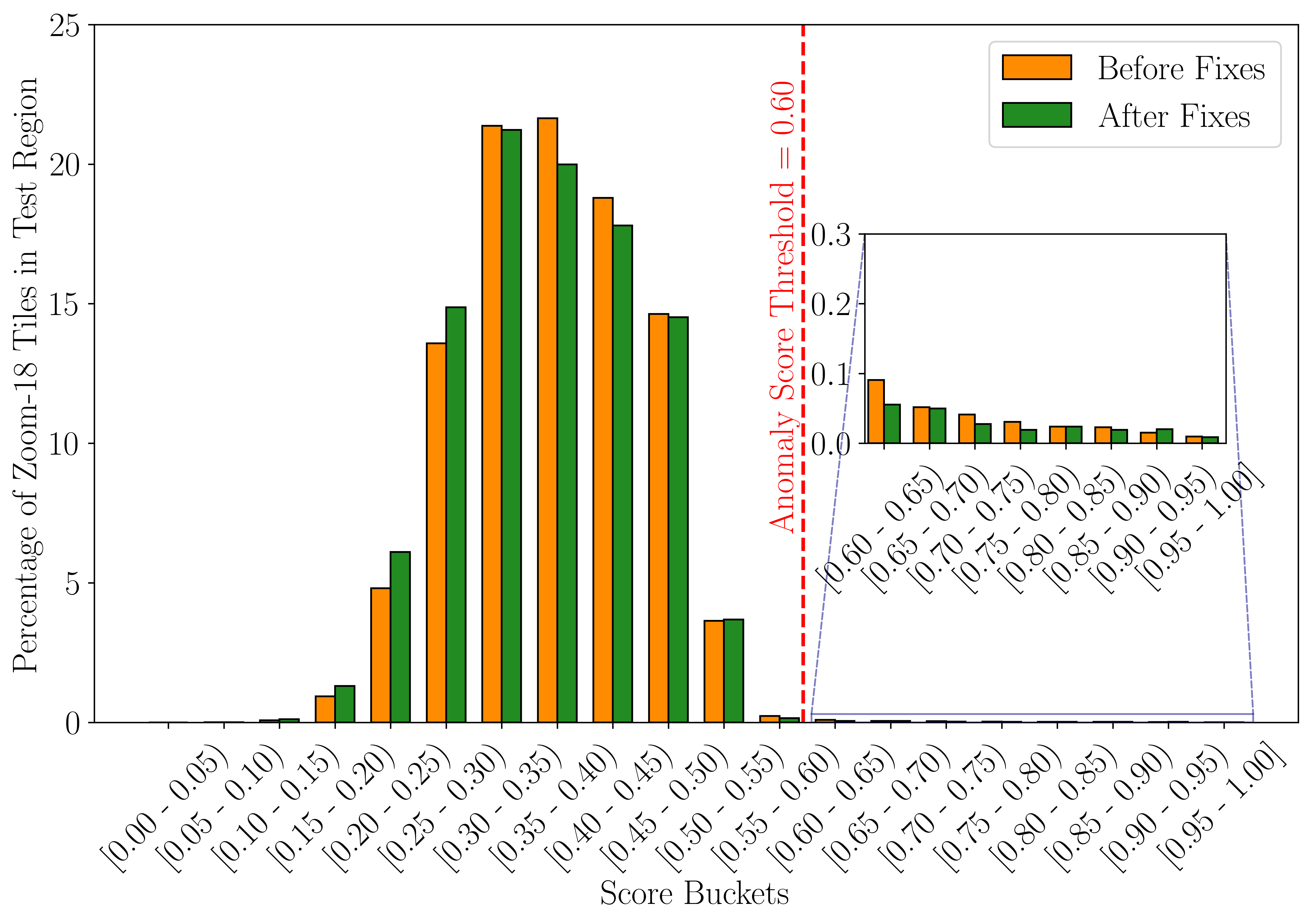}
    \caption{SeMAnD can be used to monitor the health of the multimodal geospatial data in a region. Shown here is the score distribution before and after fixes were made to address data regressions in a chosen test region. SeMAnD quantifies the improvement in the data (growth of green bars on the left and the associated reduction of orange bars on the right) and allows for targeted querying of the remaining anomalies (remaining green bars in the right tail).}
    \label{figure:region_health_statistics}
\end{figure}

Figure \ref{figure:region_health_statistics} exemplifies the deployment approaches proposed above. The figure compares the histogram of tile scores in a chosen test region (unseen during training) using proprietary data described in \cref{subsection::dataset_description} scored using SeMAnD. Due to multiple sources of real world changes (for example, road construction, road closures, new roads, etc.), there are regressions in the data which need to be fixed. The orange bars show the score distribution before the fixes while the green bars show the scores on the same region after the fixes. It is observed that:
\begin{itemize}
    \item Most of the region (99.78\%) is scored below the threshold of 0.6 which suggests that most of the data in the region is in good health and no major systemic issues exist in the data. 
    \item The green bars on the left side (closer to normal score of 0) are taller than the orange bars while the green bars on the right side (closer to anomalous score of 1) are shorter than the orange bars. This quantifies the extent to which the procedural fixes have addressed the data regression due to real world changes by fixing anomalous tiles and bringing their score closer to 0.
    \item Zooming into the right anomalous tail (inset of figure), we observe that even though the number of errors have reduced, some errors remain and further fixes need to be made in the locations identified by SeMAnD.
    \item Score distributions of the type shown in figure \ref{figure:region_health_statistics} can be computed for individual region stratas (e.g., downtown areas, residential areas, landmarks, high population density or activity areas) for a more detailed view of the health of the data in a region. 
\end{itemize}
This example demonstrates the potential of SeMAnD as a comprehensive, scalable and automated anomaly detection and data health monitoring solution.

\section{Conclusion}
This work demonstrates the effectiveness of SeMAnD, a self-supervised multimodal anomaly detection technique that combines (i) multimodal learning, (ii) RandPolyAugment which is an augmentation strategy capable of generating diverse augmentations, and (iii) meaningful self-supervised training objectives with self-predictive and contrastive components. SeMAnD shows superior performance in detecting real-world geometric anomalies across multiple diverse geographies. SeMAnD can be extended to include many other (reference and augmentable) data modalities as long as they can be transformed to image-like tensor representations. The success of SeMAnD highlights the importance of multimodal and self-supervised representation learning for anomaly detection.

\begin{acks}
{\small The authors sincerely thank Justin Spelbrink, Melissa Britton, Jason Borak, and Matt Lindsay for their continuous and constructive feedback during the development phase as well as helping with real anomalous data collection and human evaluation of trained models.}
\end{acks}

{\small
    \bibliographystyle{ACM-Reference-Format}
    \bibliography{Paper_ACMSIGSPATIAL2023}


\begin{thebibliography}{86}


\ifx \showCODEN    \undefined \def \showCODEN     #1{\unskip}     \fi
\ifx \showDOI      \undefined \def \showDOI       #1{#1}\fi
\ifx \showISBNx    \undefined \def \showISBNx     #1{\unskip}     \fi
\ifx \showISBNxiii \undefined \def \showISBNxiii  #1{\unskip}     \fi
\ifx \showISSN     \undefined \def \showISSN      #1{\unskip}     \fi
\ifx \showLCCN     \undefined \def \showLCCN      #1{\unskip}     \fi
\ifx \shownote     \undefined \def \shownote      #1{#1}          \fi
\ifx \showarticletitle \undefined \def \showarticletitle #1{#1}   \fi
\ifx \showURL      \undefined \def \showURL       {\relax}        \fi
\providecommand\bibfield[2]{#2}
\providecommand\bibinfo[2]{#2}
\providecommand\natexlab[1]{#1}
\providecommand\showeprint[2][]{arXiv:#2}

\bibitem[Abadi et~al\mbox{.}(2016)]%
        {abadi2016tensorflow}
\bibfield{author}{\bibinfo{person}{Mart{\'\i}n Abadi}, \bibinfo{person}{Paul Barham}, \bibinfo{person}{Jianmin Chen}, \bibinfo{person}{Zhifeng Chen}, \bibinfo{person}{Andy Davis}, \bibinfo{person}{Jeffrey Dean}, \bibinfo{person}{Matthieu Devin}, \bibinfo{person}{Sanjay Ghemawat}, \bibinfo{person}{Geoffrey Irving}, \bibinfo{person}{Michael Isard}, {et~al\mbox{.}}} \bibinfo{year}{2016}\natexlab{}.
\newblock \showarticletitle{{Tensorflow: A system for large-scale machine learning}}. In \bibinfo{booktitle}{\emph{Osdi}}, Vol.~\bibinfo{volume}{16}. Savannah, GA, USA, \bibinfo{pages}{265--283}.
\newblock


\bibitem[Akbari et~al\mbox{.}(2021)]%
        {akbari2021vatt}
\bibfield{author}{\bibinfo{person}{Hassan Akbari}, \bibinfo{person}{Liangzhe Yuan}, \bibinfo{person}{Rui Qian}, \bibinfo{person}{Wei-Hong Chuang}, \bibinfo{person}{Shih-Fu Chang}, \bibinfo{person}{Yin Cui}, {and} \bibinfo{person}{Boqing Gong}.} \bibinfo{year}{2021}\natexlab{}.
\newblock \showarticletitle{{VATT: Transformers for multimodal self-supervised learning from raw video, audio and text}}.
\newblock \bibinfo{journal}{\emph{Advances in Neural Information Processing Systems}}  \bibinfo{volume}{34} (\bibinfo{year}{2021}).
\newblock


\bibitem[Alayrac et~al\mbox{.}(2020)]%
        {alayrac2020}
\bibfield{author}{\bibinfo{person}{Jean-Baptiste Alayrac}, \bibinfo{person}{Adria Recasens}, \bibinfo{person}{Rosalia Schneider}, \bibinfo{person}{Relja Arandjelovic}, \bibinfo{person}{Jason Ramapuram}, \bibinfo{person}{Jeffrey De~Fauw}, \bibinfo{person}{Lucas Smaira}, \bibinfo{person}{Sander Dieleman}, {and} \bibinfo{person}{Andrew Zisserman}.} \bibinfo{year}{2020}\natexlab{}.
\newblock \showarticletitle{{Self-supervised multimodal versatile networks}}.
\newblock \bibinfo{journal}{\emph{Proceedings of the 34th International Conference on Neural Information Processing Systems}} \bibinfo{volume}{2}, \bibinfo{number}{6} (\bibinfo{year}{2020}), \bibinfo{pages}{7}.
\newblock


\bibitem[Alwassel et~al\mbox{.}(2020)]%
        {alwassel2020}
\bibfield{author}{\bibinfo{person}{Humam Alwassel}, \bibinfo{person}{Dhruv Mahajan}, \bibinfo{person}{Bruno Korbar}, \bibinfo{person}{Lorenzo Torresani}, \bibinfo{person}{Bernard Ghanem}, {and} \bibinfo{person}{Du Tran}.} \bibinfo{year}{2020}\natexlab{}.
\newblock \showarticletitle{{Self-supervised learning by cross-modal audio-video clustering}}.
\newblock \bibinfo{journal}{\emph{Advances in Neural Information Processing Systems}}  \bibinfo{volume}{33} (\bibinfo{year}{2020}).
\newblock


\bibitem[An and Cho(2015)]%
        {an2015variational}
\bibfield{author}{\bibinfo{person}{Jinwon An} {and} \bibinfo{person}{Sungzoon Cho}.} \bibinfo{year}{2015}\natexlab{}.
\newblock \showarticletitle{{Variational autoencoder based anomaly detection using reconstruction probability}}.
\newblock \bibinfo{journal}{\emph{Special lecture on IE}} \bibinfo{volume}{2}, \bibinfo{number}{1} (\bibinfo{year}{2015}), \bibinfo{pages}{1--18}.
\newblock


\bibitem[{Australian Research Data Commons}({[n.\,d.]})]%
        {aus_road_casement}
\bibfield{author}{\bibinfo{person}{{Australian Research Data Commons}}.} \bibinfo{year}{[n.\,d.]}\natexlab{}.
\newblock \bibinfo{title}{{Road casement polygons dataset}}.
\newblock \bibinfo{howpublished}{\url{https://researchdata.edu.au/road-casement-polygon-vicmap-property/1739655}}.
\newblock


\bibitem[Aytekin et~al\mbox{.}(2018)]%
        {aytekin2018clustering}
\bibfield{author}{\bibinfo{person}{Caglar Aytekin}, \bibinfo{person}{Xingyang Ni}, \bibinfo{person}{Francesco Cricri}, {and} \bibinfo{person}{Emre Aksu}.} \bibinfo{year}{2018}\natexlab{}.
\newblock \showarticletitle{{Clustering and unsupervised anomaly detection with l 2 normalized deep auto-encoder representations}}. In \bibinfo{booktitle}{\emph{2018 International Joint Conference on Neural Networks (IJCNN)}}. IEEE, \bibinfo{pages}{1--6}.
\newblock


\bibitem[Baevski et~al\mbox{.}(2022)]%
        {baevski2022data2vec}
\bibfield{author}{\bibinfo{person}{Alexei Baevski}, \bibinfo{person}{Wei-Ning Hsu}, \bibinfo{person}{Qiantong Xu}, \bibinfo{person}{Arun Babu}, \bibinfo{person}{Jiatao Gu}, {and} \bibinfo{person}{Michael Auli}.} \bibinfo{year}{2022}\natexlab{}.
\newblock \showarticletitle{{data2vec: A general framework for self-supervised learning in speech, vision and language}}.
\newblock \bibinfo{journal}{\emph{arXiv preprint arXiv:2202.03555}} (\bibinfo{year}{2022}).
\newblock


\bibitem[Baltru{\v{s}}aitis et~al\mbox{.}(2018)]%
        {baltruvsaitis2018}
\bibfield{author}{\bibinfo{person}{Tadas Baltru{\v{s}}aitis}, \bibinfo{person}{Chaitanya Ahuja}, {and} \bibinfo{person}{Louis-Philippe Morency}.} \bibinfo{year}{2018}\natexlab{}.
\newblock \showarticletitle{{Multimodal machine learning: A survey and taxonomy}}.
\newblock \bibinfo{journal}{\emph{IEEE transactions on pattern analysis and machine intelligence}} \bibinfo{volume}{41}, \bibinfo{number}{2} (\bibinfo{year}{2018}), \bibinfo{pages}{423--443}.
\newblock


\bibitem[Bayoudh et~al\mbox{.}(2021)]%
        {bayoudh2021}
\bibfield{author}{\bibinfo{person}{Khaled Bayoudh}, \bibinfo{person}{Raja Knani}, \bibinfo{person}{Fay{\c{c}}al Hamdaoui}, {and} \bibinfo{person}{Abdellatif Mtibaa}.} \bibinfo{year}{2021}\natexlab{}.
\newblock \showarticletitle{{A survey on deep multimodal learning for computer vision: Advances, trends, applications, and datasets}}.
\newblock \bibinfo{journal}{\emph{The Visual Computer}} (\bibinfo{year}{2021}), \bibinfo{pages}{1--32}.
\newblock


\bibitem[Bergman and Hoshen(2020)]%
        {bergman2020classification}
\bibfield{author}{\bibinfo{person}{Liron Bergman} {and} \bibinfo{person}{Yedid Hoshen}.} \bibinfo{year}{2020}\natexlab{}.
\newblock \showarticletitle{{Classification-based anomaly detection for general data}}.
\newblock \bibinfo{journal}{\emph{arXiv preprint arXiv:2005.02359}} (\bibinfo{year}{2020}).
\newblock


\bibitem[Cao et~al\mbox{.}(2023)]%
        {cao2023self}
\bibfield{author}{\bibinfo{person}{Yi Cao}, \bibinfo{person}{Swetava Ganguli}, {and} \bibinfo{person}{Vipul Pandey}.} \bibinfo{year}{2023}\natexlab{}.
\newblock \showarticletitle{{Self-supervised temporal analysis of spatiotemporal data}}.
\newblock \bibinfo{journal}{\emph{arXiv preprint arXiv:2304.13143}} (\bibinfo{year}{2023}).
\newblock


\bibitem[Caron et~al\mbox{.}(2021)]%
        {caron2021emerging}
\bibfield{author}{\bibinfo{person}{Mathilde Caron}, \bibinfo{person}{Hugo Touvron}, \bibinfo{person}{Ishan Misra}, \bibinfo{person}{Herv{\'e} J{\'e}gou}, \bibinfo{person}{Julien Mairal}, \bibinfo{person}{Piotr Bojanowski}, {and} \bibinfo{person}{Armand Joulin}.} \bibinfo{year}{2021}\natexlab{}.
\newblock \showarticletitle{{Emerging properties in self-supervised vision transformers}}. In \bibinfo{booktitle}{\emph{Proceedings of the IEEE/CVF international conference on computer vision}}. \bibinfo{pages}{9650--9660}.
\newblock


\bibitem[Chalapathy and Chawla(2019)]%
        {chalapathy2019}
\bibfield{author}{\bibinfo{person}{Raghavendra Chalapathy} {and} \bibinfo{person}{Sanjay Chawla}.} \bibinfo{year}{2019}\natexlab{}.
\newblock \showarticletitle{{Deep learning for anomaly detection: A survey}}.
\newblock \bibinfo{journal}{\emph{arXiv preprint arXiv:1901.03407}} (\bibinfo{year}{2019}).
\newblock


\bibitem[Chalapathy et~al\mbox{.}(2018)]%
        {chalapathy2018anomaly}
\bibfield{author}{\bibinfo{person}{Raghavendra Chalapathy}, \bibinfo{person}{Aditya~Krishna Menon}, {and} \bibinfo{person}{Sanjay Chawla}.} \bibinfo{year}{2018}\natexlab{}.
\newblock \showarticletitle{{Anomaly detection using one-class neural networks}}.
\newblock \bibinfo{journal}{\emph{arXiv preprint arXiv:1802.06360}} (\bibinfo{year}{2018}).
\newblock


\bibitem[Chandola et~al\mbox{.}(2009)]%
        {chandola2009anomaly}
\bibfield{author}{\bibinfo{person}{Varun Chandola}, \bibinfo{person}{Arindam Banerjee}, {and} \bibinfo{person}{Vipin Kumar}.} \bibinfo{year}{2009}\natexlab{}.
\newblock \showarticletitle{{Anomaly detection: A survey}}.
\newblock \bibinfo{journal}{\emph{ACM computing surveys (CSUR)}} \bibinfo{volume}{41}, \bibinfo{number}{3} (\bibinfo{year}{2009}), \bibinfo{pages}{1--58}.
\newblock


\bibitem[Chen et~al\mbox{.}(2017)]%
        {chen2017outlier}
\bibfield{author}{\bibinfo{person}{Jinghui Chen}, \bibinfo{person}{Saket Sathe}, \bibinfo{person}{Charu Aggarwal}, {and} \bibinfo{person}{Deepak Turaga}.} \bibinfo{year}{2017}\natexlab{}.
\newblock \showarticletitle{{Outlier detection with autoencoder ensembles}}. In \bibinfo{booktitle}{\emph{Proceedings of the 2017 SIAM international conference on data mining}}. SIAM, \bibinfo{pages}{90--98}.
\newblock


\bibitem[Chen et~al\mbox{.}(2020a)]%
        {chen2020}
\bibfield{author}{\bibinfo{person}{Ting Chen}, \bibinfo{person}{Simon Kornblith}, \bibinfo{person}{Mohammad Norouzi}, {and} \bibinfo{person}{Geoffrey Hinton}.} \bibinfo{year}{2020}\natexlab{a}.
\newblock \showarticletitle{{A simple framework for contrastive learning of visual representations}}. In \bibinfo{booktitle}{\emph{International conference on machine learning}}. PMLR, \bibinfo{pages}{1597--1607}.
\newblock


\bibitem[Chen et~al\mbox{.}(2020b)]%
        {chen2020uniter}
\bibfield{author}{\bibinfo{person}{Yen-Chun Chen}, \bibinfo{person}{Linjie Li}, \bibinfo{person}{Licheng Yu}, \bibinfo{person}{Ahmed El~Kholy}, \bibinfo{person}{Faisal Ahmed}, \bibinfo{person}{Zhe Gan}, \bibinfo{person}{Yu Cheng}, {and} \bibinfo{person}{Jingjing Liu}.} \bibinfo{year}{2020}\natexlab{b}.
\newblock \showarticletitle{{UNITER: Universal image-text representation learning}}. In \bibinfo{booktitle}{\emph{European conference on computer vision}}. Springer, \bibinfo{pages}{104--120}.
\newblock


\bibitem[Cho et~al\mbox{.}(2021)]%
        {cho2021masked}
\bibfield{author}{\bibinfo{person}{Hyunsoo Cho}, \bibinfo{person}{Jinseok Seol}, {and} \bibinfo{person}{Sang-goo Lee}.} \bibinfo{year}{2021}\natexlab{}.
\newblock \showarticletitle{{Masked contrastive learning for anomaly detection}}.
\newblock \bibinfo{journal}{\emph{arXiv preprint arXiv:2105.08793}} (\bibinfo{year}{2021}).
\newblock


\bibitem[Cubuk et~al\mbox{.}(2020)]%
        {cubuk2020randaugment}
\bibfield{author}{\bibinfo{person}{Ekin~D Cubuk}, \bibinfo{person}{Barret Zoph}, \bibinfo{person}{Jonathon Shlens}, {and} \bibinfo{person}{Quoc~V Le}.} \bibinfo{year}{2020}\natexlab{}.
\newblock \showarticletitle{{Randaugment: Practical automated data augmentation with a reduced search space}}. In \bibinfo{booktitle}{\emph{Proceedings of the IEEE/CVF conference on computer vision and pattern recognition workshops}}. \bibinfo{pages}{702--703}.
\newblock


\bibitem[DeVries and Taylor(2017)]%
        {devries2017}
\bibfield{author}{\bibinfo{person}{Terrance DeVries} {and} \bibinfo{person}{Graham~W Taylor}.} \bibinfo{year}{2017}\natexlab{}.
\newblock \showarticletitle{{Improved regularization of convolutional neural networks with cutout}}.
\newblock \bibinfo{journal}{\emph{arXiv preprint arXiv:1708.04552}} (\bibinfo{year}{2017}).
\newblock


\bibitem[Doersch et~al\mbox{.}(2015)]%
        {doersch2015}
\bibfield{author}{\bibinfo{person}{Carl Doersch}, \bibinfo{person}{Abhinav Gupta}, {and} \bibinfo{person}{Alexei~A Efros}.} \bibinfo{year}{2015}\natexlab{}.
\newblock \showarticletitle{{Unsupervised visual representation learning by context prediction}}. In \bibinfo{booktitle}{\emph{Proceedings of the IEEE international conference on computer vision}}. \bibinfo{pages}{1422--1430}.
\newblock


\bibitem[{EPSG Geodetic Parameter Registry}({[n.\,d.]})]%
        {epsg_3857}
\bibfield{author}{\bibinfo{person}{{EPSG Geodetic Parameter Registry}}.} \bibinfo{year}{[n.\,d.]}\natexlab{}.
\newblock \bibinfo{title}{{EPSG:3857}}.
\newblock \bibinfo{howpublished}{\url{https://epsg.io/3857}}.
\newblock


\bibitem[Erfani et~al\mbox{.}(2016)]%
        {erfani2016high}
\bibfield{author}{\bibinfo{person}{Sarah~M Erfani}, \bibinfo{person}{Sutharshan Rajasegarar}, \bibinfo{person}{Shanika Karunasekera}, {and} \bibinfo{person}{Christopher Leckie}.} \bibinfo{year}{2016}\natexlab{}.
\newblock \showarticletitle{{High-dimensional and large-scale anomaly detection using a linear one-class SVM with deep learning}}.
\newblock \bibinfo{journal}{\emph{Pattern Recognition}}  \bibinfo{volume}{58} (\bibinfo{year}{2016}), \bibinfo{pages}{121--134}.
\newblock


\bibitem[{ESRI}({[n.\,d.]})]%
        {esriuk_pavements_2020}
\bibfield{author}{\bibinfo{person}{{ESRI}}.} \bibinfo{year}{[n.\,d.]}\natexlab{}.
\newblock \bibinfo{title}{{Map of every pavement width in Great Britain helps local authorities with social distancing plans}}.
\newblock \bibinfo{howpublished}{\url{https://www.esriuk.com/en-gb/news/press-releases/uk/39-map-of-every-pavement-width-in-great-britain}}.
\newblock


\bibitem[{European Commission}({[n.\,d.]})]%
        {insprire_directive}
\bibfield{author}{\bibinfo{person}{{European Commission}}.} \bibinfo{year}{[n.\,d.]}\natexlab{}.
\newblock \bibinfo{title}{{INSPIRE: Infrastructure for Spatial Information in Europe, Cadastral parcels datasets}}.
\newblock \bibinfo{howpublished}{\url{https://inspire.ec.europa.eu/about-inspire/563}}.
\newblock


\bibitem[Foundation({[n.\,d.]})]%
        {spark}
\bibfield{author}{\bibinfo{person}{Apache~Software Foundation}.} \bibinfo{year}{[n.\,d.]}\natexlab{}.
\newblock \bibinfo{title}{{Apache Spark}}.
\newblock \bibinfo{howpublished}{\url{https://spark.apache.org/}}.
\newblock


\bibitem[Ganguli et~al\mbox{.}(2022)]%
        {ganguli2022reachability}
\bibfield{author}{\bibinfo{person}{Swetava Ganguli}, \bibinfo{person}{CV~Krishnakumar Iyer}, {and} \bibinfo{person}{Vipul Pandey}.} \bibinfo{year}{2022}\natexlab{}.
\newblock \showarticletitle{{Reachability Embeddings: Scalable self-supervised representation learning from mobility trajectories for multimodal geospatial computer vision}}. In \bibinfo{booktitle}{\emph{2022 23rd IEEE International Conference on Mobile Data Management (MDM)}}. IEEE, \bibinfo{pages}{44--53}.
\newblock


\bibitem[Ghafoori and Leckie(2020)]%
        {ghafoori2020deep}
\bibfield{author}{\bibinfo{person}{Zahra Ghafoori} {and} \bibinfo{person}{Christopher Leckie}.} \bibinfo{year}{2020}\natexlab{}.
\newblock \showarticletitle{{Deep multi-sphere support vector data description}}. In \bibinfo{booktitle}{\emph{Proceedings of the 2020 SIAM International Conference on Data Mining}}. SIAM, \bibinfo{pages}{109--117}.
\newblock


\bibitem[Gidaris et~al\mbox{.}(2018)]%
        {gidaris2018}
\bibfield{author}{\bibinfo{person}{Spyros Gidaris}, \bibinfo{person}{Praveer Singh}, {and} \bibinfo{person}{Nikos Komodakis}.} \bibinfo{year}{2018}\natexlab{}.
\newblock \showarticletitle{{Unsupervised representation learning by predicting image rotations}}.
\newblock \bibinfo{journal}{\emph{arXiv preprint arXiv:1803.07728}} (\bibinfo{year}{2018}).
\newblock


\bibitem[Golan and El-Yaniv(2018)]%
        {golan2018deep}
\bibfield{author}{\bibinfo{person}{Izhak Golan} {and} \bibinfo{person}{Ran El-Yaniv}.} \bibinfo{year}{2018}\natexlab{}.
\newblock \showarticletitle{{Deep anomaly detection using geometric transformations}}.
\newblock \bibinfo{journal}{\emph{Advances in neural information processing systems}}  \bibinfo{volume}{31} (\bibinfo{year}{2018}).
\newblock


\bibitem[Grill et~al\mbox{.}(2020)]%
        {grill2020}
\bibfield{author}{\bibinfo{person}{Jean-Bastien Grill}, \bibinfo{person}{Florian Strub}, \bibinfo{person}{Florent Altch{\'e}}, \bibinfo{person}{Corentin Tallec}, \bibinfo{person}{Pierre~H Richemond}, \bibinfo{person}{Elena Buchatskaya}, \bibinfo{person}{Carl Doersch}, \bibinfo{person}{Bernardo~Avila Pires}, \bibinfo{person}{Zhaohan~Daniel Guo}, \bibinfo{person}{Mohammad~Gheshlaghi Azar}, {et~al\mbox{.}}} \bibinfo{year}{2020}\natexlab{}.
\newblock \showarticletitle{{Bootstrap your own latent: A new approach to self-supervised learning}}.
\newblock \bibinfo{journal}{\emph{arXiv preprint arXiv:2006.07733}} (\bibinfo{year}{2020}).
\newblock


\bibitem[Guo et~al\mbox{.}(2019)]%
        {guo2019}
\bibfield{author}{\bibinfo{person}{Wenzhong Guo}, \bibinfo{person}{Jianwen Wang}, {and} \bibinfo{person}{Shiping Wang}.} \bibinfo{year}{2019}\natexlab{}.
\newblock \showarticletitle{{Deep multimodal representation learning: A survey}}.
\newblock \bibinfo{journal}{\emph{IEEE Access}}  \bibinfo{volume}{7} (\bibinfo{year}{2019}), \bibinfo{pages}{63373--63394}.
\newblock


\bibitem[He et~al\mbox{.}(2022)]%
        {he2022masked}
\bibfield{author}{\bibinfo{person}{Kaiming He}, \bibinfo{person}{Xinlei Chen}, \bibinfo{person}{Saining Xie}, \bibinfo{person}{Yanghao Li}, \bibinfo{person}{Piotr Doll{\'a}r}, {and} \bibinfo{person}{Ross Girshick}.} \bibinfo{year}{2022}\natexlab{}.
\newblock \showarticletitle{{Masked autoencoders are scalable vision learners}}. In \bibinfo{booktitle}{\emph{Proceedings of the IEEE/CVF Conference on Computer Vision and Pattern Recognition}}. \bibinfo{pages}{16000--16009}.
\newblock


\bibitem[He et~al\mbox{.}(2016)]%
        {he2016}
\bibfield{author}{\bibinfo{person}{Kaiming He}, \bibinfo{person}{Xiangyu Zhang}, \bibinfo{person}{Shaoqing Ren}, {and} \bibinfo{person}{Jian Sun}.} \bibinfo{year}{2016}\natexlab{}.
\newblock \showarticletitle{{Identity mappings in deep residual networks}}. In \bibinfo{booktitle}{\emph{European conference on computer vision}}. Springer, \bibinfo{pages}{630--645}.
\newblock


\bibitem[Hendrycks et~al\mbox{.}(2018)]%
        {hendrycks2018}
\bibfield{author}{\bibinfo{person}{Dan Hendrycks}, \bibinfo{person}{Mantas Mazeika}, {and} \bibinfo{person}{Thomas Dietterich}.} \bibinfo{year}{2018}\natexlab{}.
\newblock \showarticletitle{{Deep anomaly detection with outlier exposure}}. In \bibinfo{booktitle}{\emph{International Conference on Learning Representations}}.
\newblock


\bibitem[Hojjati et~al\mbox{.}(2022)]%
        {hojjati2022self}
\bibfield{author}{\bibinfo{person}{Hadi Hojjati}, \bibinfo{person}{Thi Kieu~Khanh Ho}, {and} \bibinfo{person}{Narges Armanfard}.} \bibinfo{year}{2022}\natexlab{}.
\newblock \showarticletitle{{Self-supervised anomaly detection: A survey and outlook}}.
\newblock \bibinfo{journal}{\emph{arXiv preprint arXiv:2205.05173}} (\bibinfo{year}{2022}).
\newblock


\bibitem[Iyer et~al\mbox{.}(2023)]%
        {iyer2023perspectives}
\bibfield{author}{\bibinfo{person}{CV~Krishnakumar Iyer}, \bibinfo{person}{Swetava Ganguli}, {and} \bibinfo{person}{Vipul Pandey}.} \bibinfo{year}{2023}\natexlab{}.
\newblock \showarticletitle{{Perspectives on geospatial artificial intelligence platforms for multimodal spatiotemporal datasets}}.
\newblock In \bibinfo{booktitle}{\emph{Advances in Scalable and Intelligent Geospatial Analytics}}. \bibinfo{publisher}{CRC Press}, \bibinfo{pages}{17--63}.
\newblock


\bibitem[Iyer et~al\mbox{.}(2021)]%
        {iyer2021trinity}
\bibfield{author}{\bibinfo{person}{CV~Krishnakumar Iyer}, \bibinfo{person}{Feili Hou}, \bibinfo{person}{Henry Wang}, \bibinfo{person}{Yonghong Wang}, \bibinfo{person}{Kay Oh}, \bibinfo{person}{Swetava Ganguli}, {and} \bibinfo{person}{Vipul Pandey}.} \bibinfo{year}{2021}\natexlab{}.
\newblock \showarticletitle{{Trinity: A No-Code AI platform for complex spatial datasets}}. In \bibinfo{booktitle}{\emph{Proceedings of the 4th ACM SIGSPATIAL International Workshop on AI for Geographic Knowledge Discovery}}. \bibinfo{pages}{33--42}.
\newblock


\bibitem[Jaegle et~al\mbox{.}(2021)]%
        {jaegle2021perceiver}
\bibfield{author}{\bibinfo{person}{Andrew Jaegle}, \bibinfo{person}{Felix Gimeno}, \bibinfo{person}{Andy Brock}, \bibinfo{person}{Oriol Vinyals}, \bibinfo{person}{Andrew Zisserman}, {and} \bibinfo{person}{Joao Carreira}.} \bibinfo{year}{2021}\natexlab{}.
\newblock \showarticletitle{{Perceiver: General perception with iterative attention}}. In \bibinfo{booktitle}{\emph{International Conference on Machine Learning}}. PMLR, \bibinfo{pages}{4651--4664}.
\newblock


\bibitem[Kim et~al\mbox{.}(2021)]%
        {kim2021vilt}
\bibfield{author}{\bibinfo{person}{Wonjae Kim}, \bibinfo{person}{Bokyung Son}, {and} \bibinfo{person}{Ildoo Kim}.} \bibinfo{year}{2021}\natexlab{}.
\newblock \showarticletitle{{ViLT: Vision-and-language transformer without convolution or region supervision}}. In \bibinfo{booktitle}{\emph{International Conference on Machine Learning}}. PMLR, \bibinfo{pages}{5583--5594}.
\newblock


\bibitem[Kolesnikov et~al\mbox{.}(2020)]%
        {kolesnikov2020big}
\bibfield{author}{\bibinfo{person}{Alexander Kolesnikov}, \bibinfo{person}{Lucas Beyer}, \bibinfo{person}{Xiaohua Zhai}, \bibinfo{person}{Joan Puigcerver}, \bibinfo{person}{Jessica Yung}, \bibinfo{person}{Sylvain Gelly}, {and} \bibinfo{person}{Neil Houlsby}.} \bibinfo{year}{2020}\natexlab{}.
\newblock \showarticletitle{{Big transfer (BiT): General visual representation learning}}. In \bibinfo{booktitle}{\emph{Computer Vision--ECCV 2020: 16th European Conference, Glasgow, UK, August 23--28, 2020, Proceedings, Part V 16}}. Springer, \bibinfo{pages}{491--507}.
\newblock


\bibitem[Lee et~al\mbox{.}(2019)]%
        {lee2019}
\bibfield{author}{\bibinfo{person}{Michelle~A Lee}, \bibinfo{person}{Yuke Zhu}, \bibinfo{person}{Krishnan Srinivasan}, \bibinfo{person}{Parth Shah}, \bibinfo{person}{Silvio Savarese}, \bibinfo{person}{Li Fei-Fei}, \bibinfo{person}{Animesh Garg}, {and} \bibinfo{person}{Jeannette Bohg}.} \bibinfo{year}{2019}\natexlab{}.
\newblock \showarticletitle{{Making sense of vision and touch: Self-supervised learning of multimodal representations for contact-rich tasks}}. In \bibinfo{booktitle}{\emph{2019 International Conference on Robotics and Automation (ICRA)}}. IEEE, \bibinfo{pages}{8943--8950}.
\newblock


\bibitem[Li et~al\mbox{.}(2021)]%
        {li2021}
\bibfield{author}{\bibinfo{person}{Chun-Liang Li}, \bibinfo{person}{Kihyuk Sohn}, \bibinfo{person}{Jinsung Yoon}, {and} \bibinfo{person}{Tomas Pfister}.} \bibinfo{year}{2021}\natexlab{}.
\newblock \showarticletitle{{CutPaste: Self-supervised learning for anomaly detection and localization}}. In \bibinfo{booktitle}{\emph{Proceedings of the IEEE/CVF Conference on Computer Vision and Pattern Recognition}}. \bibinfo{pages}{9664--9674}.
\newblock


\bibitem[Lin et~al\mbox{.}(2017)]%
        {lin2017}
\bibfield{author}{\bibinfo{person}{Tsung-Yi Lin}, \bibinfo{person}{Priya Goyal}, \bibinfo{person}{Ross Girshick}, \bibinfo{person}{Kaiming He}, {and} \bibinfo{person}{Piotr Doll{\'a}r}.} \bibinfo{year}{2017}\natexlab{}.
\newblock \showarticletitle{{Focal loss for dense object detection}}. In \bibinfo{booktitle}{\emph{Proceedings of the IEEE international conference on computer vision}}. \bibinfo{pages}{2980--2988}.
\newblock


\bibitem[Long et~al\mbox{.}(2017)]%
        {long2017}
\bibfield{author}{\bibinfo{person}{Mingsheng Long}, \bibinfo{person}{Zhangjie Cao}, \bibinfo{person}{Jianmin Wang}, {and} \bibinfo{person}{Michael Jordan}.} \bibinfo{year}{2017}\natexlab{}.
\newblock \showarticletitle{{Domain adaptation with randomized multilinear adversarial networks}}.
\newblock \bibinfo{journal}{\emph{ResearchGate}} (\bibinfo{date}{05} \bibinfo{year}{2017}).
\newblock


\bibitem[Loshchilov and Hutter(2017)]%
        {loshchilov2017decoupled}
\bibfield{author}{\bibinfo{person}{Ilya Loshchilov} {and} \bibinfo{person}{Frank Hutter}.} \bibinfo{year}{2017}\natexlab{}.
\newblock \showarticletitle{{Decoupled weight decay regularization}}.
\newblock \bibinfo{journal}{\emph{arXiv preprint arXiv:1711.05101}} (\bibinfo{year}{2017}).
\newblock


\bibitem[Lu et~al\mbox{.}(2019)]%
        {lu2019vilbert}
\bibfield{author}{\bibinfo{person}{Jiasen Lu}, \bibinfo{person}{Dhruv Batra}, \bibinfo{person}{Devi Parikh}, {and} \bibinfo{person}{Stefan Lee}.} \bibinfo{year}{2019}\natexlab{}.
\newblock \showarticletitle{{ViLBERT: Pretraining task-agnostic visiolinguistic representations for vision-and-language tasks}}. In \bibinfo{booktitle}{\emph{Advances in Neural Information Processing Systems}}, Vol.~\bibinfo{volume}{32}.
\newblock


\bibitem[Meyer et~al\mbox{.}(2020)]%
        {meyer2020}
\bibfield{author}{\bibinfo{person}{Johannes Meyer}, \bibinfo{person}{Andreas Eitel}, \bibinfo{person}{Thomas Brox}, {and} \bibinfo{person}{Wolfram Burgard}.} \bibinfo{year}{2020}\natexlab{}.
\newblock \showarticletitle{{Improving unimodal object recognition with multimodal contrastive learning}}. In \bibinfo{booktitle}{\emph{2020 IEEE/RSJ International Conference on Intelligent Robots and Systems (IROS)}}. IEEE, \bibinfo{pages}{5656--5663}.
\newblock


\bibitem[Moreira-Matias et~al\mbox{.}(2013)]%
        {ucigpsdataset13}
\bibfield{author}{\bibinfo{person}{L. Moreira-Matias} {et~al\mbox{.}}} \bibinfo{year}{2013}\natexlab{}.
\newblock \showarticletitle{{Predicting taxi–passenger demand using streaming data}}.
\newblock \bibinfo{journal}{\emph{IEEE Trans. Intelligent Transportation Sys.}}  \bibinfo{volume}{14} (\bibinfo{year}{2013}), \bibinfo{pages}{1393--1402}.
\newblock


\bibitem[Noroozi and Favaro(2016)]%
        {noroozi2016}
\bibfield{author}{\bibinfo{person}{Mehdi Noroozi} {and} \bibinfo{person}{Paolo Favaro}.} \bibinfo{year}{2016}\natexlab{}.
\newblock \showarticletitle{{Unsupervised learning of visual representations by solving jigsaw puzzles}}. In \bibinfo{booktitle}{\emph{European conference on computer vision}}. Springer, \bibinfo{pages}{69--84}.
\newblock


\bibitem[O'Flaherty(2018)]%
        {oflaherty2018}
\bibfield{author}{\bibinfo{person}{Coleman~A O'Flaherty}.} \bibinfo{year}{2018}\natexlab{}.
\newblock \bibinfo{booktitle}{\emph{{Transport planning and traffic engineering}}}.
\newblock \bibinfo{publisher}{CRC Press}.
\newblock


\bibitem[Oord et~al\mbox{.}(2018)]%
        {oord2018}
\bibfield{author}{\bibinfo{person}{Aaron van~den Oord}, \bibinfo{person}{Yazhe Li}, {and} \bibinfo{person}{Oriol Vinyals}.} \bibinfo{year}{2018}\natexlab{}.
\newblock \showarticletitle{{Representation learning with contrastive predictive coding}}.
\newblock \bibinfo{journal}{\emph{arXiv preprint arXiv:1807.03748}} (\bibinfo{year}{2018}).
\newblock


\bibitem[OpenStreetMap({[n.\,d.]})]%
        {osmgpstraces21}
\bibfield{author}{\bibinfo{person}{OpenStreetMap}.} \bibinfo{year}{[n.\,d.]}\natexlab{}.
\newblock \bibinfo{title}{{Public GPS Traces}}.
\newblock \bibinfo{howpublished}{\url{https://www.openstreetmap.org/traces}}.
\newblock


\bibitem[Owens and Efros(2018)]%
        {owens2018}
\bibfield{author}{\bibinfo{person}{Andrew Owens} {and} \bibinfo{person}{Alexei~A Efros}.} \bibinfo{year}{2018}\natexlab{}.
\newblock \showarticletitle{{Audio-visual scene analysis with self-supervised multisensory features}}. In \bibinfo{booktitle}{\emph{Proceedings of the European Conference on Computer Vision (ECCV)}}. \bibinfo{pages}{631--648}.
\newblock


\bibitem[Pang et~al\mbox{.}(2021)]%
        {pang2021}
\bibfield{author}{\bibinfo{person}{Guansong Pang}, \bibinfo{person}{Chunhua Shen}, \bibinfo{person}{Longbing Cao}, {and} \bibinfo{person}{Anton Van~Den Hengel}.} \bibinfo{year}{2021}\natexlab{}.
\newblock \showarticletitle{{Deep learning for anomaly detection: A review}}.
\newblock \bibinfo{journal}{\emph{ACM Computing Surveys (CSUR)}} \bibinfo{volume}{54}, \bibinfo{number}{2} (\bibinfo{year}{2021}), \bibinfo{pages}{1--38}.
\newblock


\bibitem[Panzarino(2018)]%
        {tcrunch18}
\bibfield{author}{\bibinfo{person}{Matthew Panzarino}.} \bibinfo{year}{2018}\natexlab{}.
\newblock \showarticletitle{{Apple is rebuilding Maps from the ground up}}.
\newblock \bibinfo{journal}{\emph{TechCrunch}} (\bibinfo{year}{2018}).
\newblock


\bibitem[Purushwalkam and Gupta(2020)]%
        {purushwalkam2020demystifying}
\bibfield{author}{\bibinfo{person}{Senthil Purushwalkam} {and} \bibinfo{person}{Abhinav Gupta}.} \bibinfo{year}{2020}\natexlab{}.
\newblock \showarticletitle{{Demystifying contrastive self-supervised learning: Invariances, augmentations and dataset biases}}.
\newblock \bibinfo{journal}{\emph{Advances in Neural Information Processing Systems}}  \bibinfo{volume}{33} (\bibinfo{year}{2020}), \bibinfo{pages}{3407--3418}.
\newblock


\bibitem[Reiss et~al\mbox{.}(2021)]%
        {reiss2021panda}
\bibfield{author}{\bibinfo{person}{Tal Reiss}, \bibinfo{person}{Niv Cohen}, \bibinfo{person}{Liron Bergman}, {and} \bibinfo{person}{Yedid Hoshen}.} \bibinfo{year}{2021}\natexlab{}.
\newblock \showarticletitle{{Panda: Adapting pretrained features for anomaly detection and segmentation}}. In \bibinfo{booktitle}{\emph{Proceedings of the IEEE/CVF Conference on Computer Vision and Pattern Recognition}}. \bibinfo{pages}{2806--2814}.
\newblock


\bibitem[Reiss et~al\mbox{.}(2023)]%
        {reiss2023anomaly}
\bibfield{author}{\bibinfo{person}{Tal Reiss}, \bibinfo{person}{Niv Cohen}, \bibinfo{person}{Eliahu Horwitz}, \bibinfo{person}{Ron Abutbul}, {and} \bibinfo{person}{Yedid Hoshen}.} \bibinfo{year}{2023}\natexlab{}.
\newblock \showarticletitle{{Anomaly detection requires better representations}}. In \bibinfo{booktitle}{\emph{Computer Vision--ECCV 2022 Workshops: Tel Aviv, Israel, October 23--27, 2022, Proceedings, Part IV}}. Springer, \bibinfo{pages}{56--68}.
\newblock


\bibitem[Rippel et~al\mbox{.}(2021a)]%
        {rippel2021}
\bibfield{author}{\bibinfo{person}{Oliver Rippel}, \bibinfo{person}{Patrick Mertens}, {and} \bibinfo{person}{Dorit Merhof}.} \bibinfo{year}{2021}\natexlab{a}.
\newblock \showarticletitle{{Modeling the distribution of normal data in pre-trained deep features for anomaly detection}}. In \bibinfo{booktitle}{\emph{25th International Conference on Pattern Recognition (ICPR)}}. IEEE, \bibinfo{pages}{6726--6733}.
\newblock


\bibitem[Rippel et~al\mbox{.}(2021b)]%
        {rippel2021modeling}
\bibfield{author}{\bibinfo{person}{Oliver Rippel}, \bibinfo{person}{Patrick Mertens}, {and} \bibinfo{person}{Dorit Merhof}.} \bibinfo{year}{2021}\natexlab{b}.
\newblock \showarticletitle{{Modeling the distribution of normal data in pre-trained deep features for anomaly detection}}. In \bibinfo{booktitle}{\emph{2020 25th International Conference on Pattern Recognition (ICPR)}}. IEEE, \bibinfo{pages}{6726--6733}.
\newblock


\bibitem[Ruff et~al\mbox{.}(2021)]%
        {ruff2021unifying}
\bibfield{author}{\bibinfo{person}{Lukas Ruff}, \bibinfo{person}{Jacob~R Kauffmann}, \bibinfo{person}{Robert~A Vandermeulen}, \bibinfo{person}{Gr{\'e}goire Montavon}, \bibinfo{person}{Wojciech Samek}, \bibinfo{person}{Marius Kloft}, \bibinfo{person}{Thomas~G Dietterich}, {and} \bibinfo{person}{Klaus-Robert M{\"u}ller}.} \bibinfo{year}{2021}\natexlab{}.
\newblock \showarticletitle{{A unifying review of deep and shallow anomaly detection}}.
\newblock \bibinfo{journal}{\emph{Proc. IEEE}} \bibinfo{volume}{109}, \bibinfo{number}{5} (\bibinfo{year}{2021}), \bibinfo{pages}{756--795}.
\newblock


\bibitem[Ruff et~al\mbox{.}(2018)]%
        {ruff2018}
\bibfield{author}{\bibinfo{person}{Lukas Ruff}, \bibinfo{person}{Robert Vandermeulen}, \bibinfo{person}{Nico Goernitz}, \bibinfo{person}{Lucas Deecke}, \bibinfo{person}{Shoaib~Ahmed Siddiqui}, \bibinfo{person}{Alexander Binder}, \bibinfo{person}{Emmanuel M{\"u}ller}, {and} \bibinfo{person}{Marius Kloft}.} \bibinfo{year}{2018}\natexlab{}.
\newblock \showarticletitle{{Deep one-class classification}}. In \bibinfo{booktitle}{\emph{International conference on machine learning}}. PMLR, \bibinfo{pages}{4393--4402}.
\newblock


\bibitem[Salehi et~al\mbox{.}(2021)]%
        {salehi2021unified}
\bibfield{author}{\bibinfo{person}{Mohammadreza Salehi}, \bibinfo{person}{Hossein Mirzaei}, \bibinfo{person}{Dan Hendrycks}, \bibinfo{person}{Yixuan Li}, \bibinfo{person}{Mohammad~Hossein Rohban}, {and} \bibinfo{person}{Mohammad Sabokrou}.} \bibinfo{year}{2021}\natexlab{}.
\newblock \showarticletitle{{A unified survey on anomaly, novelty, open-set, and out-of-distribution detection: Solutions and future challenges}}.
\newblock \bibinfo{journal}{\emph{arXiv preprint arXiv:2110.14051}} (\bibinfo{year}{2021}).
\newblock


\bibitem[Sehwag et~al\mbox{.}(2021)]%
        {sehwag2021ssd}
\bibfield{author}{\bibinfo{person}{Vikash Sehwag}, \bibinfo{person}{Mung Chiang}, {and} \bibinfo{person}{Prateek Mittal}.} \bibinfo{year}{2021}\natexlab{}.
\newblock \showarticletitle{{SSD: A unified framework for self-supervised outlier detection}}.
\newblock \bibinfo{journal}{\emph{arXiv preprint arXiv:2103.12051}} (\bibinfo{year}{2021}).
\newblock


\bibitem[Selvaraju et~al\mbox{.}(2017)]%
        {selvaraju2017}
\bibfield{author}{\bibinfo{person}{Ramprasaath~R Selvaraju}, \bibinfo{person}{Michael Cogswell}, \bibinfo{person}{Abhishek Das}, \bibinfo{person}{Ramakrishna Vedantam}, \bibinfo{person}{Devi Parikh}, {and} \bibinfo{person}{Dhruv Batra}.} \bibinfo{year}{2017}\natexlab{}.
\newblock \showarticletitle{{GradCAM: Visual explanations from deep networks via gradient-based localization}}. In \bibinfo{booktitle}{\emph{Proceedings of the IEEE international conference on computer vision}}. \bibinfo{pages}{618--626}.
\newblock


\bibitem[Sohn(2016)]%
        {sohn2016}
\bibfield{author}{\bibinfo{person}{Kihyuk Sohn}.} \bibinfo{year}{2016}\natexlab{}.
\newblock \showarticletitle{{Improved deep metric learning with multi-class n-pair loss objective}}. In \bibinfo{booktitle}{\emph{Advances in neural information processing systems}}. \bibinfo{pages}{1857--1865}.
\newblock


\bibitem[Sohn et~al\mbox{.}(2020)]%
        {sohn2020learning}
\bibfield{author}{\bibinfo{person}{Kihyuk Sohn}, \bibinfo{person}{Chun-Liang Li}, \bibinfo{person}{Jinsung Yoon}, \bibinfo{person}{Minho Jin}, {and} \bibinfo{person}{Tomas Pfister}.} \bibinfo{year}{2020}\natexlab{}.
\newblock \showarticletitle{{Learning and evaluating representations for deep one-class classification}}.
\newblock \bibinfo{journal}{\emph{arXiv preprint arXiv:2011.02578}} (\bibinfo{year}{2020}).
\newblock


\bibitem[Tack et~al\mbox{.}(2020)]%
        {tack2020csi}
\bibfield{author}{\bibinfo{person}{Jihoon Tack}, \bibinfo{person}{Sangwoo Mo}, \bibinfo{person}{Jongheon Jeong}, {and} \bibinfo{person}{Jinwoo Shin}.} \bibinfo{year}{2020}\natexlab{}.
\newblock \showarticletitle{{Csi: Novelty detection via contrastive learning on distributionally shifted instances}}.
\newblock \bibinfo{journal}{\emph{Advances in neural information processing systems}}  \bibinfo{volume}{33} (\bibinfo{year}{2020}), \bibinfo{pages}{11839--11852}.
\newblock


\bibitem[Taleb et~al\mbox{.}(2021)]%
        {taleb2021}
\bibfield{author}{\bibinfo{person}{Aiham Taleb}, \bibinfo{person}{Christoph Lippert}, \bibinfo{person}{Tassilo Klein}, {and} \bibinfo{person}{Moin Nabi}.} \bibinfo{year}{2021}\natexlab{}.
\newblock \showarticletitle{{Multimodal self-supervised learning for medical image analysis}}. In \bibinfo{booktitle}{\emph{International Conference on Information Processing in Medical Imaging}}. Springer, \bibinfo{pages}{661--673}.
\newblock


\bibitem[Tan and Bansal(2019)]%
        {tan2019lxmert}
\bibfield{author}{\bibinfo{person}{Hao Tan} {and} \bibinfo{person}{Mohit Bansal}.} \bibinfo{year}{2019}\natexlab{}.
\newblock \showarticletitle{{LXMERT: Learning cross-modality encoder representations from transformers}}.
\newblock \bibinfo{journal}{\emph{arXiv preprint arXiv:1908.07490}} (\bibinfo{year}{2019}).
\newblock


\bibitem[Tax(2002)]%
        {tax2002}
\bibfield{author}{\bibinfo{person}{David Martinus~Johannes Tax}.} \bibinfo{year}{2002}\natexlab{}.
\newblock \showarticletitle{{One-class classification: Concept learning in the absence of counter-examples}}.
\newblock \bibinfo{journal}{\emph{PhD Thesis, TU Delft}} (\bibinfo{year}{2002}).
\newblock


\bibitem[Tian et~al\mbox{.}(2020)]%
        {tian2020}
\bibfield{author}{\bibinfo{person}{Yonglong Tian}, \bibinfo{person}{Dilip Krishnan}, {and} \bibinfo{person}{Phillip Isola}.} \bibinfo{year}{2020}\natexlab{}.
\newblock \showarticletitle{{Contrastive multiview coding}}. In \bibinfo{booktitle}{\emph{Proceedings of the European Conference on Computer Vision (ECCV)}}. Springer, \bibinfo{pages}{776--794}.
\newblock


\bibitem[Wang and Isola(2020)]%
        {wang2020}
\bibfield{author}{\bibinfo{person}{Tongzhou Wang} {and} \bibinfo{person}{Phillip Isola}.} \bibinfo{year}{2020}\natexlab{}.
\newblock \showarticletitle{{Understanding contrastive representation learning through alignment and uniformity on the hypersphere}}. In \bibinfo{booktitle}{\emph{International Conference on Machine Learning}}. PMLR, \bibinfo{pages}{9929--9939}.
\newblock


\bibitem[Wu et~al\mbox{.}(2018)]%
        {wu2018}
\bibfield{author}{\bibinfo{person}{Zhirong Wu}, \bibinfo{person}{Yuanjun Xiong}, \bibinfo{person}{Stella~X Yu}, {and} \bibinfo{person}{Dahua Lin}.} \bibinfo{year}{2018}\natexlab{}.
\newblock \showarticletitle{{Unsupervised feature learning via non-parametric instance discrimination}}. In \bibinfo{booktitle}{\emph{Proceedings of the IEEE conference on computer vision and pattern recognition}}. \bibinfo{pages}{3733--3742}.
\newblock


\bibitem[Xiao et~al\mbox{.}(2020)]%
        {xiao2020vae}
\bibfield{author}{\bibinfo{person}{Xuerong Xiao}, \bibinfo{person}{Swetava Ganguli}, {and} \bibinfo{person}{Vipul Pandey}.} \bibinfo{year}{2020}\natexlab{}.
\newblock \showarticletitle{{VAE-Info-cGAN: Generating synthetic images by combining pixel-level and feature-level geospatial conditional inputs}}. In \bibinfo{booktitle}{\emph{Proceedings of the 13th ACM SIGSPATIAL International Workshop on Computational Transportation Science}}. \bibinfo{pages}{1--10}.
\newblock


\bibitem[Yang et~al\mbox{.}(2021)]%
        {yang2021}
\bibfield{author}{\bibinfo{person}{Jingkang Yang}, \bibinfo{person}{Kaiyang Zhou}, \bibinfo{person}{Yixuan Li}, {and} \bibinfo{person}{Ziwei Liu}.} \bibinfo{year}{2021}\natexlab{}.
\newblock \showarticletitle{{Generalized out-of-distribution detection: A survey}}.
\newblock \bibinfo{journal}{\emph{arXiv preprint arXiv:2110.11334}} (\bibinfo{year}{2021}).
\newblock


\bibitem[Yuan et~al\mbox{.}(2010)]%
        {msrtaxi11}
\bibfield{author}{\bibinfo{person}{Jing Yuan} {et~al\mbox{.}}} \bibinfo{year}{2010}\natexlab{}.
\newblock \showarticletitle{{T-Drive: Driving directions based on taxi trajectories}}. In \bibinfo{booktitle}{\emph{ACM SIGSPATIAL}}.
\newblock


\bibitem[Zenati et~al\mbox{.}(2018)]%
        {zenati2018adversarially}
\bibfield{author}{\bibinfo{person}{Houssam Zenati}, \bibinfo{person}{Manon Romain}, \bibinfo{person}{Chuan-Sheng Foo}, \bibinfo{person}{Bruno Lecouat}, {and} \bibinfo{person}{Vijay Chandrasekhar}.} \bibinfo{year}{2018}\natexlab{}.
\newblock \showarticletitle{{Adversarially learned anomaly detection}}. In \bibinfo{booktitle}{\emph{2018 IEEE International conference on data mining (ICDM)}}. IEEE, \bibinfo{pages}{727--736}.
\newblock


\bibitem[Zhai et~al\mbox{.}(2016)]%
        {zhai2016deep}
\bibfield{author}{\bibinfo{person}{Shuangfei Zhai}, \bibinfo{person}{Yu Cheng}, \bibinfo{person}{Weining Lu}, {and} \bibinfo{person}{Zhongfei Zhang}.} \bibinfo{year}{2016}\natexlab{}.
\newblock \showarticletitle{{Deep structured energy based models for anomaly detection}}. In \bibinfo{booktitle}{\emph{International conference on machine learning}}. PMLR, \bibinfo{pages}{1100--1109}.
\newblock


\bibitem[Zheng(2015)]%
        {zheng15}
\bibfield{author}{\bibinfo{person}{Y. Zheng}.} \bibinfo{year}{2015}\natexlab{}.
\newblock \showarticletitle{{Trajectory data mining: An overview}}.
\newblock \bibinfo{journal}{\emph{ACM TIST}} \bibinfo{volume}{6}, \bibinfo{number}{3} (\bibinfo{year}{2015}), \bibinfo{pages}{1--41}.
\newblock


\bibitem[Zhong et~al\mbox{.}(2020)]%
        {zhong2020}
\bibfield{author}{\bibinfo{person}{Zhun Zhong}, \bibinfo{person}{Liang Zheng}, \bibinfo{person}{Guoliang Kang}, \bibinfo{person}{Shaozi Li}, {and} \bibinfo{person}{Yi Yang}.} \bibinfo{year}{2020}\natexlab{}.
\newblock \showarticletitle{{Random erasing data augmentation}}. In \bibinfo{booktitle}{\emph{Proceedings of the AAAI Conference on Artificial Intelligence}}, Vol.~\bibinfo{volume}{34}. \bibinfo{pages}{13001--13008}.
\newblock


\bibitem[Zhou and Paffenroth(2017)]%
        {zhou2017anomaly}
\bibfield{author}{\bibinfo{person}{Chong Zhou} {and} \bibinfo{person}{Randy~C Paffenroth}.} \bibinfo{year}{2017}\natexlab{}.
\newblock \showarticletitle{{Anomaly detection with robust deep autoencoders}}. In \bibinfo{booktitle}{\emph{Proceedings of the 23rd ACM SIGKDD international conference on knowledge discovery and data mining}}. \bibinfo{pages}{665--674}.
\newblock


\bibitem[Zong et~al\mbox{.}(2018)]%
        {zong2018deep}
\bibfield{author}{\bibinfo{person}{Bo Zong}, \bibinfo{person}{Qi Song}, \bibinfo{person}{Martin~Renqiang Min}, \bibinfo{person}{Wei Cheng}, \bibinfo{person}{Cristian Lumezanu}, \bibinfo{person}{Daeki Cho}, {and} \bibinfo{person}{Haifeng Chen}.} \bibinfo{year}{2018}\natexlab{}.
\newblock \showarticletitle{{Deep autoencoding gaussian mixture model for unsupervised anomaly detection}}. In \bibinfo{booktitle}{\emph{International conference on learning representations}}.
\newblock


\end{thebibliography}
}

\clearpage
\appendix

\end{document}